\documentclass[10pt,journal,compsoc]{IEEEtran}
\ifCLASSOPTIONcompsoc
  \usepackage[nocompress]{cite}
\else
  \usepackage{cite}
\fi
\usepackage[pdftex]{graphicx}
\usepackage{amsmath}
\usepackage{array, boldline, makecell, booktabs}
\usepackage{amssymb}
\usepackage{caption}
\usepackage{acronym}
\usepackage{algorithmic}
\usepackage{breakurl}
\usepackage[edges]{forest}
\usepackage{upgreek}
\usepackage{dsfont}
\usepackage{rotating}
\usepackage{titlesec}
\titleformat{\paragraph}
{\normalfont\normalsize\em}{\theparagraph}{1em}{}
\titlespacing*{\paragraph}
{0pt}{3.25ex plus 1ex minus .2ex}{1.5ex plus .2ex}

\newcommand{\myparagraph}[1]{\vspace{1pt}\noindent{\bf{#1}}}
\newcommand{\newparagraph}[1]{\vspace{1pt}{{\em{#1}}}}
\usepackage{hyperref}
\hypersetup{
    colorlinks=true,
    linkcolor=blue,
    filecolor=magenta,      
    urlcolor=cyan,
}
\usepackage{review}
\usepackage{enumitem}
\usepackage{multirow}

\newcommand{\ycd}[1]{\textcolor{cyan}{}}

\def\eg{e.g.}
\def\ie{i.e.}

\begin{document}
\title{Semi-Supervised and Unsupervised \\ Deep Visual Learning: A Survey} 

\author{
Yanbei~Chen, 
Massimiliano~Mancini,
Xiatian~Zhu,
and 
Zeynep~Akata
\IEEEcompsocitemizethanks{\IEEEcompsocthanksitem This work was done when Y.Chen was with the University of
Tübingen.\protect \\
E-mail: yanbeic@gmail.com
\IEEEcompsocthanksitem M. Mancini is with the University of
Tübingen.\protect\\
E-mail: massimiliano.mancini@uni-tuebingen.de
\IEEEcompsocthanksitem X. Zhu is with the University of Surrey. E-mail: xiatian.zhu@surrey.ac.uk
\IEEEcompsocthanksitem Z. Akata is with the University of
Tübingen, MPI for Informatics and MPI for Intelligent Systems. E-mail: zeynep.akata@uni-tuebingen.de}
}

\IEEEtitleabstractindextext{
\begin{abstract}
State-of-the-art deep learning models are often trained with a large amount of costly labeled training data. However, requiring exhaustive manual annotations may degrade the model's generalizability in the limited-label regime. Semi-supervised learning and unsupervised learning offer promising paradigms to learn from an abundance of unlabeled visual data. Recent progress in these paradigms has indicated the strong benefits of leveraging unlabeled data to improve model generalization and provide better model initialization. In this survey, we review the recent advanced deep learning algorithms on semi-supervised learning (SSL) and unsupervised learning (UL) for visual recognition from a unified perspective. To offer a holistic understanding of the state-of-the-art in these areas, we propose a unified taxonomy. We categorize existing representative SSL and UL with comprehensive and insightful analysis to highlight their design rationales in different learning scenarios and applications in different computer vision tasks. Lastly, we discuss the emerging trends and open challenges in SSL and UL to shed light on future critical research directions.
\end{abstract}

}

\maketitle

\IEEEdisplaynontitleabstractindextext
\IEEEpeerreviewmaketitle

\ifCLASSOPTIONcompsoc
\IEEEraisesectionheading{\section{Introduction}\label{sec:introduction}}
\else
\section{Introduction}
\label{sec:introduction}
\fi

\IEEEPARstart{O}{ver} the last decade, deep learning algorithms and architectures~\cite{lecun2015deep,goodfellow2016deep} have been pushing the state of the art in a wide variety of computer vision tasks, ranging from object recognition~\cite{krizhevsky2012imagenet}, retrieval~\cite{schroff2015facenet}, detection~\cite{ren2015faster}, to segmentation~\cite{chen2017deeplab}. To achieve human-level performance, deep learning models are typically built by supervised training upon a tremendous amount of labeled training data.However, collecting large-scale labeled training sets manually is not only expensive and time-consuming, but may also be legally prohibited due to privacy, security, and ethics restrictions. Moreover, supervised deep learning models tend to memorize the labeled data and incorporate the annotator’s bias, 
which weakens their generalization to new scenarios with unseen data distributions in practice. 

Cheaper imaging technologies and more convenient access to web data, makes obtaining large unlabeled visual data no longer challenging. Learning from unlabeled data thus becomes a natural and promising way to scale models towards practical scenarios where it is  infeasible to collect a large labeled training set that covers all types of visual variations in illumination, viewpoint, resolution, occlusion, and background clutter induced by different scenes, camera positions, times of the day, and weather conditions.{Semi-supervised learning}~\cite{zhou2005semi,chapelle2010semi} and {unsupervised learning}~\cite{weinberger2006unsupervised,bengio2013representation,doersch2015unsupervised,chen2020simple} stand out as two most representative paradigms for leveraging unlabeled data. 
Built upon different assumptions, these paradigms are often developed independently, whilst sharing the same aim to learn more powerful representations and models using unlabeled data. 

\begin{figure}[!ht]
\centering
\includegraphics[width=0.47\textwidth]{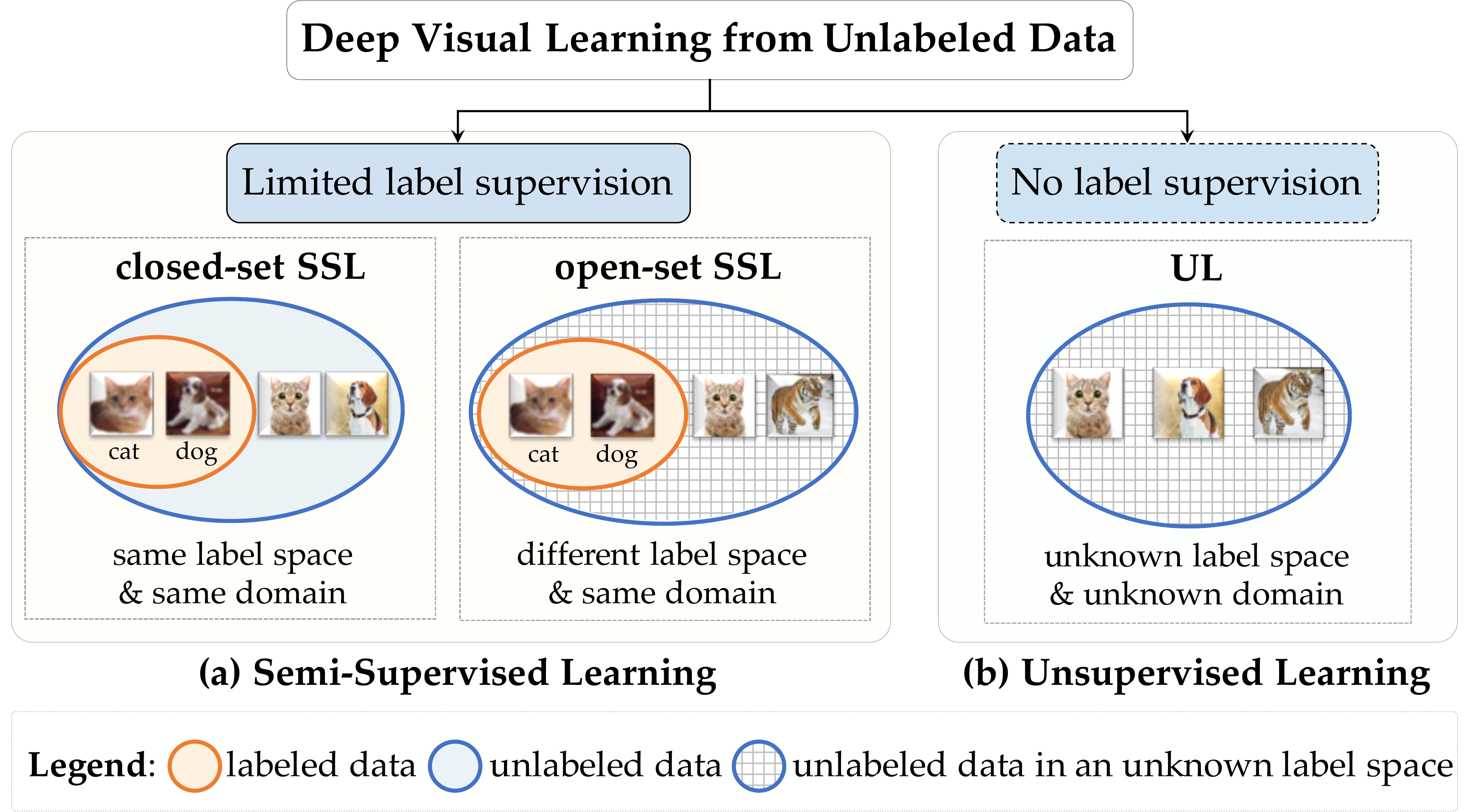}
\caption{An overview of semi-supervised and unsupervised learning paradigms -- both aim to learn from unlabeled data.}
\label{fig:overview}
\end{figure}
Figure \ref{fig:overview} summarizes the two paradigms covered in this survey, which both utilize unlabeled data for visual representation learning. 
According to whether label annotations are given for a small portion or none of the training data, 
we categorize the paradigms as semi-supervised learning, and unsupervised learning 
as defined explicitly in the following.

\begin{itemize}[itemsep=1pt,topsep=1pt]
\item[(a)]  {\bf Semi-Supervised Learning (SSL)}
aims to jointly learn from sparsely labeled data 
and a large amount of auxiliary unlabeled data often drawn from the same underlying data distribution as the labeled data. 
In standard {closed-set SSL}~\cite{zhu2006semi,chapelle2010semi}, the labeled and unlabeled data belong to the {same set of classes} from the same domain. 
In {open-set SSL}~\cite{chen2020semi,guo2020safe}, they may not lie in the same label space, i.e., the unlabeled data may contain {unknown} and/or mislabeled classes.

\item[(b)]  {\bf Unsupervised Learning (UL)} 
aims to learn from only unlabeled data without utilizing any task-relevant label supervision. Once trained, the model can be  
fine-tuned using labeled data to achieve better model generalization in a {downstream task}~\cite{he2019momentum}. 

\end{itemize}	 

Following the above definitions, let the sets of labeled data and unlabeled data be denoted as  $\mathcal{D}_l$ and $\mathcal{D}_u$. The overall unified learning objective for SSL and UL is: 
\begin{equation}
\underset{{\theta}}{\text{min}} 
\ \lambda_l \sum_{(\textbf{x},y) \in \mathcal{D}_L} \mathcal{L}_{\text{sup}}(\textbf{x},y, \theta) + 
\lambda_u \sum_{\textbf{x} \in \mathcal{D}_U} \mathcal{L}_{\text{unsup}}(\textbf{x}, \theta), 
\label{eq:common}
\end{equation}
where $\theta$ refers to the model parameters of a deep neural network (DNN); $\textbf{x}$ is an input image and $y$ is the corresponding label; $\mathcal{L}_{\text{sup}}$ and $\mathcal{L}_{\text{unsup}}$ are the supervised and unsupervised loss terms; $\lambda_l$ and $\lambda_u$ are balancing hyperparameters. In SSL, both loss terms are jointly optimized. In UL, only the unsupervised loss term is used for unsupervised model pre-training (\ie, $\lambda_l=0$). Although SSL and UL share the same rationale of learning with an unsupervised objective, they differ in the learning setup, leading to different unique challenges. Specifically, SSL assumes the availability of limited labeled data, and its core challenge is to expand the labeled set with abundant unlabeled data. UL assumes no labeled data for the main learning task and its key challenge is to learn task-generic representations from unlabeled data. 

We focus on providing a timely and comprehensive review of the advances in leveraging unlabeled data to improve model generalization, covering the representative state-of-the-art methods in SSL and UL, their application domains, to the emerging trends in self-supervised learning. Importantly, we propose a unified taxonomy of the advanced deep learning methods to offer researchers a systematic overview that helps to understand the current state of the art and identify open challenges for future research.

\myparagraph{Comparison with previous surveys.} Our survey is related to other surveys on semi-supervised learning~\cite{zhu2006semi,chapelle2010semi,van2020survey}, self-supervised learning~\cite{jing2020self,schmarje2020survey}, or both topics~\cite{qi2020small}. While these surveys mostly focus on a single particular learning setup~\cite{zhu2006semi,chapelle2010semi,van2020survey,jing2020self}, non-deep learning methods~\cite{zhu2006semi,chapelle2010semi}, or lacking a comprehensive taxonomy on methods and discussion on applications~\cite{qi2020small}, our work covers a wider review of representative SSL and UL algorithms involving unlabeled visual data. Importantly, we categorize the state-of-the-art SSL and UL algorithms with novel taxonomies and draw connections among different methods. Beyond intrinsic challenges with each learning paradigm, we distill their underlying connections from the problem and algorithmic perspectives, discuss unique insights into different existing techniques, and their practical applicability.

\myparagraph{Survey organization and contributions.} Our contributions are three fold. First, to our knowledge, this is the first deep learning survey of its kind to provide a comprehensive review of three prevalent machine learning paradigms in exploiting unlabeled data for visual recognition, including semi-supervised learning (SSL, \S \ref{sec:ssl}), unsupervised learning (UL, \S \ref{sec:ul}), and a further discussion on SSL and UL (\S \ref{sec:sslul}). Second, we provide a unified, insightful taxonomy and analysis of the existing methods in both the learning setup and model formulation to uncover their underlying algorithmic connections. Finally, we outlook the emerging trends and future research directions in \S \ref{sec:future} to shed light on those under-explored and potentially critical open avenues.

\section{Semi-Supervised Learning (SSL)}
\label{sec:ssl}

Semi-Supervised Learning (SSL)~\cite{zhu2006semi,chapelle2010semi} aims at exploiting large unlabeled data together with sparsely labeled data. SSL is explored in various application domains, such as image search~\cite{fergus2009semi}, medical data analysis~\cite{papernot2017semi}, web-page classification~\cite{blum1998combining}, document retrieval~\cite{nigam2000analyzing}, genetics and genomics~\cite{libbrecht2015machine}. More recently, SSL has been used for learning generic visual representations to facilitate many computer vision tasks such as image classification~\cite{berthelot2019mixmatch,berthelot2019remixmatch}, image retrieval~\cite{Jang_2020_CVPR}, object detection~\cite{Gao_2019_ICCV,tang2021humble}, semantic segmentation~\cite{Kalluri_2019_ICCV,Ouali_2020_CVPR,Ibrahim_2020_CVPR}, and pose estimation~\cite{Chen_2019_ICCV,radosavovic2018data,Mitra_2020_CVPR}. While our review mainly covers generic semi-supervised learners for image classification ~\cite{laine2016temporal,berthelot2019mixmatch,berthelot2019remixmatch,sohn2020fixmatch}, the ideas behind thembe generalized to solve other vision recognition tasks. 

We define the SSL problem setup and discuss its assumptions in \S \ref{sec:overviewssl}. We provide a taxonomy and analysis of the existing semi-supervised deep learning methods in \S \ref{sec:taxonomyssl}. 

\subsection{The Problem Setting of SSL}
\label{sec:overviewssl}

\begin{figure}[!t]
\centering
\includegraphics[width=0.48\textwidth]{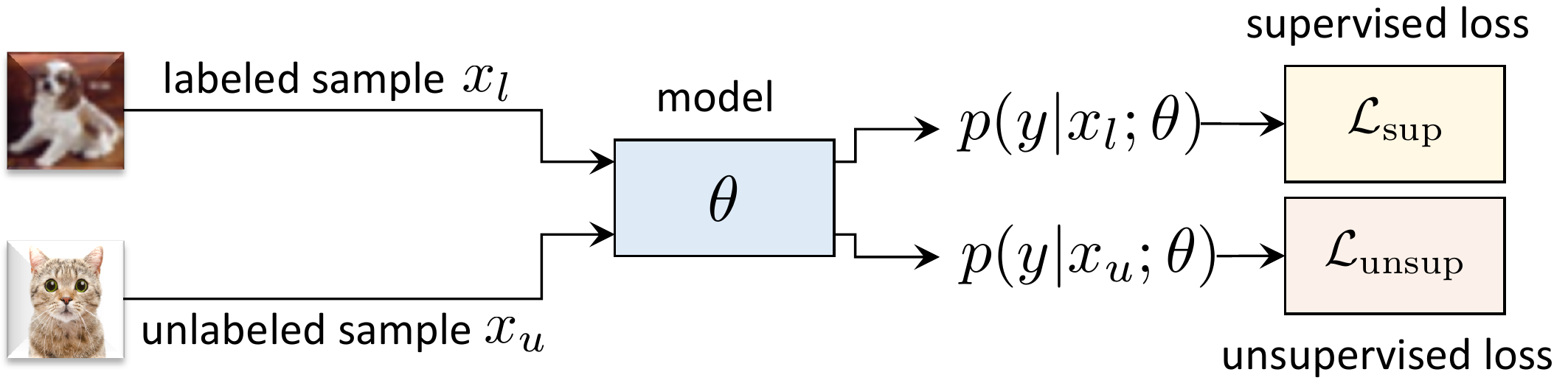}
\caption{Semi-supervised learning (SSL) aims to learn jointly from a small set of labeled and a large set of unlabeled data.}
\label{fig:ssl}
\end{figure}

\myparagraph{Problem Definition.} In SSL, we often have access to a limited amount of labeled samples $\mathcal{D}_l = \{\textbf{x}_{i,l}, y_i\}_{i=1}^{N_l}$ and a large amount of unlabeled samples $\mathcal{D}_u = \{\textbf{x}_{i,u}\}_{i=1}^{N_u}$. Each labeled sample $\textbf{x}_{i,l}$ belongs to one of $K$ class labels $\mathcal{Y} = \{y_k\}_{k=1}^{K}$. For training, the SSL loss function $\mathcal{L}$ for a deep neural network (DNN) $\theta$ can generally be expressed as Eq.~\eqref{eq:common}, \ie, $\mathcal{L} = \lambda_l \mathcal{L}_{\text{sup}} + \lambda_u \mathcal{L}_{\text{unsup}}$. In many SSL methods, the hyperparameters $\lambda_u$ in Eq.~\eqref{eq:common} is often a ramp-up weighting function (\ie, $\lambda=w(t)$ and $t$ is training iteration), which gradually increases the importance of the unsupervised loss term during training~\cite{laine2016temporal,tarvainen2017mean,oliver2018realistic,athiwaratkun2018there,chen2020semi}. At test time, the model is deployed to recognize the $K$ known classes. See Figure \ref{fig:ssl} for an illustration of SSL. 

\myparagraph{Evaluation Protocol.} To test whether an SSL model utilizes the unlabeled data effectively, two evaluation criteria are commonly adopted. First, the model needs to outperform its supervised baseline that learns from merely the labeled data. Second, when increasing the proportion of unlabeled samples in the training set, the improved margins upon the supervised baseline are expected to increase accordingly. Overall, these improved margins indicate the effectiveness and robustness of an SSL method. 

\myparagraph{Assumptions.} The main assumptions for SSL include the {smoothness assumption}~\cite{zhou2004learning} and {manifold assumption}~\cite{zhou2004learning,chapelle2010semi} -- the latter is also known as {cluster assumption}~\cite{chapelle2002cluster},  {structure assumption}~\cite{weston2012deep}, and {low-density separation assumption}~\cite{chapelle2005semi}. Specifically, the {smoothness assumption} considers that the nearby data points are likely to share the same class label. The {manifold assumption} considers data points lying within the same structure (\ie, the same cluster or manifold) should share the same class label. In other words, the former assumption is imposed locally for nearby data points, while the latter is imposed globally based on the underlying data structure formed by clusters or graphs. 
{
\setlength{\tabcolsep}{1.5pt}
\renewcommand{\arraystretch}{1.5}
\begin{table*}[!t]
\fontsize{8pt}{8pt}\selectfont
\centering
\caption{A taxonomy on semi-supervised deep learning methods, including five representative families in \S \ref{sec:ssl-consistency} -- \S \ref{sec:ssl-self}.}
\label{tab:ssl}
\begin{tabular}{l|l|l}
\hline
\bf Families of Models & \bf Model Rationale & \bf Representative Strategies and Methods \\ \hline
\multirow{6}{*}{\em Consistency regularization} 
& \bf Random augmentation & $\Pi$-model~\cite{sajjadi2016regularization,laine2016temporal}, ensemble transformations \cite{wang2020enaet}  \\
& \bf Adversarial perturbation & Virtual Adversarial Training (VAT)~\cite{miyato2015distributional,miyato2018virtual} \\
& \bf MixUp & MixMatch~\cite{berthelot2019mixmatch}, ICT~\cite{verma2019interpolation} \\
& \bf Automated augmentation & ReMixMatch~\cite{berthelot2019remixmatch}, UDA~\cite{xie2020unsupervised}, FixMatch~\cite{sohn2020fixmatch} \\
\cline{2-3}
& \bf Stochastic perturbation & Pseudo-Ensembles~\cite{bachman2014learning}, Ladder Network~\cite{rasmus2015semi}, Virtual Adversarial Dropout~\cite{park2017adversarial}, WCP \cite{zhang2020wcp}  \\
& \bf Ensembling & Temporal Ensembling~\cite{laine2016temporal}, Mean Teacher~\cite{tarvainen2017mean}, SWA~\cite{athiwaratkun2018there}, UASD~\cite{chen2020semi} \\ 
\hline
\multirow{3}{*}{\em Self-training} 
& \bf Entropy minimization & Pseudo-Label~\cite{lee2013pseudo}, MixMatch~\cite{berthelot2019mixmatch}, ReMixMatch~\cite{berthelot2019remixmatch}, Memory~\cite{chen2018semi} \\
& \bf Co-training  & Deep Co-training~\cite{qiao2018deep}, Tri-training~\cite{dong2018tri} \\
& \bf Distillation & 
model distillation (Noisy Student Training~\cite{xie2020self}, UASD~\cite{chen2020semi}), data distillation~\cite{radosavovic2018data} \\
\hline
\multirow{2}{*}{\em Graph-based regularization} 
& \bf Graph-based feature regularizer  & EmbedNN~\cite{weston2012deep}, Teacher Graph~\cite{luo2018smooth}, Graph Convolutional Networks~\cite{kipf2016semi} \\
& \bf Graph-based prediction regularizer & Label Propagation~\cite{iscen2019label}\\
\hline
\multirow{2}{*}{\em Deep generative models} 
& \bf Variational auto-encoders & Class-conditional VAE~\cite{kingma2014semi}, ADGM~\cite{maaloe2016auxiliary} \\ 
& \bf Generative adversarial networks & CatGAN~\cite{springenberg2015unsupervised}, FM-GAN~\cite{salimans2016improved}, ALI~\cite{dumoulin2016adversarially}, BadGAN~\cite{dai2017good}, Localized GAN \cite{qi2018global}  \\ 
\hline 
{\em Self-supervised learning} & 
\bf Self-supervision 
& S4L~\cite{zhai2019s4l}, SimCLR~\cite{chen2020simple}, SimCLRv2~\cite{chen2020big} \\
\hline
\end{tabular}
\end{table*}
}

\subsection{Taxonomy on SSL Algorithms}
\label{sec:taxonomyssl}

Existing SSL methods generally assume that the unlabeled data is closed-set and {task-specific}, \ie, all unlabeled training samples belong to a pre-defined set of classes. The idea shared by most existing works is to assign each unlabeled sample with a class label based on a certain underlying data structure, \eg, manifold structure~\cite{zhou2004learning,weston2012deep}, and graph structure~\cite{zhu2002learning}. We divide the most representative semi-supervised deep learning methods into five categories: {consistency regularization}, {self-training}, {graph-based regularization}, {deep generative models}, and {self-supervised learning} (Table \ref{tab:ssl}), and provide their general model formulations in \S \ref{sec:ssl-consistency}, \S \ref{sec:self-train}, \S \ref{sec:ssl-graph}, \S \ref{sec:DGMs} and \S \ref{sec:ssl-self}.

\subsubsection{Consistency Regularization} 
\label{sec:ssl-consistency}

\begin{figure}[!t]
\centering
\includegraphics[width=0.475\textwidth]{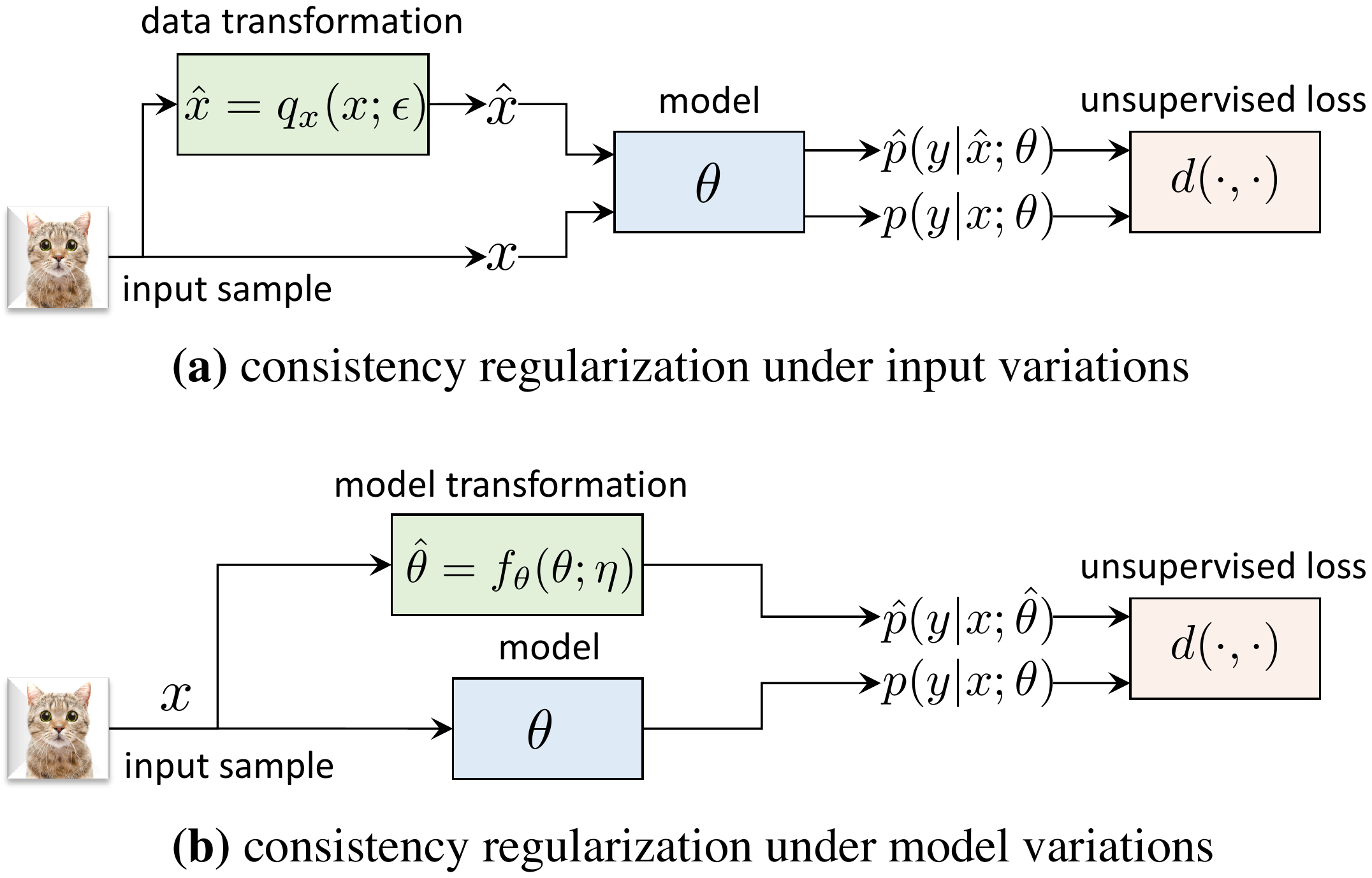}
\caption{In consistency regularization (\S \ref{sec:ssl-consistency}) {\bf (a)} input variations vs {\bf (b)} model variations, where 
variations can be induced by transformation on input data or model weights.}
\label{fig:ssl-consistency}
\end{figure}

\newparagraph{Consistency regularization} includes a number of successful and prevalent methods \cite{sajjadi2016regularization,laine2016temporal,tarvainen2017mean,miyato2018virtual,verma2019interpolation,berthelot2019mixmatch,berthelot2019remixmatch,suzuki2020adversarial,xie2020unsupervised}. The basic rationale is to enforce consistent model outputs under variations in the {input space} and (or) {model space}. The variations are often implemented by adding noise, perturbations or forming variants of the same input or model. Formally, the objective in case of input variation is: 
\begin{equation}
\underset{\theta}{\text{min}} \ 
\ \sum_{x \in \mathcal{D}}
d(p(y|x; \theta), \hat{p}(y|\hat{x}; \theta)),
\label{eq:consist1}
\end{equation}
and in case of model variation is: 
\begin{equation}
\underset{\theta}{\text{min}} \ 
\ \sum_{x \in \mathcal{D}} 
d(p(y|x; \theta), \hat{p}(y|x; \hat{\theta})).
\label{eq:consist2}
\end{equation}
In Eq.~\eqref{eq:consist1}, $\hat{x} = q_x(x; \epsilon)$ is a variant of the original input $x$, which is derived through a data transformation operation $ q_x(\cdot, \epsilon)$ with  $\epsilon$ being the noise added via data augmentation and stochastic perturbation. Similarly, in Eq.~\eqref{eq:consist2}, $\hat{\theta} = f_{\theta}(\theta; \eta)$ is a variant of the model $\theta$ derived via a transformation function $f_{\theta}(\cdot; \eta)$ with $\eta$ being the randomness added via stochastic perturbation on model weights and model ensembling strategies. In both equations, the consistency is measured as the discrepancy $d(\cdot, \cdot)$ between two network outputs $p(y|\cdot, \cdot)$ and $\hat{p}(y|\cdot, \cdot)$, typically quantified by divergence or distance metrics such as Kullback-Leibler (KL) divergence~\cite{miyato2018virtual}, cross-entropy~\cite{xie2020unsupervised}, and mean square error (MSE)~\cite{laine2016temporal}. See Figure~\ref{fig:ssl-consistency} for an illustration of consistency regularization. 
 
\paragraph{Consistency regularization under input variations}
\label{sec:coninput}

Various strategies aim to generate different versions of the same input ($\hat{x}$ in Eq.~\eqref{eq:consist1}) enforcing consistency (distributional smoothness) under input variations as depicted in Fig. \ref{fig:ssl-consistency} (a). Techniques range from simple {random augmentation}~\cite{sajjadi2016regularization,laine2016temporal}, to more advanced transformations such as {adversarial perturbation}~\cite{miyato2018virtual}, {MixUp}~\cite{zhang2017mixup,berthelot2019mixmatch}, as well as stronger {automated augmentation} such as AutoAugment~\cite{cubuk2018autoaugment}, 
RandAugment~\cite{cubuk2020randaugment}, CTAugment~\cite{berthelot2019remixmatch} and Cutout~\cite{devries2017improved}. Below we review these four streams of models.

\myparagraph{Random augmentation} is a standard data transformation strategy widely adopted~\cite{sajjadi2016regularization,laine2016temporal,tarvainen2017mean} via adding Gaussian noise and applying simple domain-specific jittering such as flipping and cropping on image data. For instance, the $\Pi$-model~\cite{sajjadi2016regularization,laine2016temporal}, applies random data augmentation on the same input and minimizes a consistency regularization term (MSE) between two network outputs. Ensemble transformations \cite{wang2020enaet} introduces more diverse data augmentation on input images, including spatial transformations (i.e., projective, affine, similarity, euclidean transformations) to modify the spatial aspect ratio, as well as non-spatial transformations to change the color, contrast, brightness, and sharpness. This way, the model learns representations invariant to various transformations. 

\myparagraph{Adversarial perturbation} augments the input data by adding adversarial noise aiming to alter the model predictions, \eg, reducing predictive confidence or changing the predicted correct label~\cite{szegedy2013intriguing,kurakin2016adversarial}. Adversarial noise is introduced for SSL to augment data and learn from the unlabeled data with adversarial transformations~\cite{miyato2015distributional,miyato2018virtual,najafi2019robustness,suzuki2020adversarial}. Virtual Adversarial Training (VAT)~\cite{miyato2015distributional,miyato2018virtual} is the {first} representative SSL method that perturbs input data adversarially. In VAT, a small adversarial perturbation is added to each input and a consistency regularization term (\ie, KL divergence) is imposed to encourage distributional robustness of the model against the virtual adversarial direction. Notably, it has been discovered that semi-supervised learning with adversarial perturbed unlabeled data does not only improve model generalization, but it also enhances robustness to adversarial attacks~\cite{najafi2019robustness,carmon2019unlabeled}. 

\myparagraph{MixUp} is a simple and data-agnostic augmentation strategy by performing linear interpolations on two inputs and their corresponding labels~\cite{zhang2017mixup}. It is also introduced as an effective regularizer for SSL~\cite{verma2019interpolation,berthelot2019mixmatch}. The Interpolation Consistency Training (ICT)~\cite{verma2019interpolation} interpolates two unlabeled samples and their network outputs. MixMatch~\cite{berthelot2019mixmatch} further considers to mix a labeled sample and unlabeled sample as the input, and the groundtruth label (of labeled data) and the predicted label (of unlabeled data) as the output targets. Both methods impose consistency regularization to guide the learning of a mapping between the interpolated input and interpolated output to learn from unlabeled data. 

\myparagraph{Automated augmentation} learns augmentation strategies from data to produce strong samples, alleviating the need to manually design domain-specific data augmentation~\cite{cubuk2018autoaugment,lim2019fast,ho2019population,zhang2019adversarial,cubuk2020randaugment}. It is introduced for SSL by enforcing that the predicted labels of a weakly-augmented or clean sample and its strongly augmented versions derived from {automated augmentation}~\cite{berthelot2019remixmatch,xie2020unsupervised} are consistent. Inspired by the advances of AutoAugment~\cite{cubuk2018autoaugment}, 
ReMixMatch~\cite{berthelot2019remixmatch} introduces CTAugment 
to learn an automated augmentation policy. 
Unsupervised Data Augmentation (UDA)~\cite{xie2020unsupervised} adopts  RandAugment~\cite{cubuk2020randaugment} to produce more diverse and strongly augmented samples by uniformly sampling a set of standard transformations based on the Python Image Library. Later on, FixMatch~\cite{sohn2020fixmatch} unifies multiple augmentation strategies including Cutout~\cite{devries2017improved}, CTAugment~\cite{berthelot2019remixmatch}, and RandAugment~\cite{cubuk2020randaugment} and produces even more strongly augmented samples as input. 

\paragraph{Consistency regularization under model variations}
\label{sec:conmodel}

To impose the predictive consistency under {model variations} (\ie, variations made in the model's parameter space) as in Eq.~\eqref{eq:consist2}, 
{stochastic perturbation}~\cite{bachman2014learning,rasmus2015semi,park2017adversarial} and {ensembling}~\cite{tarvainen2017mean,izmailov2018averaging,laine2016temporal} are proposed. Via non-identical models they produce different outputs for the same input -- a new model variant is denoted by $\hat{\theta}$ in Eq.~\eqref{eq:consist2}. Below we review these two streams of works as depicted in Fig. \ref{fig:ssl-consistency} (b). 

\myparagraph{Stochastic perturbation} introduces slight modifications on the model weights by adding Gaussian noise, dropout, or adversarial noise in a class-agnostic manner \cite{bachman2014learning,rasmus2015semi,park2017adversarial}. For example, Ladder Network injects layer-wise Gaussian noises into the network and minimizes a denoising L2 loss between outputs from the original network and the noisy-corrupted network~\cite{rasmus2015semi}. Pseudo-Ensemble applies dropout on the model's parameters to obtain a collection of models (a pseudo-ensemble), while minimizing the disagreements (KL divergence) between the pseudo-ensemble and the model~\cite{bachman2014learning}. Similarly, Virtual Adversarial Dropout introduces adversarial dropout to selectively deactivates network neurons and minimizes the discrepancy between outputs from the original model and the perturbed model~\cite{park2017adversarial}. Worst-Case Perturbations (WCP) introduces both addictive perturbations and drop connections on model parameters, where drop connections set certain model weights to zero to further change the network structure \cite{zhang2020wcp}. Notably, these  perturbation mechanisms promote the model robustness against noise in network parameters or structure.

\myparagraph{Ensembling} learns a set of models covering different regions of the version space~\cite{mitchell1982generalization,schapire1990strength,breiman2001random}. As demonstrated by seminal machine learning models such as boosting~\cite{freund1999short} and random forest~\cite{breiman2001random}, a set of different models can often provide more reliable predictions than a single model. Moreover, ensembling offers a rich inference uncertainty to mitigate the overconfidence issue in deep neural networks ~\cite{lakshminarayanan2017simple}. For SSL, an ensemble model is typically derived by computing an {exponential moving average} (EMA) or {equal average} in the prediction space or weight space~\cite{tarvainen2017mean,laine2016temporal,athiwaratkun2018there,chen2020semi}. Temporal Ensembling~\cite{laine2016temporal} and Mean Teacher~\cite{tarvainen2017mean} are two representatives that first propose to ensemble all the networks produced during training by maintaining an EMA in the weight space~\cite{tarvainen2017mean} or prediction space~\cite{laine2016temporal}. Stochastic Weight Averaging (SWA)~\cite{athiwaratkun2018there} applies an equal average of the model parameters in the weight space to provide a more stable target for deriving the consistency cost. {Later on}, Uncertainty-Aware Self-Distillation (UASD)~\cite{chen2020semi} computes an equal average of all the preceding model predictions during training to derive {soft targets} as the regularizer.

\myparagraph{\underline{Remarks.}} Consistency regularization can be treated as an  auxiliary task where the model learns from the unlabeled data to minimize its {predictive variance}  towards the variations in the {input space} or {weight space}. The {predictive variance} is generally quantified as the discrepancy between two predictive probability distributions or network outputs. By minimizing the consistency regularization loss, the model is encouraged to learn more powerful representations invariant towards variations added on each sample, without utilizing any additional label annotation.

\begin{figure}[!t]
\centering
\includegraphics[width=0.475\textwidth]{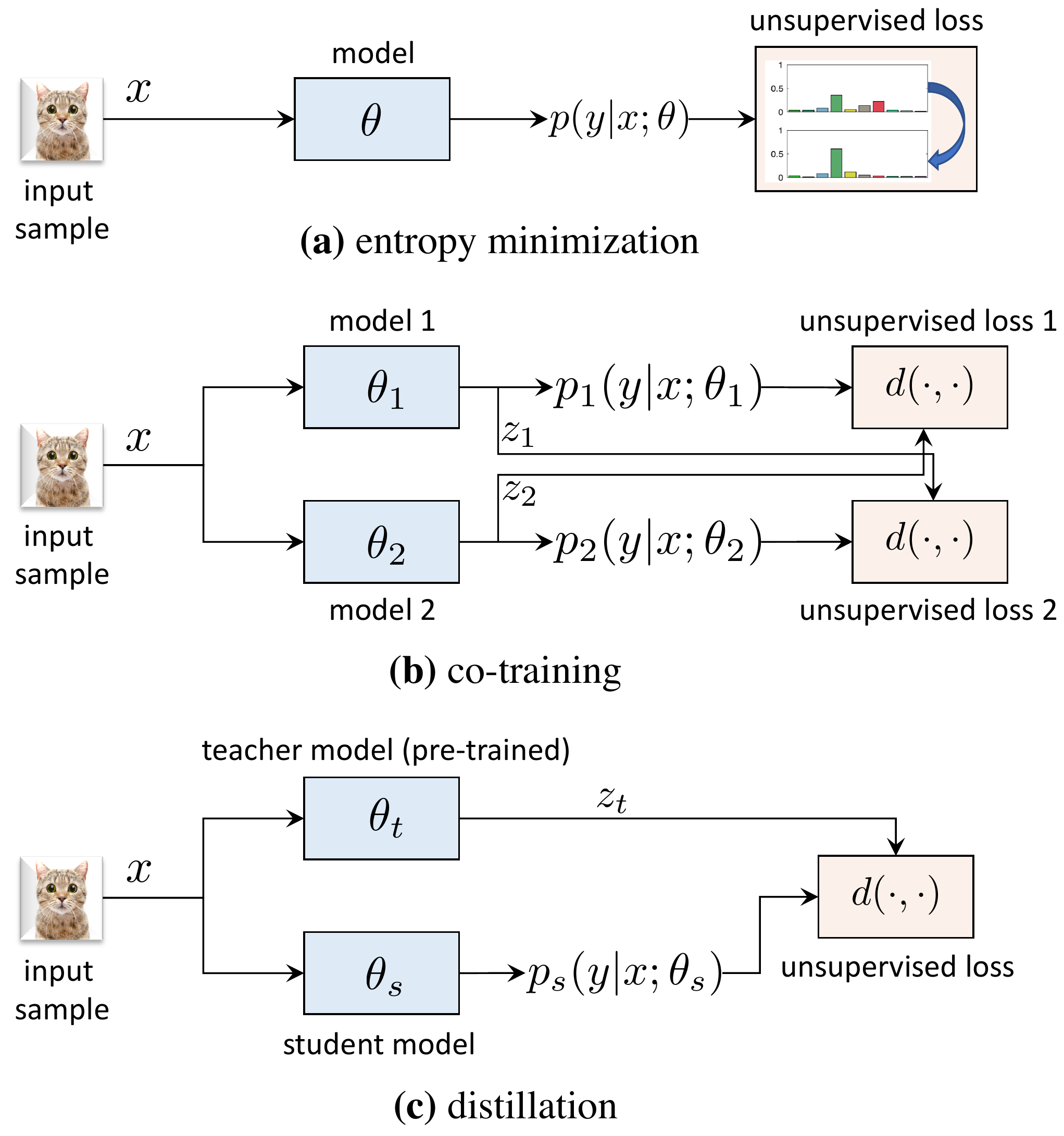}
\caption{In self-training, {\bf (a)} the model prediction is enforced to have low entropy, {\bf (b)} two models learn from each other and {\bf (c)} the student model learns from the teacher model. }
\label{fig:ssl-self-training}
\end{figure}

\subsubsection{Self-Training} 
\label{sec:self-train}

\newparagraph{Self-training} methods learn from unlabeled data by imputing the labels for samples predicted with high confidence~\cite{grandvalet2005semi,blum1998combining,nigam2000analyzing}. It is originally proposed for conventional machine learning models such as logistic regression~\cite{grandvalet2005semi}, bipartite graph~\cite{blum1998combining} and Naive Bayes classifier~\cite{nigam2000analyzing}. It is re-visited in deep neural networks to learn from massive unlabeled data along with limited labeled data. We review three representative lines of works in {self-training}, including {entropy minimization}, {co-training} and {distillation} as follows. See Figure \ref{fig:ssl-self-training} for an illustration of self-training. 

\myparagraph{Entropy minimization} regularizes the model training based on the {low density separation} assumption ~\cite{grandvalet2005semi,chapelle2005semi}, to enforce that the class decision boundary is placed in the low density regions. This is also in line with the {cluster assumption} and {manifold assumption} \cite{zhou2004learning,weston2012deep}, which hypothesizes that data points from the same class are likely to share the same {cluster} or {manifold}. Formally, the entropy minimization objective can be formulated as: 
\begin{equation}
\underset{\theta}{\text{min}} \ 
\ \sum_{x \in \mathcal{D}}
\Big{(}-\sum{_{j=1}^{K}}p(y_j|x; \theta) \log \ p(y_j|x; \theta)\Big{)},
\label{eq:ent}
\end{equation}
where $K$ refers to the number of classes. $p(y_j|x; \theta)$ is the probability of assigning the sample $x$ to the class $y_j$. This measures the class overlap. As a lower entropy indicates a higher confidence in model prediction, minimizing Eq.~\eqref{eq:ent} enforces each unlabeled sample to be assigned to the class predicted with the highest probability. Although {entropy minimization} is originally proposed for logistic regression to impute the labels of samples classified with high confidence~\cite{grandvalet2005semi}, it is later extended to train deep neural networks in SSL setting by minimizing the entropy of the class assignments either derived in the {prediction space}~\cite{lee2013pseudo,sajjadi2016mutual,miyato2018virtual,berthelot2019mixmatch,berthelot2019remixmatch,sohn2020fixmatch} or the {feature space}~\cite{chen2018semi}, as detailed next. 

Entropy minimization can be imposed in the {prediction space}, \eg, Pseudo-Label~\cite{lee2013pseudo} directly assigns each sample to the class label predicted with the maximum probability, which implicitly minimizes the entropy of model predictions. When pseudo labels are one-hot vectors, they could easily cause error propagation due to the wrong label assignments. To alleviate this risk, MixMatch~\cite{berthelot2019mixmatch} uses an ensemble of predictions over different input augmentations, and softly sharpens the one-hot pseudo labels with a temperature hyperparameter. Similarly, FixMatch~\cite{sohn2020fixmatch} assigns the one-hot labels only when the confidence scores of the model predictions are higher than a certain threshold. 

Entropy minimization can also be imposed in the {feature space}, as it is feasible to derive the class assignments based on proximities to class-level prototypes (\eg, cluster centers) in the feature space~\cite{snell2017prototypical,chen2018semi}. In~\cite{chen2018semi}, a Memory module learns a center per class that is derived based on proximities to all the cluster centers. Each unlabeled sample is assigned to the nearest cluster center by minimizing the entropy.

\myparagraph{Co-training} learns two or more classifiers on more than one view of the same sample coming from different sources~\cite{blum1998combining,zhou2005semi,nigam2000analyzing,qiao2018deep,dong2018tri}. Conceptually, a co-training framework~\cite{blum1998combining,nigam2000analyzing} trains two independent classifier models on two different but complementary data views and imputes the predicted labels in a cross-model manner. It is later extended for deep visual learning~\cite{qiao2018deep,dong2018tri,ke2019dual}, \eg, Deep Co-training (DCT)~\cite{qiao2018deep} trains a network with two or more classification layers, and passes different views (\eg, the original view and the adversarial view~\cite{goodfellow2014explaining}) to individual classifiers for co-training, while an unsupervised loss is imposed to minimize the similarity of predictions from different views. The basic idea of co-training can be extended from {dual-view}~\cite{qiao2018deep} to triple~\cite{dong2018tri} or multi-view~\cite{qiao2018deep} -- \eg, in Tri-training~\cite{dong2018tri}, three classifiers are trained together, with labels assigned to the unlabeled data when two of the classifiers agree on the predictions and the confidence scores are higher than a threshold. Formally, the deep co-training objective can be written as: 
\begin{equation}
\underset{\theta}{\text{min}} \ 
\ \sum_{x \in \mathcal{D}}
d(p_1(y|x; \theta_1), z_2) + d(p_2(y|x; \theta_2), z_1), 
\label{eq:cot}
\end{equation}
where $p_1, p_2$ are predictions of two independent classifiers $\theta_1, \theta_2$ trained on different data views. $d(\cdot,\cdot)$ introduces the similarity metric to learn from the imputed targets $z_1, z_2$ from each other, \eg, cross-entropy on one-hot targets~\cite{dong2018tri}, or Jensen-Shannon divergence between output targets~\cite{qiao2018deep}. 

\myparagraph{Distillation} is originally proposed to transfer the {knowledge} learned by a teacher model to a student model, where the {soft targets} from the teacher model (\eg, an ensemble of networks or a larger network) can serve as an effective regularizer or a model compression strategy to train a student model~\cite{hinton2015distilling,bucilua2006model,ba2014deep}. Recent works in SSL use distillation to impute learning targets on the unlabeled data for training the student network~\cite{radosavovic2018data,xie2020self,yalniz2019billion,chen2020semi}. Formally, an unsupervised distillation objective is introduced on a student model $\theta_s$ to learn from the unlabeled data as: 
\begin{equation}
\underset{\theta}{\text{min}} \ 
\ \sum_{x \in \mathcal{D}}
d(p_s(y|x; \theta_s), z_t),
\label{eq:distill}
\end{equation}
where the student prediction $p_s$ is enforced to align with the targets $z_t$ produced by a teacher model $\theta_t$ on either the unlabeled data or all the data. Compared to co-training (Eq.~\eqref{eq:cot}), distillation in SSL (Eq.~\eqref{eq:distill}) does not optimize multiple networks simultaneously, but instead trains more than one network in different stages. 
In distillation, the existing works can be further grouped into {model distillation} and {data distillation}, which generate {learning targets} for unlabeled data using the teacher model output or multiple forward passes of the same input data, as detailed next. 

In {model distillation}, labels from a teacher are assigned to a student~\cite{xie2020self,yalniz2019billion,chen2020semi}. The teacher model can be formed, \eg, via a pre-trained model or an ensemble of models. In Noisy Student Training~\cite{xie2020self}, an iterative self-training process iterates the teacher-student training by first training a teacher to impute labels on unlabeled data for the student, and reuses the student as the teacher in the next iteration. In Uncertainty-Aware Self-Distillation (USAD)~\cite{chen2020semi}, the teacher averages all the preceding network predictions to impute labels on unlabeled data for updating the student network itself. In {model distillation},  both soft targets and one-hot labels from the teacher model can serve as the learning targets on the unlabeled data~\cite{xie2020self,chen2020semi}.

In {data distillation}, the teacher model predicts learning targets on unlabeled data by ensembling the outputs of the same input under different data transformations~\cite{radosavovic2018data}. Specifically, the ensembled teacher predictions (\ie, soft targets) are derived by averaging the outputs of the same inputs under multiple data transformations; while the student model is then trained with the soft targets. {Data distillation} transforms the input data multiple times rather than training multiple networks to impute the ensembled predictions on unlabeled data. This is similar to {consistency regularization with random data augmentation}; however, in {data distillation}, two training stages are involved -- the first stage involves pre-training the teacher model; while the second stage involves training the student network to mimic the teacher model by distillation. 

\myparagraph{\underline{Remarks.}} Similar to {consistency regularization}, {self-training} can be considered as an unsupervised auxiliary task learned along with the supervised learning task. In general, it also enforces the {predictive invariance} towards instance-wise variations or the teacher's predictions. However, {self-training} differs in design. While consistency regularization generally trains one model, {self-training} may require more than one model to be trained, \eg, {co-training} requires at least two models trained in parallel while {distillation} requires to train a teacher and a student model sequentially.

\subsubsection{Graph-based Regularization} 
\label{sec:ssl-graph}
\begin{figure}[!t]
\centering
\includegraphics[width=0.48\textwidth]{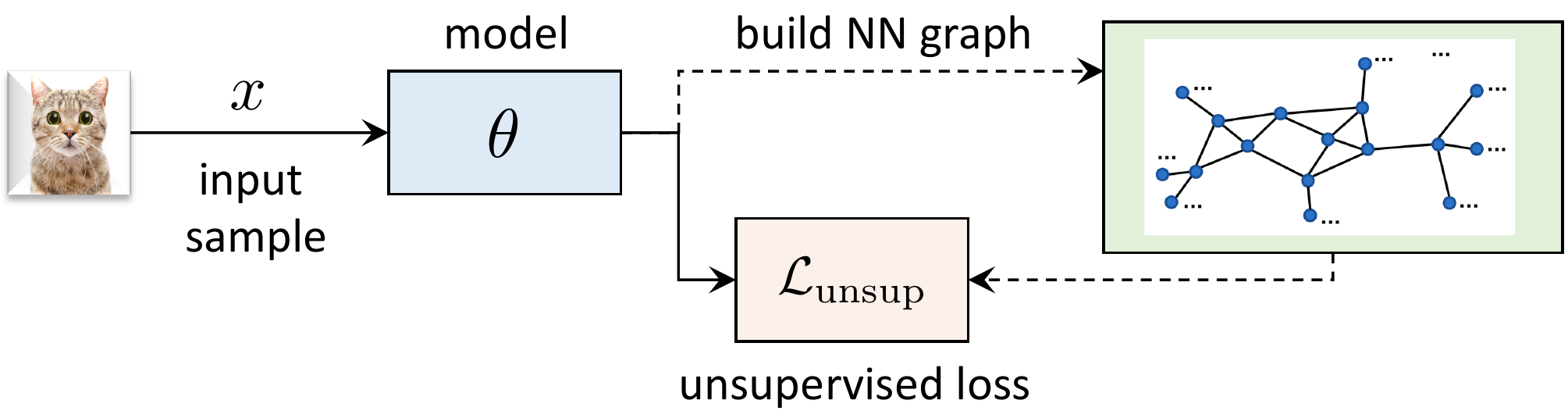}
\caption{In graph-based regularization (\S \ref{sec:ssl-graph}) pseudo labels are propagated over the Nearest Neighbor graph based on neighbourhood consistency and an unsupervised regularization term is imposed on the feature or prediction space.}
\label{fig:ssl-graph}
\end{figure}

\newparagraph{Graph-based regularization} is a family of transductive learning methods originally proposed for non-deep semi-supervised learning algorithms~\cite{zhu2002learning,zhu2003semi,zhou2004learning,belkin06a,wang2013dynamic}, such as transductive Support Vector Machine~\cite{zhou2004learning,belkin06a} and Gaussian random field model~\cite{zhu2003semi}. Most algorithms from this family build a weighted graph to exploit relationships among the data samples. Specifically, both labeled and unlabeled samples are represented as nodes, while the edge weights encode the similarities between different samples. The labels can be propagated over the graph based on the {smoothness assumption}~\cite{zhou2004learning}, \ie, neighboring data points should share the same class label as shown in Figure \ref{fig:ssl-graph}. 

A graph-based regularization term is used in model optimization by imposing various forms of smoothness constraints to minimize the pairwise similarities between nearby data points. Graph-based regularization is later reformulated for semi-supervised learning with deep neural networks, such as EmbedNN~\cite{weston2012deep}, Graph Convolutional Network~\cite{kipf2016semi,jiang2019semi}, Teacher Graph~\cite{luo2018smooth}, and Label Propagation~\cite{iscen2019label}.  Although this line of works share the same smoothness assumption for model optimization, graph-based regularization can be imposed differently in either the {feature space} or {prediction space}, detailed as follows. 

\myparagraph{Graph-based feature regularization} is typically done by building a learnable nearest neighbor (NN) graph that augments the original DNN to encode the affinity between data points in the feature space, as represented by EmbedCNN~\cite{weston2012deep} and Teacher Graph~\cite{luo2018smooth}. Each node in the graph is encoded by the visual feature extracted from the intermediate network layer or the output from the last layer; while an affinity matrix $W_{ij}$ is computed to encode the pairwise similarities between all the nodes. To exploit unlabeled data, a graph-based regularization term can be formed as a metric learning loss, such as the {margin-based contrastive loss} for Siamese networks~\cite{bromley1993signature,hadsell2006dimensionality} which constrains feature learning by enforcing the local smoothness:
\begin{equation}
\underset{\theta}{\text{min}} 
\sum_{x_i, x_j \in \mathcal{D}}
\left\{  \begin{aligned}  
&
\|h(x_i)-h(x_j)\|^2, & \text{if} \ W_{ij}{=}1 \\
&
\text{max}(0, m-||h(x_i)-h(x_j)||)^2, & \text{if} \ W_{ij}{=}0
\end{aligned} \right.
\label{eq:graph}
\end{equation}
ensuring that features $h(x_i), h(x_j)$ of nearest neighbors (\ie, $W_{ij}{=}1$) are close to and dissimilar pairs (\ie, $W_{ij}{=}0$) are away from each other with a distance margin $m$. 

Beyond augmenting a DNN with a graph, a more flexible way is to use graph convolutions, \ie, Graph Convolutional Networks (GCN)~\cite{kipf2016semi}, which derive new feature representations for each node subject to the graph structure \cite{jiang2019semi,lin2020shoestring}. Specifically, a GCN takes the data and affinity matrix as input, and learns to estimate the class labels of unlabeled data under a supervised cross-entropy loss on labeled data. 

\myparagraph{Graph-based prediction regularization} operates in the prediction space~\cite{iscen2019label,li2020density}, as in Label Propagation~\cite{iscen2019label}. Driven by the same rationale of building a learnable NN-graph as above, in label propagation, an NN-graph encidong the similarity between data points is used to propagate the labels from the labeled data to the unlabeled data based on transitivity via with a cross-entropy loss. While being similar to the approach {Pseudo-Labels}~\cite{lee2013pseudo}, the propagated labels are derived with an external NN-graph that encodes the global manifold structure. Further, label propagation on the graph and the update of DNN are performed alternatively to propagate more reliable labels. 

\myparagraph{\underline{Remarks.}} Graph-based regularization shares several similarities with consistency regularization and self-training in SSL. First, it introduces an unsupervised auxiliary task to train a DNN with propagated learning targets (\eg, pseudo labels) on the unlabeled data. Second, its learning objective can be formulated as a  cross-entropy loss or metric learning loss. Notably, while consistency regularization and self-training are inductive approaches that estimate a learning target per instance, graph-based regularization methods are {transductive} approaches that propagate learning targets based on a graph constructed on the dataset. Beyond concrete details, however, the three techniques all share the same fundamental idea of seeking for unsupervised targets.

\subsubsection{Deep Generative Models}
\label{sec:DGMs}

\begin{figure}[!t]
\centering
\includegraphics[width=0.48\textwidth]{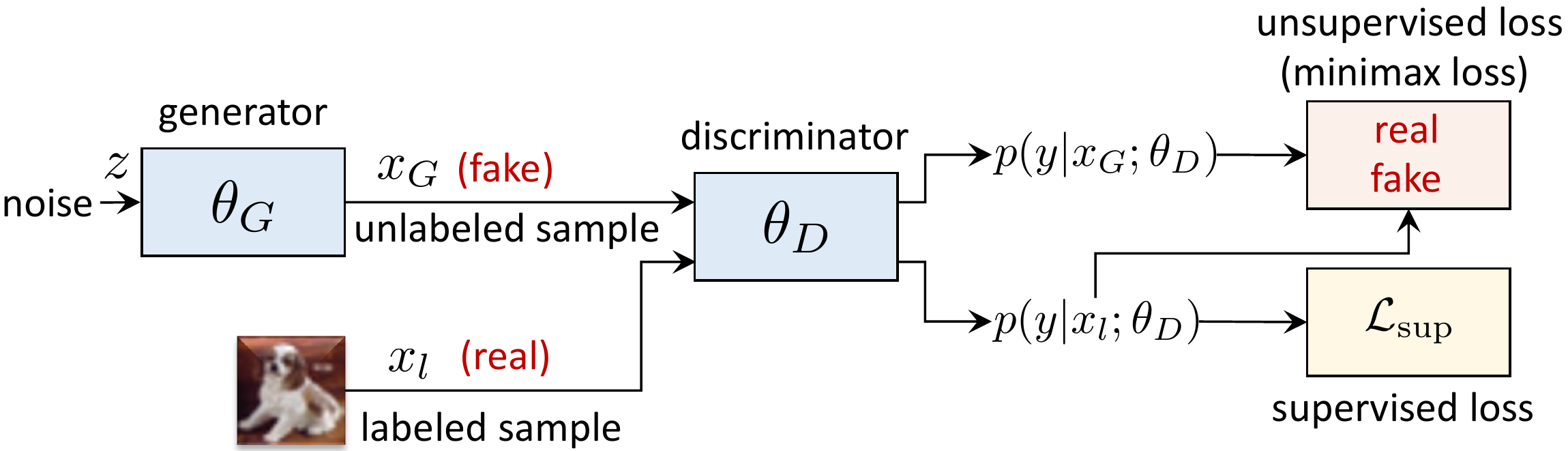}
\caption{In GAN-based deep generative models (\S \ref{sec:DGMs}), the discriminator assigns the labeled samples to the $K$ classes and the generated unlabeled data to an auxiliary class ($K+1$). At test time, the discriminator acts as the classifier.}
\label{fig:ssl-gdm}
\end{figure}
\newparagraph{Deep generative models} 
are a class of unsupervised learning models that learn to approximate the data distributions without labels~\cite{kingma2013auto,goodfellow2014generative}. 
By integrating the generative unsupervised learning concept into a supervised model, 
a semi-supervised learning framework can be formulated to unify the merits of supervised and unsupervised learning.
Two main streams of deep generative models are Variational Auto-Encoders (VAEs) and Generative Adversarial Networks (GANs), as detailed below.   
See Figure \ref{fig:ssl-gdm} for an illustration of a GAN framework for SSL. 

\myparagraph{Variational auto-encoders} (VAEs) are probabilistic models based on variational inference for unsupervised learning of a complex data distribution~\cite{kingma2013auto,doersch2016tutorial}. A standard VAE model contains a network that encodes an input sample to a latent variable and a network that decodes the latent variable to reconstruct the input; maximizing a variational lower bound. In semi-supervised learning \cite{kingma2014semi,maaloe2016auxiliary,ehsan2017infinite}, an unsupervised VAE model is generally combined with a supervised classifier. For instance, to predict task-specific class information required in SSL, Class-conditional VAE~\cite{kingma2014semi} and ADGM~\cite{maaloe2016auxiliary} introduce the class label as an extra latent variable in the latent feature space to explicitly disentangle the class information (content) and the stochastic information (style), and impose an explicit classification loss on the labeled data along with the vanilla VAE loss.

\myparagraph{Generative adversarial networks} (GANs)~\cite{goodfellow2014generative} learn to capture the data distribution by an adversarial minimax game. Specifically, a {generator} is trained to generate as realistic images as possible while a {discriminator} is trained to discriminate between real and generated samples. When re-formulated as a semi-supervised representation learner, GANs can leverage the benefits of both {unsupervised generative modeling} and {supervised discriminative learning} \cite{springenberg2015unsupervised,salimans2016improved,dumoulin2016adversarially,kumar2017semi,li2017triple,dai2017good,RAYLLS16,RAMTLS16,XLSA18,XSSA19}. 

The generic idea is to augment the standard GAN framework with  supervised learning on the labeled real samples (\ie, discriminative) and unsupervised learning on the generated samples. Formally, this enhances the original discriminator with an extra supervised learning capability. For example, Categorical GAN (CatGAN)~\cite{springenberg2015unsupervised} introduces a $K$-class discriminator, and minimizes a supervised cross-entropy loss on the real labeled samples, while imposing a uniform distribution constraint on the generated samples by maximizing the prediction's entropy. Similarly, feature matching GAN (FM-GAN)~\cite{salimans2016improved}, ALI~\cite{dumoulin2016adversarially}, BadGAN~\cite{dai2017good} and Localized GAN \cite{qi2018global} formulate a $(K{+}1)$-class discriminator for SSL, whereby a real labeled sample $x_l$ is considered as one of the $K$ classes and a generated sample $x_G$ as the $(K+1)_{\text{th}}$ class. The supervised and unsupervised learning objective for the $(K{+}1)$-class discriminator is formulated as;
\begin{align}
     & 
\underset{\theta}{\text{max}} 
\ \sum_{x \in \mathcal{D}}
\text{log} \ p(y|x_l,y{<}K{+}1),  
\label{eq:gans}
\\
     & 
\underset{\theta}{\text{max}} 
\ \sum_{x \in \mathcal{D}}
\text{log} \ (1{-}p(y{=}K{+}1|x_l)) 
- \text{log} \ p(y{=}K{+}1|x_G), 
\label{eq:ganu}
\end{align}
where Eq.~\eqref{eq:gans} is the supervised classification loss on the labeled samples $x_l$; Eq.~\eqref{eq:ganu} is an unsupervised GAN loss that discriminates between the real labeled samples $x_l$ and the generated fake samples $x_G$ from the image generator. To constrain the generated samples, Localized GAN \cite{qi2018global} introduces a regularizer on the generator to ensure the generated samples lie in the neighborhood of an original sample on the manifold, thus allowing to train a locally consistent classifier based on the generated samples in a semi-supervised fashion.

\myparagraph{\underline{Remarks.}} Unlike previously discussed discriminative SSL techniques DGMs can naturally learn from unlabeled data {without} the need to estimate their labels. In other words, DGMs are native unsupervised representation learners. To enable SSL in DGMs, the key in model reformulation is thus to integrate the label supervision into training, \eg, adding a class label latent variable in VAEs or an extra class discriminator in GANs. Further, one also needs to tackle more difficult model optimization in a GAN framework. 

\subsubsection{Self-Supervised Learning}
\label{sec:ssl-self}
 
\newparagraph{Self-supervised learning} is a class of unsupervised representation learners designed based on unsupervised surrogate (pretext) tasks~\cite{dosovitskiy2014discriminative,dosovitskiy2015discriminative,doersch2015unsupervised,noroozi2016unsupervised,noroozi2017representation,gidaris2018unsupervised}. Self-supervision differs from self-training algorithms in \S \ref{sec:self-train}, as self-supervised learning objectives are task-agnostic and could be trained without any label supervision. The former is originally proposed to learn from only unlabeled data with task-agnostic unsupervised learning objectives, but it is also explored for SSL~\cite{zhai2019s4l,chen2020big,chen2020simple}. In SSL, task-agnostic self-supervision signals on all training data are often integrated with a supervised learning objective on labeled data. For instance, S4L~\cite{zhai2019s4l} uses self-supervision for SSL based on multiple self-supervision signals such as predicting rotation degree~\cite{gidaris2018unsupervised} and enforcing invariance to exemplar transformation~\cite{dosovitskiy2014discriminative} to train the model along with supervised learning. SimCLR~\cite{chen2020simple} and SimCLRv2~\cite{chen2020big} are follow-up works introducing self-supervised contrastive learning for task-agnostic unsupervised pre-training, followed by supervised or semi-supervised fine-tuning with label supervision as the downstream task.  

\myparagraph{\underline{Remarks.}} A unique advantage of self-supervision for SSL is that task-specific label supervision is not required during training. While the aforementioned semi-supervised learners typically solve a supervised task and an auxiliary unsupervised task jointly, self-supervised semi-supervised learners can be trained in a fully task-agnostic fashion. This suggests the great flexibility of self-supervision for SSL. Thus, the self-supervised training can be introduced as unsupervised pre-training or as an auxiliary unsupervised task solved along with supervised learning. Although self-supervision is relatively new for SSL, it has been more widely explored for unsupervised learning, which is detailed more extensively in \S \ref{sec:pretext} and \S \ref{sec:dis}. 

{
\setlength{\tabcolsep}{1.5pt}
\renewcommand{\arraystretch}{1.5}
\begin{table*}[!t]
\fontsize{8pt}{8pt}\selectfont
\centering
\caption{A taxonomy of unsupervised deep learning methods, including three representative families in \S \ref{sec:pretext} -- \S \ref{sec:ul-dgm}.}
\label{tab:ul}
\begin{tabular}{l|l|l|l}
\hline
\bf Families of Models & \bf Model Rationale & \multicolumn{2}{l}{\bf Representative Strategies and Methods} \\ \hline
\multirow{2}{*}{\em Pretext tasks} 
& \bf Pixel-level & \multicolumn{2}{l}{reconstruction~\cite{hinton2006reducing,masci2011stacked}, inpainting~\cite{pathak2016context}, MAE \cite{he2022masked},  denoising~\cite{vincent2008extracting}, colorization~\cite{zhang2016colorful,zhang2017split,larsson2017colorization}} \\
& \bf Instance-level & \multicolumn{2}{l}{predict image rotations~\cite{gidaris2018unsupervised}, scaling and tiling~\cite{noroozi2017representation}, patch ordering~\cite{doersch2015unsupervised}, patch re-ordering~\cite{noroozi2016unsupervised}} \\
\hline
\multirow{5}{*}{\em Discriminative models} 
& \multirow{3}{*}{\bf Instance discrimination} 
    & negative sampling & large batch size (SimLR~\cite{chen2020simple}), memory bank (InstDis~\cite{wu2018unsupervised}), queue (MoCo~\cite{he2019momentum}) \\ 
    & 
    & input transformation & data augmentation (PIRL~\cite{misra2020self}), multi-view augmentation (CMC~\cite{tian2019contrastive}) \\
   \cline{3-4}
  & & \multirow{1}{*}{negative-sample-free}  
& simple siamese (SimSiam~\cite{chen2020exploring}), Bootstrap (BYOL~\cite{grill2020bootstrap}), 
DirectPred~\cite{tian2021understanding} \\
\cline{2-4} 
& \multirow{2}{*}{\bf Deep clustering} 
    & offline clustering & DeepCluster~\cite{caron2018deep}, JULE~\cite{yang2016joint}, SeLa~\cite{asano2019self} \\
    & & online clustering & 
    IIC~\cite{ji2019invariant}, PICA~\cite{huang2020deep}, AssociativeCluster~\cite{haeusser2018associative}, SwAV~\cite{caron2020unsupervised} \\
\hline
\multirow{2}{*}{\em Deep generative models} 
& \bf Discriminator-level & \multicolumn{2}{l}{DCGAN~\cite{radford2015unsupervised}, Self-supervised GAN~\cite{chen2019self}, 
Transformation GAN~\cite{wang2020transformation}} \\ 
& \bf Generator-level & \multicolumn{2}{l}{BiGAN~\cite{brock2018large}, BigBiGAN~\cite{donahue2019large}} \\
\hline
\end{tabular}
\end{table*}
}

\section{Unsupervised Learning (UL)}
\label{sec:ul}

\newparagraph{Unsupervised Learning} (UL) aims to learn representations without utilizing any label supervision. The learned representation is not only expected to capture the underlying semantic information, but also be transferable to tackle unseen {downstream tasks} such as visual recognition, detection, and segmentation~\cite{he2019momentum}, visual retrieval~\cite{wang2015unsupervised}, and tracking~\cite{vondrick2018tracking}. 

UL is attractive in computer vision for multiple reasons. First, due to costly label annotations, large labeled datasets may not be available in many application scenarios, \eg, medical imaging~\cite{aganj2018unsupervised}. Second, as there are often data/label distribution drifts (or gaps) across tasks and application scenarios, pre-training on a large labeled dataset cannot always guarantee good model initialization for unseen situations~\cite{he2019rethinking}. Third, UL could supply strong pre-trained models that may perform on par with or even outperform supervised pre-training~\cite{he2019momentum,chen2020simple,feichtenhofer2021large}. 

\myparagraph{\underline{Remarks.}} {\bf UL} and {\bf SSL} share the same aim to learn from unlabeled data, and leverage similar modeling principles to formulate unsupervised surrogate supervision signals without any label annotation. However, instead of assuming the availability of task-specific information (\ie, class labels) as in SSL, UL considers model learning from purely task-agnostic unlabeled data. Given that unlabeled data are abundantly available in different scenarios (\eg, Internet), UL offers an appealing strategy to provide good pre-trained models that could facilitate various downstream tasks.  

Focusing on unsupervised visual learners trained on image classification datasets, we define the UL problem setup in \S \ref{sec:overviewul}, and provide a taxonomy and analysis of the existing representative  unsupervised deep learning methods in \S \ref{sec:taxonomyul}. 

\subsection{The Problem Setting of UL}
\label{sec:overviewul}

\begin{figure}[!t]
\centering
\includegraphics[width=0.4\textwidth]{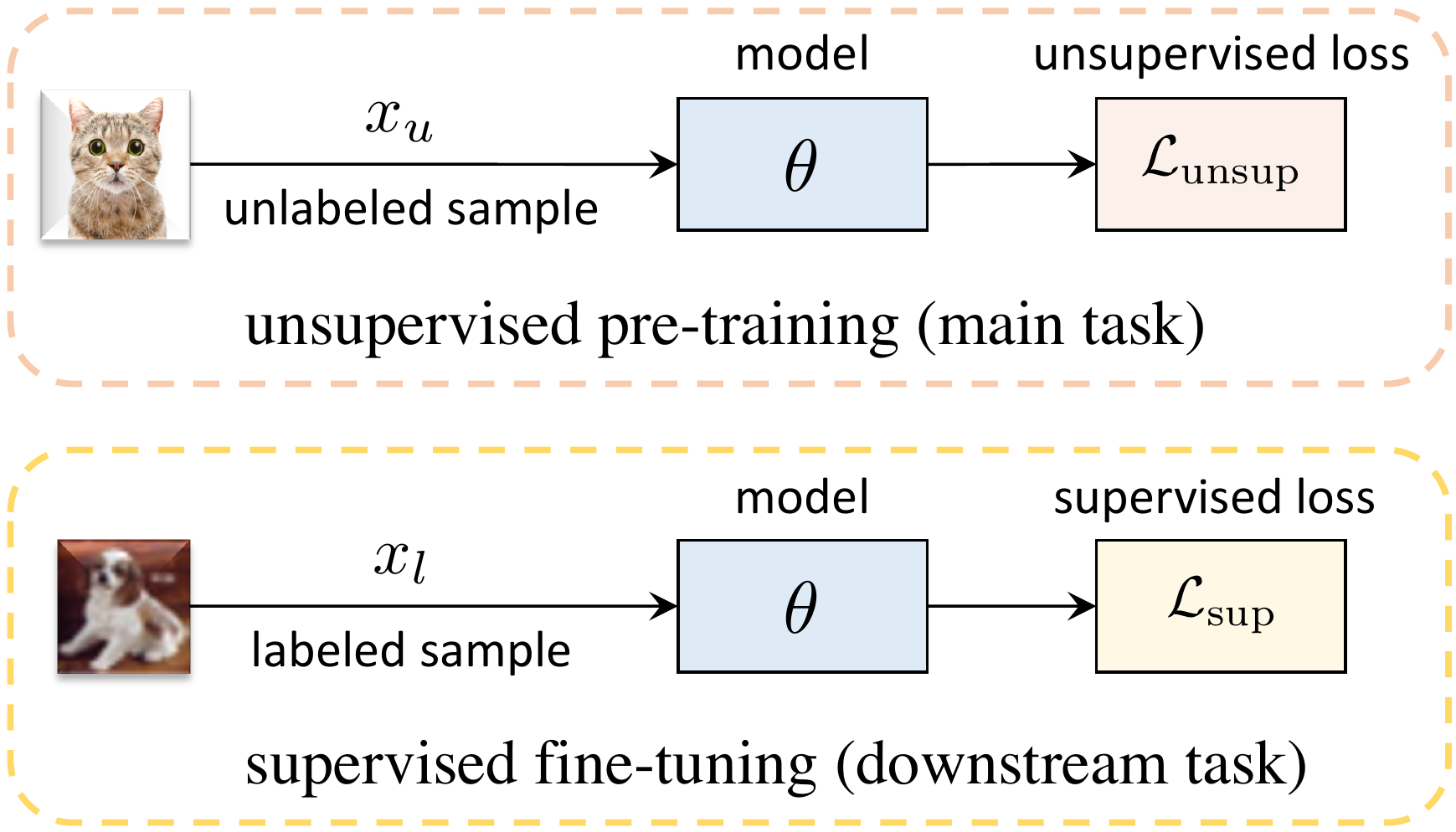}
\caption{Unsupervised learning trains a generalizable model using purely unlabeled data. The model can later be fine-tuned with labeled data and tested on a downstream task.}
\label{fig:ul}
\end{figure}

\myparagraph{Problem Definition.} In UL, we have access to an unlabeled dataset $\mathcal{D}_u = \{\textbf{x}_{i}\}_{i=1}^{N_u}$. As label information is unknown, the UL loss function $\mathcal{L}$ for training a DNN $\theta$ can generally be expressed as Eq.~\eqref{eq:common}, \ie, $\mathcal{L} = \lambda_l \mathcal{L}_{\text{sup}} + \lambda_u \mathcal{L}_{\text{unsup}}$ with $\lambda_l=0$. In {discriminative} models, the unsupervised objective $\mathcal{L}_{\text{unsup}}$ requires certain {pseudo/proxy targets} to learn semantically meaningful and generalizable representations. In {generative models}, $\mathcal{L}_{\text{unsup}}$ is imposed to explicitly model the {data distribution}. See Figure \ref{fig:ul} for an illustration of UL. 

\myparagraph{Evaluation Protocol.} The performance of UL methods are often evaluated via  two protocols, commonly known as the (1) linear classification protocol, and (2) fine-tuning on downstream tasks. In (1), the pre-trained DNN is frozen to extract the features for an image dataset, while a linear classifier (\eg, a fully-connected layer or a kNN classifier) is trained to classify the extracted features. In (2), the pre-trained DNN is used to initialize a model for any downstream task, followed by fine-tuning with a task-specific objective, such as fine-tuning an object detector initialized from an unsupervised pre-trained backbone (\eg, FasterR-CNN~\cite{ren2016faster}) on object detection datasets (\eg, PASCAL VOC~\cite{everingham2010pascal}), or fine-tuning a segmentation model (\eg, Mask R-CNN~\cite{he2017mask}) with a pre-trained backbone on segmentation datasets (\eg, COCO~\cite{lin2014microsoft}). 

\subsection{Taxonomy on UL Algorithms}
\label{sec:taxonomyul}

Existing unsupervised deep learning models can be mainly grouped into three families: {pretext tasks}, {discriminative models} and {generative models} (Table \ref{tab:ul}). Pretext tasks and discriminative models are also known as {self-supervised learning}, which drive model learning by a proxy protocol/task and construct pseudo label supervision to formulate unsupervised surrogate losses. Generative models is inherently unsupervised and explicitly models the data distribution to learn representations without label supervision. We review these models in \S \ref{sec:pretext}, \S \ref{sec:dis} and \S \ref{sec:ul-dgm}. 

\subsubsection{Pretext Tasks}
\label{sec:pretext}

\begin{figure}[!t]
\centering
\includegraphics[width=0.45\textwidth]{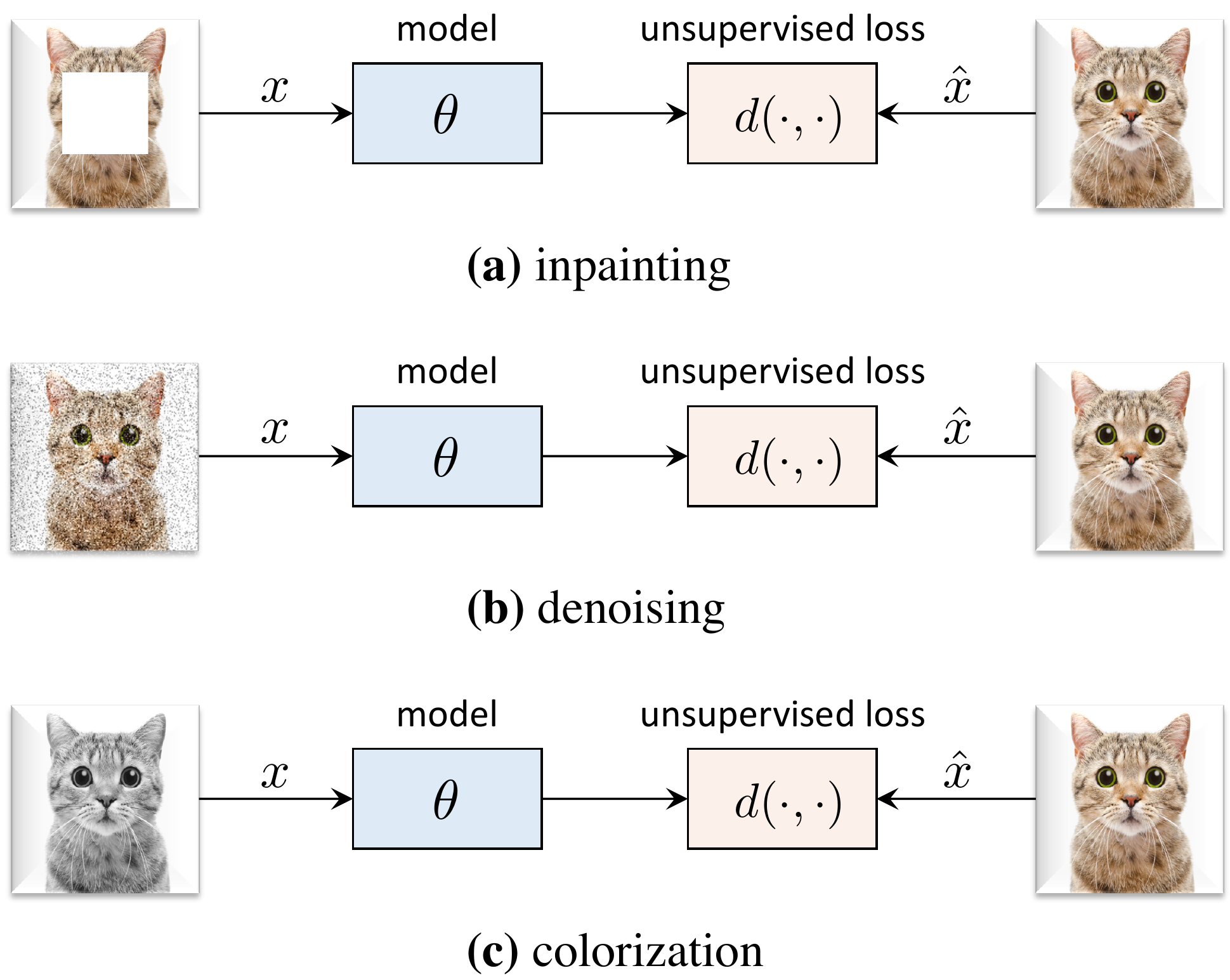}
\caption{In {pixel-level} pretext tasks (\S \ref{sec:pretext}), 
the aim is to reconstruct the original image $\hat x$ from a corrupted input $x$.}
\label{fig:ul-pixel-pretext}
\end{figure}

\begin{figure}[!t]
\centering
\includegraphics[width=0.475\textwidth]{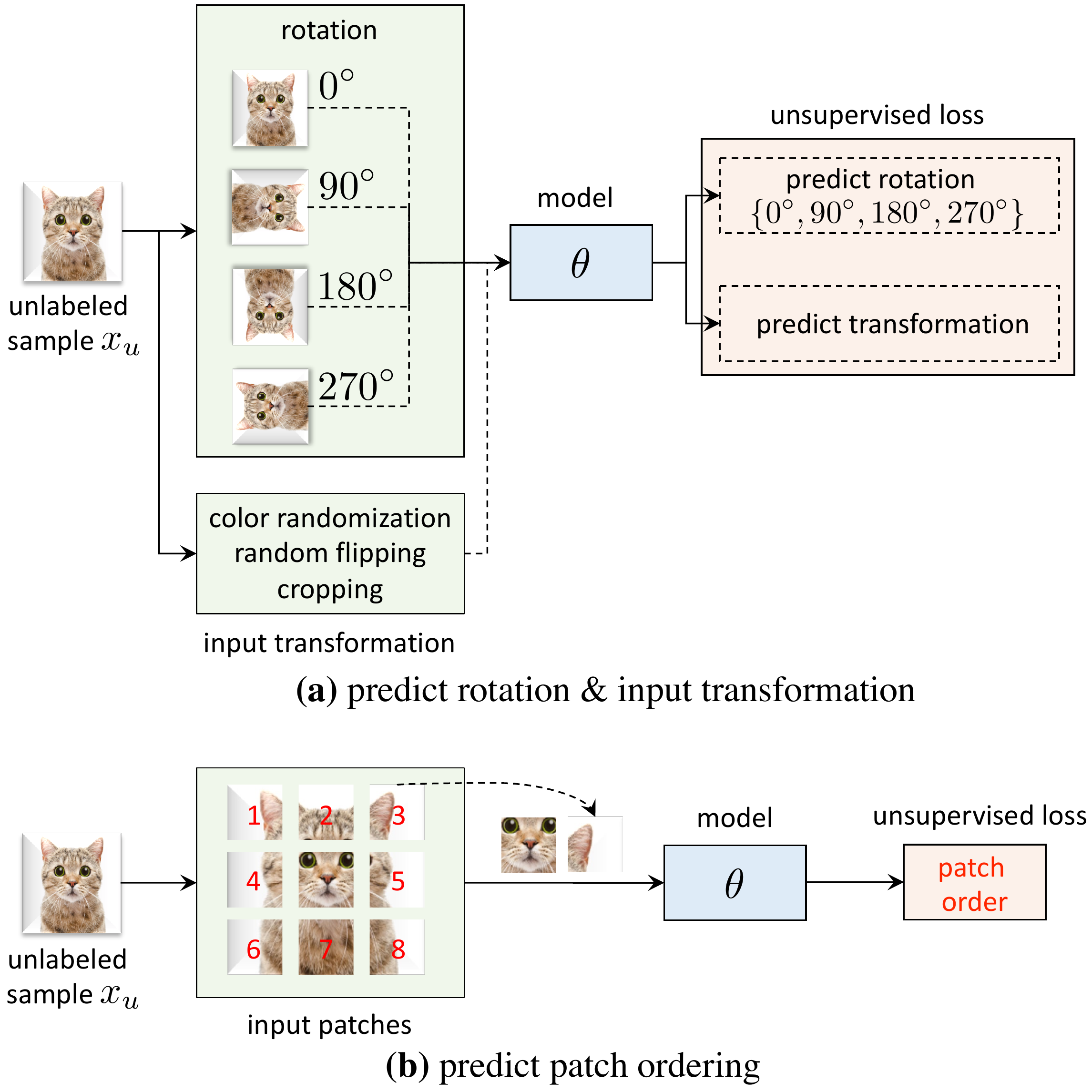}
\caption{In {instance-level} pretext tasks (\S \ref{sec:pretext}) the aim is to predict the transformation on the input.}
\label{fig:self-supervised}
\end{figure}

\newparagraph{Pretext Tasks} refer to hand-crafted proxy tasks manually designed to predict certain task-agnostic properties of the input data, which do not require any label supervision for training. By formulating self-supervised learning objectives with free labels, meaningful visual representations can be learned in a fully unsupervised manner. In the following, we review two classes of pretext tasks, which introduce the self-supervision signals at the {pixel-level} (illustrated in Figure \ref{fig:ul-pixel-pretext}) or {instance-level} (illustrated in Figure \ref{fig:self-supervised}). 

\myparagraph{Pixel-level} pretext task is generally designed as a dense prediction task that aims to predict the expected pixel values of an output image as a self-supervision signal~\cite{hinton2006reducing,vincent2008extracting,masci2011stacked,pathak2016context,zhang2016colorful,zhang2017split,larsson2017colorization,he2022masked}. Auto-Encoder~\cite{hinton2006reducing,masci2011stacked} is one of the most representative and primitive unsupervised models that learn representations by reconstructing input images. In addition to standard reconstruction, pixel-level pretext tasks introduce more advanced image generation tasks to hallucinate the pixel colour values of the corrupted input images, as represented by three standard low-level image processing tasks: (1) {image inpainting}~\cite{pathak2016context,he2022masked} learns by inpainting the masked-out missing regions in the input images, which is also known as masked auto-encoders (MAE) \cite{he2022masked}; (2) {denoising}~\cite{vincent2008extracting} learns to denoise the partial destructed input; and  (3) {colorization}~\cite{zhang2016colorful,zhang2017split,larsson2017colorization} aims to predict the colour values of the grayscale images. These self-supervised models are trained with an image generation task objective (\eg, a mean square error) to enforce predicting the expected pixel values:
\begin{equation}
\underset{{\theta}}{\text{min}} \ 
\sum_{x \in \mathcal{D}} ||G_{\theta}(x)-\hat{x}||^2, 
\label{eq:l2}
\end{equation}
where $G_{\theta}(\cdot)$ is an image generation network (typically implemented as an encoder-decoder network architecture) trained to predict the expected output image $\hat{x}$ per pixel. Once trained, part of the network $G_{\theta}(\cdot)$ (\eg, encoder) can be used to initialize the model weights or extract the intermediate features for solving the downstream task. 

\myparagraph{Instance-level} pretext tasks introduce sparse semantic labels for each image sample by designing a surrogate proxy task that can be solved per instance without any label annotations \cite{doersch2015unsupervised,noroozi2016unsupervised,noroozi2017representation,santa2018visual,gidaris2018unsupervised,nathan2018improvements,wei2019iterative,goyal2019scaling}. In general, these pretext tasks involve applying different {image transformations} to generate diverse input variations, whereby an artificial supervision signal is imposed to predict the applied transformation on each image instance. Among this line of works, the representative ones consider mainly two classes of instance-wise transformations on input images. The first one is classifying global transformations, such as {rotations}~\cite{gidaris2018unsupervised}, {scaling} and {tiling}~\cite{noroozi2017representation}, where the learning objective is to recognize the geometric transformation applied on an image. The second one is predicting local transformations, such as {patch orderings}~\cite{doersch2015unsupervised} and {patch re-orderings}~\cite{noroozi2016unsupervised,santa2018visual,wei2019iterative}, which cut each image into multiple local patches. The goal of {patch orderings} is to recognize the order of a given cut-out patch, while {patch re-orderings}, also known as the {jigsaw puzzles}, permute the cut-out patches randomly and the goal is to predict the permuted configurations. In summary, the objective of an instance-level pretext task can be written as:  
\begin{equation}
\underset{{\theta}}{\text{min}} 
\ \sum_{{x} \in \mathcal{D}} \mathcal{L}_{\text{unsup}}(\Phi_z({x}),z, \theta), 
\label{eq:pretext}
\end{equation}
where $\mathcal{L}_{\text{unsup}}(\cdot)$ can be various loss functions (\eg, cross-entropy loss~\cite{gidaris2018unsupervised}) that learn a
mapping from a transformed input image $\Phi_z({x})$ to a discrete category or a configuration of the applied transformation $z$. Once trained, the representations are covariant with the transformations $\Phi_z(\cdot)$, thus being aware of the 
spatial context information, \eg, how an image is rotated or how the local patches are permuted. 

\myparagraph{\underline{Remarks.}} Although self-supervised learning objectives of pixel-level or instance-level {pretext tasks} are generally not explicitly related to the downstream task objectives (\eg, image classification, detection and segmentation), they permit to learn from unlabeled data by predicting the spatial context or structured correlation in images, such as inpainting missing regions, and predicting the applied rotations. As these self-supervision signals can implicitly uncover the semantic content (e.g. human interpretable concepts~\cite{goh2021multimodal}) or spatial context in images, they often yield a meaningful pre-trained model for initialization in unseen downstream tasks, or even serve as a flexible and effective regularizer to facilitate other machine learning setups, such as semi-supervised learning~\cite{zhai2019s4l} and domain generalization~\cite{carlucci2019domain}. 

\subsubsection{Discriminative Models}
\label{sec:dis}

\newparagraph{Discriminative models} hereby refer to the class of unsupervised discriminative models that learn visual representations from the unlabeled data by enforcing invariance towards various task-irrelevant visual variations at either instance-level, neighbor-level or group-level. These visual variations can be {intra-instance variations} such as different views of the same instance~\cite{hjelm2018learning,tian2019contrastive,bachman2019learning,tschannen2019mutual,tian2020makes}, or {inter-instance variations} between neighbor instances~\cite{huang2019unsupervised,van2020learning} or across a group of instances~\cite{caron2018deep,caron2019unsupervised,caron2020unsupervised}. 

\begin{figure}[!t]
\centering
\includegraphics[width=0.475\textwidth]{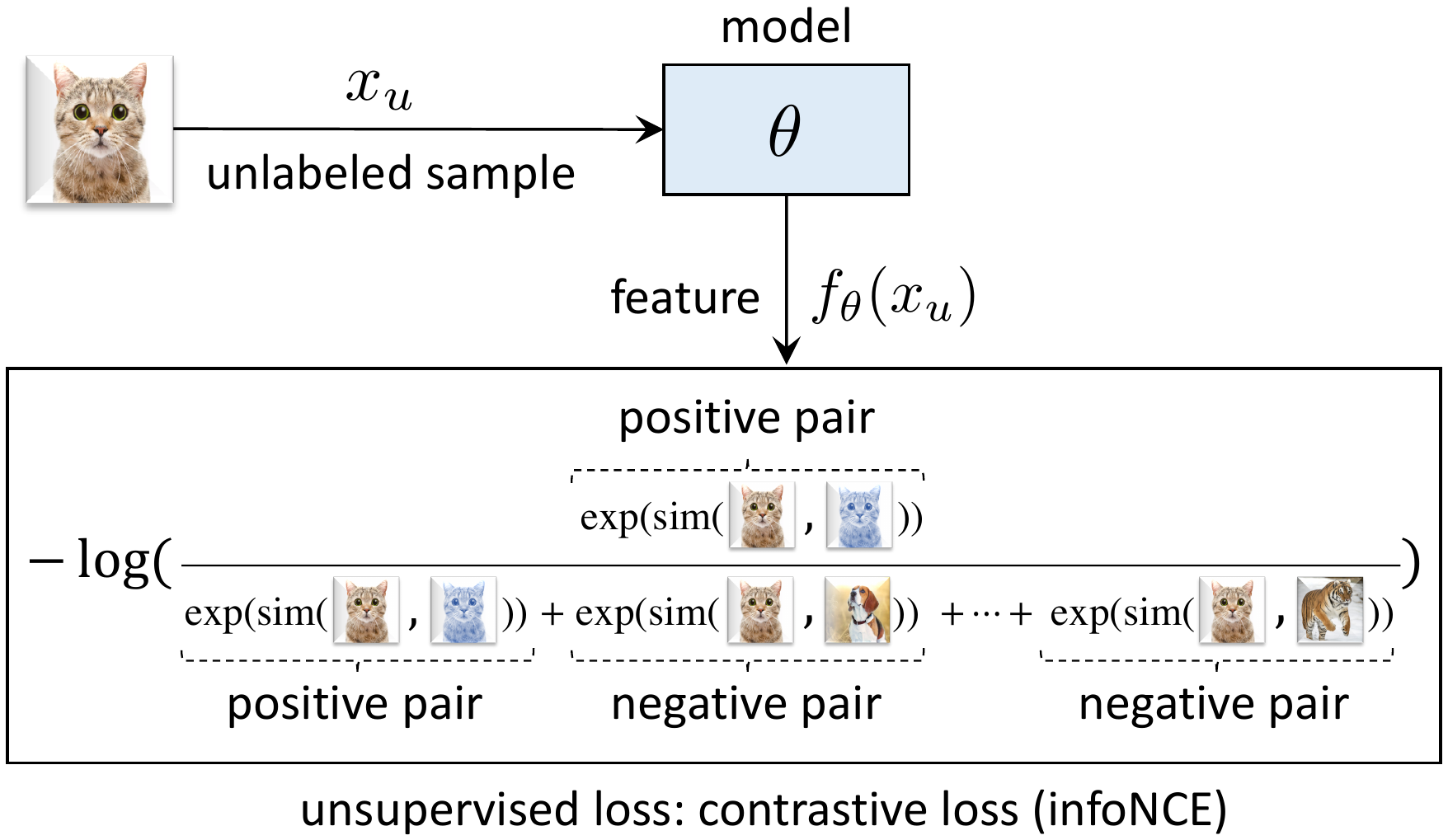}
\caption{The unsupervised discriminative model using contrastive learning (\S \ref{sec:dis}) aims to pull together the positive pairs and push away the negative ones.}
\label{fig:ul-contrastive}
\end{figure}

\begin{figure}[!t]
\centering
\includegraphics[width=0.475\textwidth]{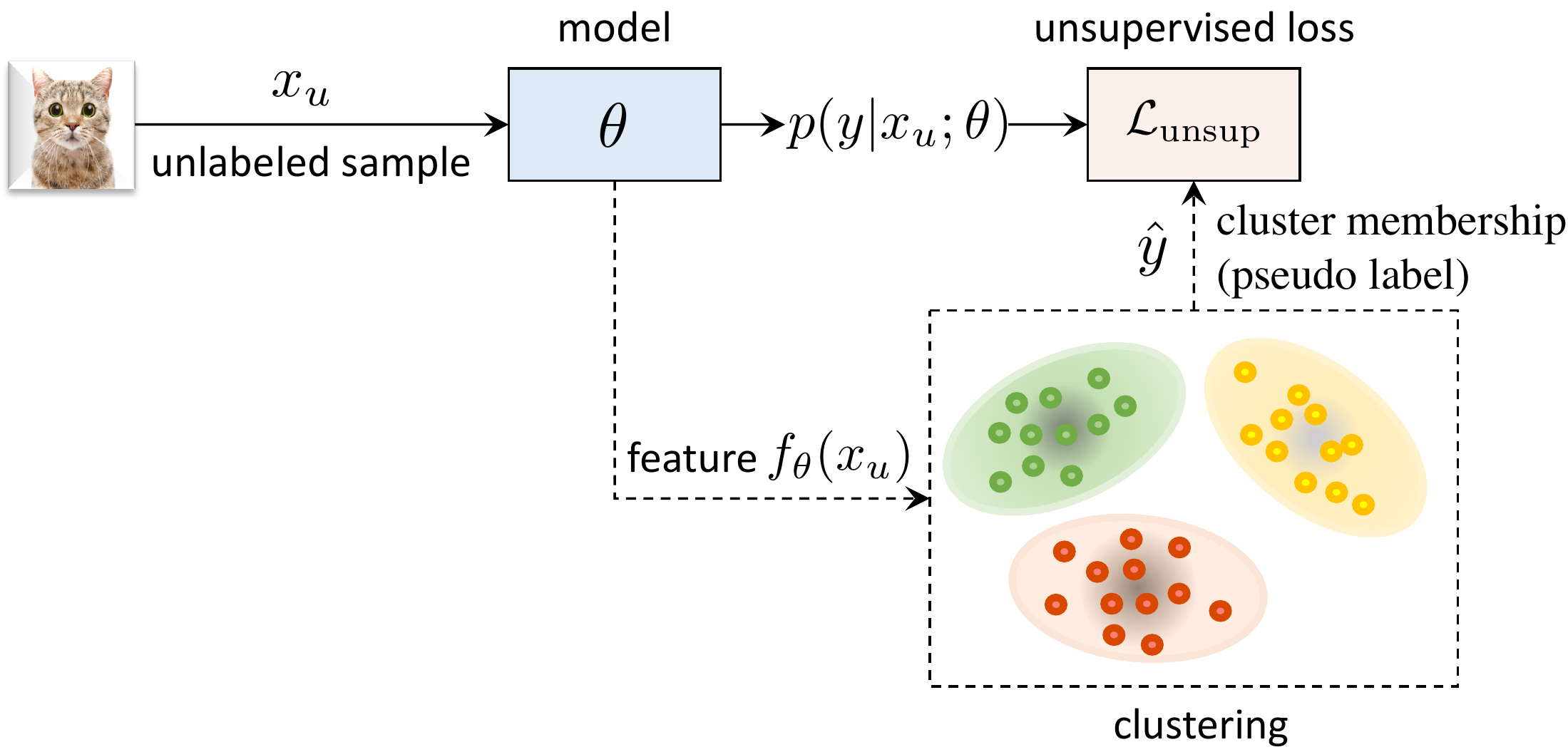}
\caption{In unsupervised discriminative models using deep clustering (\S \ref{sec:dis}), unlabeled samples are assigned to a set of clusters by {\em online} or {\em offline} clustering, while the cluster memberships are utilized as pseudo labels for training.} 
\label{fig:ul-clustering}
\end{figure}

In the following, we review two representative classes of unsupervised discriminative models that offer the state of the art in unsupervised visual feature learning, including instance discrimination (see Figure \ref{fig:ul-contrastive}) and deep clustering (see Figure \ref{fig:ul-clustering}). The former imposes self-supervision per instance by treating each instance as a class, while the latter introduces supervision per group by considering a group of similar instances as a class. 

\myparagraph{Instance discrimination} \label{sec:instance} models learn discriminative representations by enforcing invariance towards different viewing conditions, data augmentations or various parts of the same image instance~\cite{dosovitskiy2014discriminative,dosovitskiy2015discriminative,novotny2018self,hjelm2018learning,wu2018unsupervised,tschannen2019mutual,bachman2019learning,ye2019unsupervised,misra2020self,he2019momentum,chen2020simple,tian2019contrastive,tian2020makes,chen2020improved,chen2020big} -- also known as {exemplar learning}~\cite{dosovitskiy2014discriminative,dosovitskiy2015discriminative}. 

The most prevalent scheme in instance discrimination is {\bf contrastive learning}, which was initially proposed to learn invariant representations by mapping {similar} inputs to {nearby} points in the latent space~\cite{hadsell2006dimensionality,bromley1993signature}. The state-of-the-art contrastive learning models for self-supervised learning generally aim to obtain an invariance property by optimizing a contrastive loss formulated upon the noise contrastive estimation (NCE) principle~\cite{gutmann2010noise}, which maximizes the mutual information across different views. The multi-view information bottleneck model~\cite{FDFKA20} extends the original information bottleneck principle to unsupervised learning and trains an encoder to retain all the relevant information for predicting the label while minimizing the excess information in the representation. Formally, contrastive learners such as SimLR~\cite{chen2020simple} and MoCo~\cite{he2019momentum} are generally optimized by an instance-wise contrastive loss (\ie, infoNCE loss)~\cite{hadsell2006dimensionality,oord2018representation}:
\begin{equation}
\underset{{\theta}}{\text{min}} 
\ \sum_{{x_i} \in \mathcal{D}} 
- \text{log} \ \frac{\text{exp}(f_{\theta}(x_{i}) \cdot f_{\theta}({x}^+_{i}) / \tau)}
{\sum_{j=1}^{M}\text{exp}(f_{\theta}(x_{i}) \cdot f_{\theta}({x}_{j}) /\tau)}, 
\label{eq:infonce}
\end{equation}
where $\tau$ is a temperature, $f_{\theta}$ is the feature encoder, \ie, a DNN; $f_{\theta}(x_{i}), f_{\theta}({x}^+_{i})$ are the feature embeddings of two different augmentations, or views of the same image; $\{x_j\}_{j=1}^M$ includes $(M{-}1)$ negative samples and $1$ positive (\ie, $x_i^+$) sample. Eq.~\eqref{eq:infonce} optimizes the network by enforcing the positive pairs (\ie, embeddings of the same instance) to lie closer, while pushing apart the negative pairs (\ie, embeddings of different instances). Minimizing the InfoNCE loss is equivalent to maximizing a lower bound on the mutual information between $f_{\theta}(x_{i})$ and $f_{\theta}({x}^+_{i})$~\cite{hjelm2018learning}. 

To derive a tractable yet meaningful contrastive distribution in Eq.~\eqref{eq:infonce}, a large amount of negative pairs are often required per training batch. To this aim, existing state-of-the-art methods are typically featured with different {negative sampling} strategies to collect more negative pairs. For instance, a large batch size of 4096 is adopted in SimCLR~\cite{chen2020simple}. In InstDis~\cite{wu2018unsupervised}, MoCo~\cite{he2019momentum}, PIRL~\cite{misra2020self}, and CMC~\cite{tian2019contrastive}, a memory bank is used to maintain all the instance prototypes by keeping moving average of their feature representations over training iterations. Finally, {running queue} enqueues the features of samples in the latest batches and dequeues the old mini-batches of samples to store a fraction of sample's features from the preceding mini-batches~\cite{he2019momentum,chen2020improved,misra2020self}. 

\label{para:new}
Inspired by deep metric learning, various training strategies further boost contrastive learning. For instance, a hard negative sampling strategy \cite{robinson2020contrastive} mines the negative pairs that are similar to the samples but likely belong to different classes. To train negative pairs and (or) positive pairs by adversarial training \cite{hu2021adco,wang2022caco} learn a set of ``adversarial negatives'' confused with the given samples, or ``cooperative positives'' similar to the given samples. These strategies are designed to find the better negative and positive pairs for improving contrastive learning. 

In addition to {negative sampling}, it is essential to apply various {image transformations} for generating multiple diverse variants (\ie, views) of the same instance to construct the positive pairs. The most typical way is to apply common {data augmentation} such as random cropping and color jittering~\cite{wu2018unsupervised,bachman2019learning,ye2019unsupervised,he2019momentum,misra2020self,chen2020simple}, or {pretext transformation}~\cite{misra2020self} like patch re-ordering~\cite{noroozi2016unsupervised} and rotation~\cite{gidaris2018unsupervised}. An alternative way is to artificially construct {multiple views} of a single image by using different image channels like luminance and chrominance~\cite{tian2019contrastive}, or by extracting the local and global patches of the same image~\cite{hjelm2018learning}. In a nutshell, although there are different strategies in negative sampling and image transformations to construct the negative and positive pairs for contrastive learning, these strategies share the same aim to learn visual representations invariant to diverse input transformations~\cite{novotny2018self,misra2020self}. 

\label{sec:newul}
While {\bf contrastive learning} approaches rely on obtaining a sufficient amount of negative pairs to derive the contrastive loss (Eq.~\eqref{eq:infonce}), another alternative {\bf non-contrastive} scheme for instance discrimintation operates in a {\bf negative-sample-free} manner~\cite{grill2020bootstrap,chen2020exploring,tian2021understanding,zbontar2021barlow},  as exemplified by bootstrap (BYOL)~\cite{grill2020bootstrap} and simple siamese networks (SimSiam)~\cite{chen2020exploring}). In particular, in BYOL and SimSiam, two views (obtained from data augmentation) of the same images are passed towards the networks and the mean squared error is minimized between the representations of two views to enforce invariances. Importantly, a stop gradient scheme is adopted to prevent representational collapse, i.e. avoid mapping all the samples to the same representations. Another related method is Barlow Twins~\cite{zbontar2021barlow}, which computes a cross-correlation matrix between the distorted versions of a batch of training samples and enforce the matrix to be an identity matrix, thus learning self-supervised representations invariant to different distortions. Although these non-contrastive methods adopt other loss formulations, they all share the similar spirit as contrastive learning given that meaningful representations are learned by enforcing invariances to different views of the same instance. 

\myparagraph{Deep clustering} \label{sec:clustering} models learn discriminative representations by grouping similar instances from the same cluster together~\cite{xie2016unsupervised,yang2016joint,hu2017learning,caron2018deep,zhuang2019local,caron2019unsupervised,asano2019self,huang2020deep,yan2020clusterfit,wang2020unsupervised,caron2020unsupervised,gidaris2020learning,van2020learning}. In training, the entire dataset is generally divided into groups by associating each instance to a certain cluster centroid based on pairwise similarities. Although clustering algorithms are longstanding machine learning techniques ~\cite{jain1999data,coates2012learning,von2007tutorial}, they have been re-designed to be seamlessly integrated with DNNs to learn discriminative representations without label supervision. Conceptually, the cluster memberships can be considered as some pseudo labels to supervise the model training, as written in Eq.~\eqref{eq:clustering}. 
\begin{equation}
\underset{{\theta}}{\text{min}} 
\ \sum_{x \in \mathcal{D}} \mathcal{L}_{\text{unsup}}(x,\hat{y}, \theta),
\label{eq:clustering}
\end{equation}
where $\hat{y}$ is the cluster membership of sample $x$, $\mathcal{L}_{\text{unsup}}(\cdot, \cdot, \theta)$ is the loss function that constrains the mapping from $x$ to $y$, such as a classification loss. Deep clustering algorithms can be further grouped into two categories according to whether the assignments of cluster memberships are derived in an {offline} or {online} manner, as detailed in the following.

In {offline clustering}, unsupervised training is alternated between a {cluster assignment} step and a {network training} step~\cite{xie2016unsupervised,yang2016joint,chang2017deep,guo2017improved,yang2017towards,asano2019self,caron2019unsupervised,van2020learning}. While the former step estimates the cluster memberships of all the training samples, the latter uses the assigned cluster memberships as pseudo labels to train the network. Representative offline clustering models include DeepCluster~\cite{caron2018deep}, JULE~\cite{yang2016joint} and SeLa~\cite{asano2019self}, which mainly differ in the clustering algorithms. Specifically, {DeepCluster}~\cite{caron2018deep,caron2019unsupervised} groups visual features using k-means clustering~\cite{coates2012learning}. JULE~\cite{yang2016joint} uses agglomerative clustering~\cite{gowda1978agglomerative} that merges similar clusters to iteratively derive new cluster memberships. SeLa~\cite{asano2019self} casts clustering as an optimal transport problem solved by Sinkhorn-Knopp algorithm~\cite{cuturi2013sinkhorn} to obtain the cluster memberships as pseudo labels. 

In {online clustering}, the {cluster assignment} step and {network training} step are coupled in an end-to-end training framework, as represented by IIC~\cite{ji2019invariant}, AssociativeCluster~\cite{haeusser2018associative}, PICA~\cite{huang2020deep}, and SwAV~\cite{caron2020unsupervised}. Compared to offline clustering, online clustering could better scale to large-scale datasets, as it does not require clustering the entire dataset iteratively. This is typically achieved in two ways: (1) training a classifier that parameterizes the cluster memberships (\eg, IIC and PICA); (2) learning a set of cluster centroids/prototypes (\eg, AssociativeCluster and SwAV). For instance, IIC~\cite{ji2019invariant} learns the cluster memberships by maximizing the mutual information between predictions of an original instance and a randomly perturbed instance obtained from data augmentation. SwAV~\cite{caron2020unsupervised} learns a set of prototypes (\ie, cluster centroids) in the feature space and assigns each sample to the closest prototype. 

\myparagraph{\underline{Remarks.}} Recent advances of discriminative unsupervised models include both contrastive learning and deep clustering, which have set the new state of the art. On one side, contrastive learning discriminates individual instances by imposing transformation invariance at the {instance-level}. Interestingly, this opposes some instance-level pretext tasks that instead learn by predicting the applied transformations. Contrastive learning also closely relates to {consistency regularization} in SSL in the sense of enforcing invariance to transformations, although different loss functions are often used. However, as shown in~\cite{chen2020exploring}, a pairwise loss objective -- often used for consistency regularization in SSL -- can be also effective as contrastive loss (Eq.~\eqref{eq:infonce}). This suggests that the essential idea behind them is {identical} -- {imposing transformation invariance at instance level}. Deep clustering, on the other hand, discriminates between groups of instances for discovering the underlying semantic boundaries, and enforces {group-level} invariance. The idea of {consistency regularization} is also adopted by several deep clustering methods~\cite{huang2020deep,ji2019invariant}, conforming its more generic efficacy beyond SSL. Lastly, discriminative unsupervised learning can also be conducted at both instance-level and group-level to learn more powerful representations \cite{wang2020unsupervised,li2020contrastive}.

\begin{figure}[!t]
\centering
\includegraphics[width=0.475\textwidth]{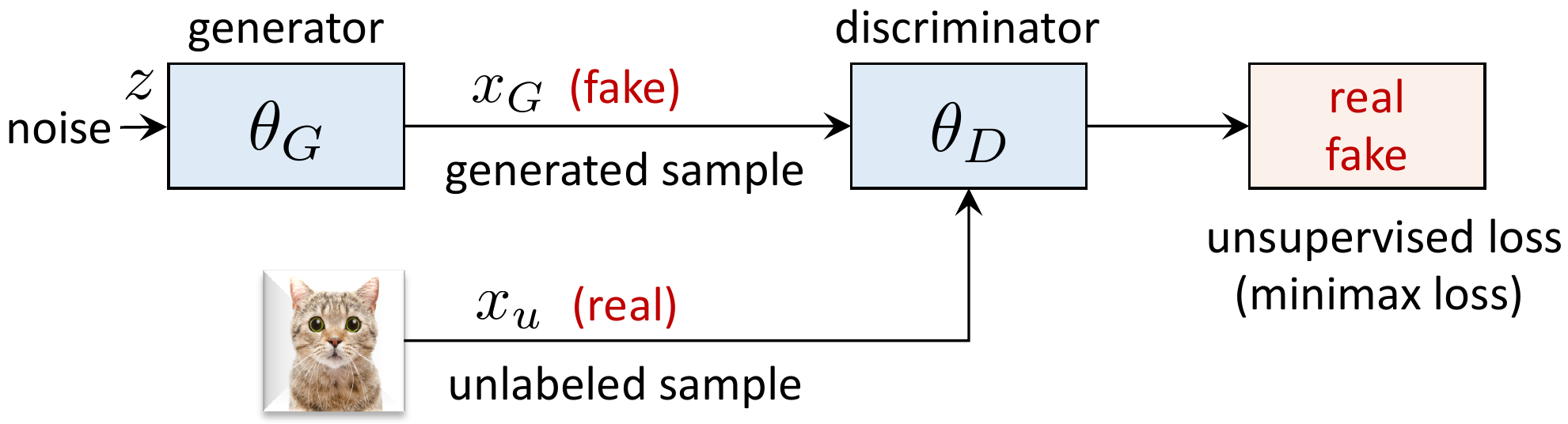}
\caption{In GANs (\S \ref{sec:ul-dgm}), 
a generator and a discriminator are trained with a minimax game (Eq.~\eqref{eq:minmax}) in an unsupervised manner, 
whilst their intermediate features lead to discriminative visual representations.}
\label{fig:ul-dgm}
\end{figure}

\subsubsection{Deep Generative Models}
\label{sec:ul-dgm}

\newparagraph{Deep generative models} (DGMs), as introduced in \S \ref{sec:DGMs}, are inherent unsupervised learners that explicitly model the data distribution \cite{hinton2006fast,kingma2013auto,goodfellow2014generative,gabbay2019demystifying}. 
DGMs are applicable for both semi-supervised and unsupervised learning. A typical Generative Adversarial Network (GAN) \cite{springenberg2015unsupervised,donahue2016adversarial,donahue2019large,chen2019self,wang2020transformation} contains a {discriminator} $D$ to differentiate real and fake samples, and a {generator} $G$ that can serve as an image encoder to capture the semantics in latent space, as trained by a min-max game:
\begin{align}
\underset{G}{\text{min}} \ \underset{D}{\text{max}} \ 
\mathds{E}_{x\sim p_{\text{data}}(x)}[\text{log} D(x)] {+}
\mathds{E}_{z\sim p_{z}(z)}[\text{log} (1{-}D(G(z)))],
\label{eq:minmax}
\end{align}
where $z$ is sampled from an input noise distribution $p_{z}(z)$. GANs can learn representations at both the {discriminator} and the {generator} level. See Figure \ref{fig:ul-dgm} for an illustration of deep generative model based on a GAN.  

To learn representations at the {\bf discriminator-level}, Deep Convolutional Generative Adversarial Network ({DCGAN}) \cite{radford2015unsupervised} adopts a pre-trained convolutional discriminator to extract features for tackling a downstream image classification task. Later on, Self-supervised GAN~\cite{chen2019self} and Transformation GAN~\cite{wang2020transformation} further imbue the discriminator with a self-supervised pretext task to predict the applied image transformation, thus enabling the representations to capture the latent visual structures. 

To learn representations at the {\bf generator-level}, Bidirectional Generative Adversarial Networks ({BiGAN}) \cite{donahue2016adversarial} introduces an image encoder coupled with the generator, which is trained with a {joint discriminator loss} to tie the data distribution and the latent feature distribution together. This allows the image encoder to capture the semantic variations in its latent representation, and offer discriminative visual representations for {one nearest neighbor} (1NN) classification. To further improve BiGAN, {BigBiGAN}~\cite{donahue2019large} adopts more powerful discriminator and generator architectures than BigGAN~\cite{brock2018large}, together with an additional {unary discriminator loss} to constrain the data or latent distribution independently, therefore enabling more expressive unsupervised representation learning at the {generator-level}.  

\myparagraph{\underline{Remarks.}} Although most state-of-the-art UL methods are self-supervised models that solve {pretext tasks} or perform {unsupervised discriminative learning} (as reviewed in \S \ref{sec:pretext} and \S \ref{sec:dis}), {deep generative models} are still an important class of unsupervised learners owing to their native unsupervised nature to learn expressive data representations in a probabilistic manner. Further, they do not require manual design of a meaningful discriminative learning objective, while offering a unique ability to generate abundant data. 

\section{Discussion on SSL and UL}
\label{sec:sslul}

In this section, we connect SSL and UL via further discussion on their common learning assumptions (\S \ref{sec:assumption}), and their applications 
in different computer vision tasks (\S \ref{sec:vision}).

\subsection{The learning assumptions shared by SSL and UL} \label{sec:assumption}

\begin{figure}[!t]
\centering
\includegraphics[width=0.42\textwidth]{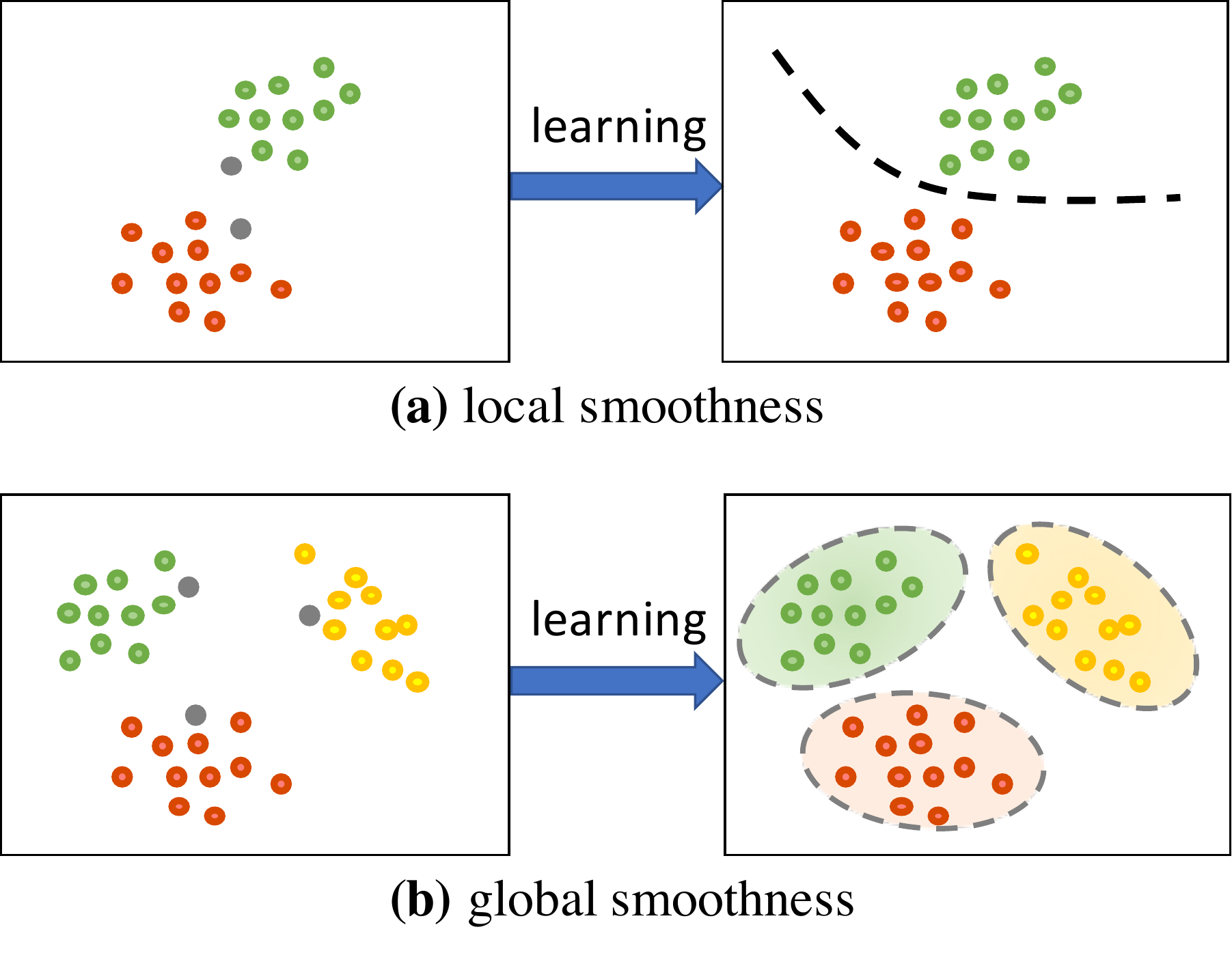}
\caption{SSL and UL share {\bf (a)} local and {\bf (b)} global smoothness assumptions. Unlabeled samples (grey dots) are assigned to class labels depending on the decision boundaries derived from the local or global smoothness assumptions.}
\label{fig:assump}
\end{figure}

As discussed in \S \ref{sec:overviewssl}, the unsupervised learning objectives in SSL are often formulated based on the smoothness assumption~\cite{zhou2004learning}. Broadly speaking, the learning assumptions of various discriminative SSL and UL algorithms can be grouped into two types of smoothness assumptions, i.e. local smoothness and global smoothness -- as visually illustrated in Figure \ref{fig:assump}. In the following, we further elaborate these assumptions and discuss the different SSL and UL algorithms that are built upon these assumptions. 

\subsubsection{Local Smoothness}\label{sec:localsm}
There are two flavors of {\bf local smoothness} assumption. First, a sample $x_i$ is assumed to share the same class label as its transformed variant $\hat{x}_i$ (Eq.~\eqref{eq:sm1}). Second, a sample $x_i$ is assumed to belong to the same class as its nearby sample $x_j$ in the latent representation space (Eq.~\eqref{eq:sm2}). An unsupervised loss term enforces local smoothness on an unlabeled sample $x_i$ via:
\begin{equation}
\underset{{\theta}}{\text{min}} 
\ \sum_{{x_i} \in \mathcal{D}} \mathcal{L}_{\text{unsup}}(f({x_i}),f(\hat{x}_i)) 
\label{eq:sm1}
\end{equation}
\begin{equation}
\underset{{\theta}}{\text{min}} 
\ \sum_{{x_i} \in \mathcal{D}} \mathcal{L}_{\text{unsup}}(f({x_i}),f(x_j))
\label{eq:sm2}
\end{equation}
where $f(\cdot)$ is the model to be trained and gives the model output (such as features or predictions). $\mathcal{L}_{\text{unsup}}(\cdot)$ could be any similarity metric that quantifies the divergence or inconsistency between two model outputs, such as a mean square error, or contrastive loss. 

Local smoothness among different views of the same sample (Eq.~\eqref{eq:sm1}) can be achieved via the consistency regularization techniques in SSL (\S \ref{sec:ssl-consistency}, Figure \ref{fig:ssl-consistency}). They enforce predictive smoothness to the same samples under different variations imposed at the input space and (or) model space, given that the different transformed versions of the same sample should lie in its own local neighborhood. Similarly, the instance discrimination algorithms in UL also implicitly enforce the same samples under different views or transformations to have locally consistent representations, as represented by contrastive learning which encourages local invariances on each sample (\S \ref{sec:instance}, Figure \ref{fig:ul-contrastive}). 

Local smoothness among the nearby samples (Eq.~\eqref{eq:sm2}) can be imposed via the graph-based regularization techniques in SSL. They often propagate the class labels to the unlabeled samples using the labels of their neighbours on the graph, as the nearby samples should likely share the same class (\S \ref{sec:ssl-graph}, Figure \ref{fig:ssl-graph}). Similarly, neighbourhood consistency is also explored in UL~\cite{huang2019unsupervised,van2020learning}, which forms the semantic training labels by mining the nearest neighbors of each sample based on feature similarity, given that nearest neighbors are likely to belong to the same semantic class. 

\subsubsection{Global Smoothness}\label{sec:globalsm} The {\bf global smoothness} assumption indicates that a sample $x_i$ could be assigned to a certain class (or target) $z_i$ based on the underlying global structures captured by the model: 
\begin{equation}
\underset{{\theta}}{\text{min}} 
\ \sum_{{x_i} \in \mathcal{D}} \mathcal{L}_{\text{unsup}}(f({x_i}),z_i)
\label{eq:sm3}
\end{equation}
where $z_i$ is the learning target (e.g. the cluster membership or the most confident predicted class), which is derived from the global class decision boundaries discovered during training (Figure \ref{fig:assump}) whilst the decision boundaries are supposed to lie in low density regions. 
Similar to Eq.~\eqref{eq:sm1} and Eq.~\eqref{eq:sm2}, $\mathcal{L}_{\text{unsup}}(\cdot)$ is a similarity metric that quantifies the inconsistency between the model output and the training target, such as a cross-entropy loss. The global smoothness assumption is also widely adopted in various SSL and UL techniques to learn from the unlabeled samples with pseudo learning targets, as detailed in the following. 

The self-training techniques in SSL (\S \ref{sec:self-train}, Figure \ref{fig:ssl-self-training}) are generally formulated based on global smoothness, as the learning targets for unlabeled data are derived based on the class decision boundaries discovered by the models. For instance, in entropy minimization (Eq.~\eqref{eq:ent}, Figure \ref{fig:ssl-self-training} (a)), the pseudo label is obtained as the class predicted with the highest confidence. In co-training and distillation (Eq.~\eqref{eq:cot}, Eq.~\eqref{eq:distill}, Figure \ref{fig:ssl-self-training} (b)(c)), the learning targets come from the model co-trained in parallel or pre-trained beforehand. 
Similarly, the deep clustering algorithms in UL (\S \ref{sec:clustering}, Figure \ref{fig:ul-clustering}) are also proposed upon global smoothness, given that the cluster memberships for unlabeled samples are acquired from an online or offline clustering algorithm which uncovers the latent class decision boundaries in the feature space.

\subsubsection{Connections between SSL and UL} \label{sec:connect}

{
\setlength{\tabcolsep}{1pt}
\renewcommand{\arraystretch}{1.5}
\begin{table}[!t]
\fontsize{8pt}{8pt}\selectfont
\centering
\caption{A common taxonomy on SSL and UL methods based on their learning assumptions.}
\label{tab:sm}
\begin{tabular}{l|c|l}
\hline
\bf Assumption & \bf Objective & \bf Corresponding SSL \& UL methods \\ \hline
\multirow{4}{*}{\bf local smoothness} 
    & \multirow{2}{*}{Eq.~\eqref{eq:sm1}} 
    & consistency regularization in SSL (\S \ref{sec:ssl-consistency}) \\
    & & instance discrimination in UL (\S \ref{sec:instance}) \\
    \cline{2-3}
    & \multirow{2}{*}{Eq.~\eqref{eq:sm2}} 
    & graph-based regularization in SSL (\S \ref{sec:ssl-graph}) \\
    & & neighbourhood consistency in UL (\S \ref{sec:dis}) \\
\hline
\multirow{2}{*}{\bf global smoothness} 
    & \multirow{2}{*}{Eq.~\eqref{eq:sm3}} 
    & self-training in SSL (\S \ref{sec:self-train}) \\
    & & deep clustering in UL (\S \ref{sec:clustering}) \\
\hline
\end{tabular}
\end{table}
}

\myparagraph{The learning rationales common in SSL and UL.} As analyzed in \S \ref{sec:localsm} and \S \ref{sec:globalsm}, most SSL and UL algorithms are formulated based on the same local smoothness or global smoothness assumption -- as summarized in Table \ref{tab:sm}. A common aspect of these SSL and UL algorithms is to design visual learning objectives that enforce invariance or equivariance towards different transformations applied on the input data, as represented by consistency regularization in SSL (\S \ref{sec:ssl-consistency}) and instance discrimination in UL (\S \ref{sec:instance}). Typical transformation strategies can range from simple data augmentation \cite{sajjadi2016regularization,laine2016temporal,tarvainen2017mean}, to more complex transformations such as adversarial perturbations \cite{miyato2015distributional,miyato2018virtual,najafi2019robustness,suzuki2020adversarial}, rotations \cite{gidaris2018unsupervised} and patch reordering \cite{noroozi2016unsupervised}, autoencoding transformations \cite{qi2019avt,zhang2019aet} and automated augmentation \cite{berthelot2019remixmatch,xie2020unsupervised,sohn2020fixmatch}. On one side, most of these SSL and UL methods hinge on learning representations invariant to data augmentation and perturbations by assigning the same underlying labels to the augmented and perturbed data samples. On the other side, other SSL and UL methods consider learning representations that are equivalent to different transformations such as rotations and patch re-ordering by learning to predict the type of transformations.

Many state-of-the-art SSL and UL methods can be well related with the same underlying learning assumptions, given that they introduce similar objectives to learn from the unlabeled samples.  In essence, the learning rationales of these SSL and UL methods could be broadly categorized as: (1) impose the consistency among different transformed versions of the same sample (Eq.~\eqref{eq:sm1}), (2) enforce the smoothness between a sample and its neighbouring one (Eq.~\eqref{eq:sm2}), and (3) derive learning targets for the unlabeled samples based on global decision boundaries (Eq.~\eqref{eq:sm3}). 

\myparagraph{The similarities and differences between problem setups.} In the problems setups, SSL and UL are similar in the sense that both labeled and unlabeled data are often involved in their training protocols before evaluating their generalized model performance on the test set. In particular, the SSL paradigm adopts {\bf \em one-stage} training and uses both labeled and unlabeled data during training (Figure \ref{fig:ssl}); while most existing UL protocols consider {\bf \em two-stage} training (Figure \ref{fig:ul}) -- one stage for {\bf \em pre-training} with unlabeled data and another stage for {\bf \em fine-tuning} with labeled data on a {\em downstream task}. 

In brief, when it comes to training protocols, UL differs from SSL in several ways: (1) the labeled data and unlabeled data are not given together at once; (2) unlabeled and labeled datasets may have different distributions. These properties make UL a more generic learning paradigm to leverage different unlabeled datasets. Nevertheless, how unsupervised pre-training upon different forms of unlabeled data benefits the model generalization on specific {downstream tasks} remains an open research question. For instance, it remains unclear how an unsupervised model pre-trained on natural colour images could generalize to a downstream task that has a different data distribution such as grayscale images in medical imaging. In this regard, SSL provides a more reliable learning paradigm to utilize the unlabeled data, given that the label set offers the prior knowledge for the models and (or) the model designers to select the useful set of unlabeled samples that are similar to the labeled data distribution.

\subsection{Applied SSL and UL in Visual Recognition} \label{sec:vision}

In \S \ref{sec:ssl} and \S \ref{sec:ul}, we mainly present the SSL and UL methods for standard image classification. However, their underlying learning rationales can be generalized to other challenging computer vision tasks, e.g., semantic segmentation~\cite{Ouali_2020_CVPR,chen2021semi}, object detection~\cite{jeong2019consistency,tang2021humble}, unsupervised domain adaptation~\cite{sun2019unsupervised-self,saito2020universal}, pose estimation~\cite{Chen_2019_ICCV,yang2021semihand}, 3D scene understanding~\cite{kim2021just}, video recognition~\cite{wang2015unsupervised,alwassel2019self}, etc. In the following, we review three core visual recognition tasks that widely benefit from SSL and UL methods to exploit unlabeled data: semantic segmentation (\S \ref{sec:segmentation}), object detection (\S \ref{sec:detection}), and unsupervised domain adaptation (\S \ref{sec:UDA}). 

\subsubsection{Semantic Segmentation}
\label{sec:segmentation}

Semantic segmentation aims to assign a semantic class label for each pixel in an input image. It is a core computer vision task that could be beneficial to various real-world applications such as medical image analysis~\cite{ronneberger2015u,huo2021atso,wu2021collaborative,huang2021graph} and autonomous driving~\cite{cordts2016cityscapes,yang2018denseaspp,behley2019semantickitti}. Supervised semantic segmentation requires tedious and expensive pixel-wise label annotations, e.g. manually annotating one single natural image in Cityscapes needs 1.5 hours~\cite{cordts2016cityscapes}. 

To reduce the annotation costs in semantic segmentation, a group of works consider only a small set of the training data annotated with per-pixel semantic labels while the rest of the training data being unlabeled -- known as {\bf semi-supervised semantic segmentation}. These works generally inherit similar learning rationales as SSL or UL for image classification, and adapt techniques such as consistency regularization~\cite{french2019semi,ouali2020semi,ke2020guided,hu2021semi}, self-training~\cite{mendel2020semi,zou2020pseudoseg,ibrahim2020semi,huo2021atso,chen2021semi,he2021re,yuan2021simple}, GAN frameworks~\cite{souly2017semi,hung2018adversarial,mittal2019semi} in SSL, or contrastive learning~\cite{lai2021semi,alonso2021semi,zhong2021pixel,zhou2021c3} in UL to learn from unlabeled images. Nevertheless, unsupervised loss terms in semantic segmentation are often required to impose in a per-pixel manner to align with the pixel-wise learning objective in semantic segmentation. In the following, we discuss the three most representative lines of state-of-the-art methods driven by recent advances in SSL and UL for semi-supervised semantic segmentation. 

{\em Consistency regularization} (\S \ref{sec:ssl-consistency}) can be generalized for pixel-wise tasks by formulating the consistency loss (Eq.~\eqref{eq:consist1}, Eq.~\eqref{eq:consist2}) at the pixel level. In a similar spirit as the standard consistency regularization in SSL, recent works in semi-supervised semantic segmentation~\cite{french2019semi,ouali2020semi,ke2020guided,hu2021semi} resort to enforcing pixel consistency among the images before and after perturbations, whilst perturbations being introduced at the input space~\cite{french2019semi} or feature space~\cite{ouali2020semi}. For instance, the first consistency regularization method in semantic segmentation~\cite{french2019semi} applies CutOut~\cite{devries2017improved} and CutMix~\cite{yun2019cutmix} augmentation techniques to perturb the input images with partial corruption, and imposes pixel-level loss terms to ensure the uncorrupted regions in perturbed images should have consistent pixel-wise predictions as the same regions in original images. A cross-consistency training~\cite{ouali2020semi} instead applies feature perturbations by injecting noise into network's activations and enforces pixel consistency between the clean and perturbed outputs. 

{\em Self-training} algorithms (\S \ref{sec:self-train}) are adapted and shown effective for semi-supervised semantic segmentation~\cite{mendel2020semi,zou2020pseudoseg,ibrahim2020semi,huo2021atso,chen2021semi,he2021re,yuan2021simple}, where pseudo segmentation maps on unlabeled images are propagated using a pre-trained teacher model~\cite{he2021re}, or a co-trained model~\cite{chen2021semi}. For example, a self-training method~\cite{he2021re} propagates pseudo segmentation labels with two steps -- (1) assigning pixel-wise pseudo labels on unlabeled data with a pre-trained teacher model; and (2) re-training a student model with the re-labeled dataset -- until no more performance gain is achieved. Another self-training approach~\cite{chen2021semi} adopts a co-training scheme by training two models to learn the per-pixel segmentation predictions from each others. 

{\em Contrastive learning} is widely used in UL and adapted to learn from unlabeled data in semantic segmentation~\cite{lai2021semi,alonso2021semi,zhong2021pixel,zhou2021c3}. To formulate the contrastive loss (Eq.~\eqref{eq:infonce}) per pixel, one needs select meaningful positive and negative pairs with consideration of pixel spatial locations. For this aim, a directional context-aware contrastive loss~\cite{lai2021semi} is proposed to crop two patches from one image, and take features at the same location as a positive pair and the rest as negative pairs. Another pixel contrastive loss~\cite{zhong2021pixel} is introduced to align the features before and after a random color augmentation by taking features at the same location as a positive pair, while sampling a fixed amount of negative pairs from different images. 

\subsubsection{Object Detection}
\label{sec:detection}

Object detection aims to predict a set of bounding boxes and the corresponding class labels for the objects of interest in an image. An object detector needs to unify classification and localization into one model by jointly training a classifier to predict class labels and a regression head to generate the bounding boxes~\cite{ren2015faster,carion2020end}. It is an important computer vision task that widely impacts different applications such as person search~\cite{xiao2017joint}, vehicle detection~\cite{qian2021robust}, logo detection~\cite{su2021multi}, text detection~\cite{feng2021semantic}, etc. Supervised object detection requires costly annotation efforts -- annotating the bounding box of a single object takes up to 42 seconds~\cite{russakovsky2015best}.

To exploit the unlabeled data without bounding box or class label information, a group of works in object detection exploit unlabeled data to boost model generalization by training on a small set of labeled data and a set of completely unlabeled images -- known as {\bf semi-supervised object detection}. These works mainly reformulate two streams of SSL techniques, including consistency regularization~\cite{jeong2019consistency,jeong2021interpolation,liu2021unbiased,tang2021humble,zhou2021instant,xu2021end} and self-training~\cite{radosavovic2018data,sohn2020simple,yang2021interactive,wang2021data,wang2021combating}, both of which introduce the learning targets for both bounding boxes and class labels to learn from the completely unlabeled data, as detailed next.

{\em Consistency regularization} (\S \ref{sec:ssl-consistency}) is introduced for semi-supervised object detection to propagate the soft label and bounding boxes assignment on unlabeled images based on dual consistency constraints on classification and regression~\cite{jeong2019consistency,jeong2021interpolation,liu2021unbiased,tang2021humble,zhou2021instant,xu2021end}. One line of works apply data augmentation such as random flipping~\cite{jeong2019consistency} and MixUp~\cite{zhang2017mixup} to generate augmented views of unlabeled images and encourage the predicted bounding boxes and its class labels remain consistent for the different views. Compared to standard consistency regularization, these methods especially need re-estimating the bounding box location in an augmented image, such as flip the bounding box~\cite{jeong2019consistency}, or calculate the overlapped bounding boxes of two mixed images in MixUp~\cite{zhang2017mixup}.
Another line of works follow a teacher-student training framework and impose teacher-student consistency~\cite{liu2021unbiased,tang2021humble,zhou2021instant,xu2021end} similar to Mean Teacher~\cite{tarvainen2017mean}. The teacher model is derived either from the student model via exponential mean average (EMA)~\cite{liu2021unbiased,tang2021humble,xu2021end}, or by applying non-maximum suppression (NMS, a filtering technique for refining the detected bounding boxes) on the instant model outputs~\cite{zhou2021instant} to obtain the pseudo bounding boxes and label annotations for training.

{\em Self-training} algorithms (\S \ref{sec:self-train}) are also introduced to annotated unlabeled images for object detection~\cite{radosavovic2018data,sohn2020simple,yang2021interactive,wang2021data,wang2021combating}. A simple self-training paradigm is to annotate the unlabeled images with bounding boxes and their class labels using a pre-trained teacher model and use these data for re-training~\cite{sohn2020simple}. However, such pseudo annotations may be rather noisy. To improve the quality of pseudo labels, recent works propose interactive self-training to progressively refine the pseudo labels with NMS~\cite{yang2021interactive}, or quantify model uncertainty to select or derive more reliable pseudo labels~\cite{wang2021data,wang2021combating} to learn from unlabeled data. 

 \subsubsection{Unsupervised Domain Adaptation} \label{sec:UDA}

 Unsupervised domain adaptation (UDA) is a special case of SSL where the labeled (source) and unlabeled (target) data lie in different distributions, a.k.a. different domains. UDA is essential for visual recognition~\cite{saenko2010adapting}, as the statistical properties of visual data are sensitive to a wider variety of factors, \eg, illumination, viewpoint, resolution, occlusion, times of the day, and weather conditions. While most UDA methods focus on tackling the {domain gap} between the labeled and unlabeled data, SSL and UL algorithms can also be adapted to learn from unlabeled data in UDA, as follows. 

{\em Consistency regularization} (\S \ref{sec:ssl-consistency}) is shown to be effective in UDA. In the same spirit of encouraging consistent outputs under perturbations, various UDA approaches apply input transformations or model ensembling to simulate variations in input or model space~\cite{wu2020dualmixup,french2018self,tarvainen2017mean,deng2019cluster}. To generate input variations, a dual MixUp regularization integrates category-level MixUp and domain-level MixUp to regularize the model with consistency constraints, thus learning from unlabeled data to enhance domain-invariance~\cite{wu2020dualmixup}. To generate model variations, self-ensembling~\cite{french2018self} utilizes the Mean Teacher~\cite{tarvainen2017mean} to impute unlabeled training targets in target domain. 

{\em Self-training} (\S \ref{sec:self-train}) has been also useful for UDA. Similar to SSL, self-training for UDA include three streams of techniques to impute pseudo labels on the unlabeled target samples, including entropy minimization, pseudo-label and co-training. To ensure the effectiveness, self-training methods are often coupled with domain distribution alignment for reducing the domain shift. 
For instance, entropy minimization (Eq.~\eqref{eq:ent}) is adopted for UDA ~\cite{carlucci2017autodial,morerio2018minimal,hu2020synchronization}, in combination with distribution alignment techniques such as domain-specific batch normalization layers~\cite{carlucci2017autodial}, aligning second-order statistics of features~\cite{morerio2018minimal}, or adversarial training and gradient synchronization~\cite{hu2020synchronization}. 
Co-training (Eq.~\eqref{eq:cot}) is also introduced for UDA, which imputes training targets from multiple co-trained classifiers to learn from unlabeled data and match cross-domain distributions~\cite{saito2017asymmetric}. 

{\em Deep generative models} (DGMs), as a class of models for SSL and UL (\S \ref{sec:DGMs}, \S \ref{sec:ul-dgm}), are widely adopted for UDA. In contrast to other UDA methods that reduce the domain shift at the feature level, DGMs provide an alternative and complementary solution to mitigate the domain discrepancy at pixel level by cross-domain image-to-image translation. 
The majority of these frameworks are based on GANs, such as PixelDA~\cite{bousmalis2017unsupervised}, generate to adapt~\cite{sankaranarayanan2018learning}, and GANs with cycle-consistency like CyCADA\cite{hoffman18cycada}, SBADA-GAN~\cite{russo2018sbadagan}, I2I Adapt~\cite{murez2018image} and CrDoCo~\cite{chen2019crdoco}.
These models typically learn a {real-to-real}~\cite{yoo2016pixel,hoffman18cycada,chen2019crdoco,russo2018sbadagan} or {synthetic-to-real}~\cite{shrivastava2017learning,bousmalis2017unsupervised,sankaranarayanan2018learning} {mapping} to render the image style from the labeled source to the unlabeled target domain, thus offering synthetic training data with pseudo labels. 

{\em Self-supervised learning} popularized in SSL and UL (\S \ref{sec:ssl-self}, \S \ref{sec:pretext}), is also introduced in UDA to construct auxiliary self-supervised learning objectives on unlabeled data. Self-supervised models often address the UDA problem by self-supervision coupled with a supervised objective on the labeled source data~\cite{carlucci2019domain,sun2019unsupervised-self,xu2019self-da,bucci2020self}. 
The pioneer work in this direction is JiGen~\cite{carlucci2019domain}, which learns jointly to classify objects and solve the jigsaw puzzles~\cite{noroozi2016unsupervised} pretext task to achieve better generalization in new domains. Recent works~\cite{sun2019unsupervised-self,xu2019self-da,bucci2020self} explored other self-supervised pretext tasks such as predicting rotation~\cite{sun2019unsupervised-self,xu2019self-da,bucci2020self}, flipping~\cite{sun2019unsupervised-self} and patch ordering~\cite{sun2019unsupervised-self}. 
Besides pretext tasks, recent UDA methods also explored discriminative self-supervision signals based on clustering or contrastive learning. For instance, DANCE~\cite{saito2020universal} performs neighborhood clustering by assigning the target samples to a “known” class prototype in the source domain or its neighbor in the target domain.
Gradient regularized contrastive learning~\cite{su2020gradient} leverages the contrastive loss to push unlabeled target samples towards the most similar labeled source samples. Similarly, \cite{wang2022cross} aligns target domain features to class prototypes in the source domain through contrastive loss, minimizing the distances between the cross-domain samples that likely belong to the same class.

\section{Emerging Trends and Open Challenges}
\label{sec:future}

In this section, we discuss the emerging trends in SSL and UL from unlabeled data, covering three directions, namely open-set learning (\S \ref{sec:open-set}), incremental learning (\S \ref{sec:continual}) and multi-modal learning (\S \ref{sec:multimodal}). We detail both recent developments and open challenges.

\subsection{Open-Set Learning from Unlabeled Data}
\label{sec:open-set}

In \S \ref{sec:ssl},  we review works addressing the relatively simple {closed-set} learning in SSL, which assume that unlabeled data share the same label space as the labeled one. However, this {closed-set} assumption may greatly hinder the effectiveness of SSL in leveraging real-world uncurated unlabeled data that contains unseen classes, \ie, out-of-distribution (OOD) samples (also known as outliers)~\cite{oliver2018realistic}. When applying most existing SSL methods to {open-set learning} with noisy unlabeled data, their model performance may degrade significantly, as the OOD samples could induce catastrophic error propagation. 

A line of works propose to address a more complex \textit{open-set SSL} scenario~\cite{chen2020semi,guo2020safe,yu2020multi,augustin2020out,huang2021trash,saito2021openmatch,huang2021universal,cao2021open}, where the unlabeled set contains task-irrelevant OOD data. In this setup (so-called {open-world} SSL), unlabeled samples are not all beneficial. To prevent possible performance hazards caused by unlabeled OOD samples, recent advances in SSL propose various sample-specific selection strategies to discount their importance or usage \cite{chen2020semi,guo2020safe,yu2020multi,augustin2020out}. The pioneer works including UASD~\cite{chen2020semi} and DS$^3$L~\cite{guo2020safe} propose to impose a dynamic weighting function to down-weight the unsupervised regularization loss term proportional to the likelihood that an unlabeled sample belongs to an unseen class. 
Follow-up works resort to curriculum learning~\cite{yu2020multi} and iterative self-training~\cite{augustin2020out} by training an OOD classifier to detect and discard the potentially detrimental samples. {More recently, OpenMatch~\cite{saito2021openmatch} propose to train a set of one-vs-all classifiers for detecting inliers and outliers and regularize the model with a consistency constraint on only the unlabeled inliers.}

\myparagraph{Open Challenges.}
The open-set SSL calls for integrating OOD detection~\cite{hendrycks2016baseline} or novel class discovery~\cite{zhong2021openmix} with semi-supervised learning in a unified model to advance selective exploitation of noisy unlabeled data. {Moreover, a more recent work propose a universal SSL benchmark~\cite{huang2021universal} which further extends the distribution mismatch problem in open-set setup as subset or intersectional class mismatch, and feature distribution mismatch.} These more realistic setups pose multiple new challenges, including confidence calibration of DNN for OOD detection~\cite{hendrycks2016baseline,hendrycks2016baseline,hendrycks2018deep,lee2018simple,hein2019relu}, imbalanced class distribution caused by real-world long-tailed distributed unlabeled data~\cite{kim2020distribution,lee2021abc}, and discovery of unseen classes in unlabeled data~\cite{han2019learning,han2020automatically,zhong2021openmix}. Although recent advances in open-set SSL have explored OOD detection, the other challenges remain to be resolved to exploit real-world unlabeled data. 

\subsection{Incremental Learning from Unlabeled Data}
\label{sec:continual}

Existing works on SSL and UL often assume all unlabeled training data is available at once, which however may not always hold in practice due to privacy concerns or computational constraints. In many realistic scenarios, we need to perform {\em incremental learning} (IL) with new data to update the model incrementally without access to past training data. Here we review research directions on IL from unlabeled data~\cite{lee2019overcoming,rao2019continual} and discuss its open challenges. 

{Incremental learning (IL) from unlabeled data has been investigated in a semi-supervised fashion~\cite{lee2019overcoming}.} IL (also known as {continual learning} and {lifelong learning}~\cite{delange2021continual}) aims to extend an existing model's knowledge without accessing the previous training data. Most existing IL approaches use regularization objectives to not forget old knowledge, \ie, reducing catastrophic forgetting~\cite{mccloskey1989catastrophic,rebuffi2017icarl,kirkpatrick2017overcoming,chaudhry2018riemannian}. 
To this aim, unlabeled data is often used in IL to prevent catastrophic forgetting by estimating the importance weights of model parameters for old tasks~\cite{aljundi2018memory}, or formulating a knowledge distillation objective~\cite{lee2019overcoming,zhang2020class} to consolidate the knowledge learned from old data. Recently, multiple works explore IL from unlabeled data that comes as a non-stationary stream~\cite{rao2019continual,li2019incremental}, with the class label space possibly varying over time~\cite{smith2019unsupervised}. In this setting, the goal is to learn a salient representation from continuous incoming unlabeled data stream. 
To expand the representations for novel classes and unlabeled data, several strategies are adopted to dynamically update representations in the latent space, such as creating new cluster centroids by online clustering~\cite{smith2019unsupervised} and updating mixture-of-Gaussians~\cite{rao2019continual}. 
{Some recent works apply self-supervised techniques on the unlabeled test-data~\cite{sun2020test,varsavsky2020test,wangtent}, which is useful to overcome possible shifts in the data distribution~\cite{hoffman2014continuous}.}

\myparagraph{Open Challenges.} Incremental learning from unlabeled data requires solving multiple challenges, ranging from catastrophic forgetting~\cite{lee2019overcoming,bobu2018adapting}, modeling new concepts~\cite{rao2019continual,smith2019unsupervised} to predicting the evolution of data streams~\cite{hoffman2014continuous}. 
Due to lacking the access to all the unlabeled training data at once, addressing these challenges is nontrivial as directly applying many existing SSL and UL methods could not guarantee good generalization performance. As an example, pseudo labels may suffer the confirmation bias problem~\cite{arazo2020pseudo} when classifying unseen unlabeled data.  
Thus, incremental learning from a stream of potentially non-\textit{i.i.d.} unlabeled data remains an open challenge.  

\subsection{Multi-Modal Learning from Unlabeled Data}
\label{sec:multimodal}

A growing number of works combine visual and non-visual modalities (\eg, text, audio) to form discriminative self-supervision signals that enable learning from multi-modal unlabeled data. To bring {vision and language} for unsupervised learning, variants of vision and language BERT models (\eg, ViLBERT~\cite{lu2019vilbert}, LXMERT~\cite{tan2019lxmert}, VL-BERT~\cite{su2019vl}, Uniter~\cite{chen2020uniter} and Unicoder-VL~\cite{li2020unicoder}) are built upon the transformer blocks~\cite{vaswani2017attention} to jointly model images and natural language in an unsupervised way. Specifically, the visual, linguistic or their joint representations can be learned in an unsupervised manner by solving the {\em Cloze task} in natural language processing which predicts the masked words in the input sentences~\cite{devlin2018bert}, or by optimizing a linguistic-visual alignment objective~\cite{tan2019lxmert,sun2019videobert}. 
Another line of works utilize the language supervision (\eg, from web data~\cite{stroud2020learning} or narrated materials \cite{alayrac2020self,miech2020end,rouditchenko2020avlnet,miech2019howto100m,radford2021learning,jia2021scaling}) to guide unsupervised representation learning by aligning images and languages in the shared latent space, as exemplified by CLIP~\cite{radford2021learning} and ALIGN~\cite{jia2021scaling}.

Similarly, to combine {audio and visual modalities} for unsupervised learning, existing works exploit the natural {audio-visual correspondence} in videos to formulate various self-supervised signals, which predict the cross-modal correspondence~\cite{arandjelovic2017look,owens2018audio}, align the temporally corresponding representations~\cite{korbar2018cooperative,morgado2020audio,patrick2020multi,miech2020end}, or cluster their representations in a shared audio-visual latent space~\cite{owens2016ambient,alwassel2019self}. 
Several works further explore {audio, vision and language} together for unsupervised representation learning by aligning different modalities in a shared multi-modal latent space~\cite{rouditchenko2020avlnet,hu2020semi} or in a hierarchical latent space for audio-vision and vision-language~\cite{alayrac2020self}.

\myparagraph{Open Challenges.} The success of multi-modal learning from unlabeled data often relies on an assumption that different modalities are semantically correlated. For instance, when clustering audio and video data for unsupervised representation learning~\cite{alwassel2019self}, or transferring text knowledge to the unlabeled image data~\cite{li2020hda-semantic}, the two data modalities are assumed to share similar semantics. However, this assumption may not hold in real-world data, leading to degraded model performance~\cite{chen2021distilling,miech2020end}. Thus, it remains an open challenge to learn from the multi-modal unlabeled data that contains a semantic gap across modalities.

\section{Conclusion}

Learning visual representations with limited or no manual supervision is critical for scalable computer vision applications. Semi-supervised learning (SSL) and unsupervised learning (UL) models provide feasible and promising solutions to learn from unlabeled visual data. In this comprehensive survey, we have introduced unified problem definitions and taxonomies to summarize and correlate a wide variety of recent advanced and popularized SSL and UL deep learning methodologies for building superior visual classification models. 
We believe that our concise taxonomies of existing algorithms and extensive discussions of emerging trends help to better understand the status quo of research in visual representation learning with unlabeled data, as well as to inspire new learning solutions for major unresolved challenges involved in the limited-label regime. 

\ifCLASSOPTIONcompsoc
  \section*{Acknowledgments}
\else
  \section*{Acknowledgment}
\fi

This work has been partially funded by the ERC (853489-DEXIM) and the DFG (2064/1–Project number 390727645).

\ifCLASSOPTIONcaptionsoff
  \newpage
\fi

\bibliographystyle{IEEEtran}
{\footnotesize \bibliography{reference}}

\begin{thebibliography}{100}
\providecommand{\url}[1]{#1}
\csname url@samestyle\endcsname
\providecommand{\newblock}{\relax}
\providecommand{\bibinfo}[2]{#2}
\providecommand{\BIBentrySTDinterwordspacing}{\spaceskip=0pt\relax}
\providecommand{\BIBentryALTinterwordstretchfactor}{4}
\providecommand{\BIBentryALTinterwordspacing}{\spaceskip=\fontdimen2\font plus
\BIBentryALTinterwordstretchfactor\fontdimen3\font minus
  \fontdimen4\font\relax}
\providecommand{\BIBforeignlanguage}[2]{{%
\expandafter\ifx\csname l@#1\endcsname\relax
\typeout{** WARNING: IEEEtran.bst: No hyphenation pattern has been}%
\typeout{** loaded for the language `#1'. Using the pattern for}%
\typeout{** the default language instead.}%
\else
\language=\csname l@#1\endcsname
\fi
#2}}
\providecommand{\BIBdecl}{\relax}
\BIBdecl

\bibitem{lecun2015deep}
Y.~LeCun, Y.~Bengio, and G.~Hinton, ``Deep learning,'' \emph{Nature}, 2015.

\bibitem{goodfellow2016deep}
I.~Goodfellow, Y.~Bengio, and A.~Courville, ``Deep learning,'' \emph{MIT
  press}, 2016.

\bibitem{krizhevsky2012imagenet}
A.~Krizhevsky, I.~Sutskever, and G.~E. Hinton, ``Imagenet classification with
  deep convolutional neural networks,'' in \emph{NeurIPS}, 2012.

\bibitem{schroff2015facenet}
F.~Schroff, D.~Kalenichenko, and J.~Philbin, ``Facenet: A unified embedding for
  face recognition and clustering,'' in \emph{CVPR}, 2015.

\bibitem{ren2015faster}
S.~Ren, K.~He, R.~Girshick, and J.~Sun, ``Faster r-cnn: Towards real-time
  object detection with region proposal networks,'' in \emph{NeurIPS}, 2015.

\bibitem{chen2017deeplab}
L.-C. Chen, G.~Papandreou, I.~Kokkinos, K.~Murphy, and A.~L. Yuille, ``Deeplab:
  Semantic image segmentation with deep convolutional nets, atrous convolution,
  and fully connected crfs,'' \emph{IEEE TPAMI}, 2017.

\bibitem{zhou2005semi}
Z.-H. Zhou and M.~Li, ``Semi-supervised regression with co-training.'' in
  \emph{IJCAI}, 2005.

\bibitem{chapelle2010semi}
O.~Chapelle, B.~Scholkopf, and A.~Zien, ``Semi-supervised learning,''
  \emph{IEEE TNNLS}, 2009.

\bibitem{weinberger2006unsupervised}
K.~Q. Weinberger and L.~K. Saul, ``Unsupervised learning of image manifolds by
  semidefinite programming,'' \emph{IJCV}, 2006.

\bibitem{bengio2013representation}
Y.~Bengio, A.~Courville, and P.~Vincent, ``Representation learning: A review
  and new perspectives,'' \emph{IEEE TPAMI}, 2013.

\bibitem{doersch2015unsupervised}
C.~Doersch, A.~Gupta, and A.~A. Efros, ``Unsupervised visual representation
  learning by context prediction,'' in \emph{ICCV}, 2015.

\bibitem{chen2020simple}
T.~Chen, S.~Kornblith, M.~Norouzi, and G.~Hinton, ``A simple framework for
  contrastive learning of visual representations,'' in \emph{ICML}, 2020.

\bibitem{zhu2006semi}
X.~J. Zhu, ``Semi-supervised learning literature survey,'' University of
  Wisconsin-Madison Department of Computer Sciences, Tech. Rep., 2005.

\bibitem{chen2020semi}
Y.~Chen, X.~Zhu, W.~Li, and S.~Gong, ``Semi-supervised learning under class
  distribution mismatch.'' in \emph{AAAI}, 2020.

\bibitem{guo2020safe}
L.-Z. Guo, Z.-Y. Zhang, Y.~Jiang, Y.-F. Li, and Z.-H. Zhou, ``Safe deep
  semi-supervised learning for unseen-class unlabeled data,'' in \emph{ICML},
  2020.

\bibitem{he2019momentum}
K.~He, H.~Fan, Y.~Wu, S.~Xie, and R.~Girshick, ``Momentum contrast for
  unsupervised visual representation learning,'' in \emph{CVPR}, 2020.

\bibitem{van2020survey}
J.~E. Van~Engelen and H.~H. Hoos, ``A survey on semi-supervised learning,''
  \emph{ML}, 2020.

\bibitem{jing2020self}
L.~Jing and Y.~Tian, ``Self-supervised visual feature learning with deep neural
  networks: A survey,'' \emph{IEEE TPAMI}, 2020.

\bibitem{schmarje2020survey}
L.~Schmarje, M.~Santarossa, S.-M. Schr{\"o}der, and R.~Koch, ``A survey on
  semi-, self-and unsupervised learning for image classification,''
  \emph{arXiv:2002.08721}, 2020.

\bibitem{qi2020small}
G.-J. Qi and J.~Luo, ``Small data challenges in big data era: A survey of
  recent progress on unsupervised and semi-supervised methods,'' \emph{IEEE
  TPAMI}, 2020.

\bibitem{fergus2009semi}
R.~Fergus, Y.~Weiss, and A.~Torralba, ``Semi-supervised learning in gigantic
  image collections,'' in \emph{NeurIPS}, 2009.

\bibitem{papernot2017semi}
N.~Papernot, M.~Abadi, {\'U}.~Erlingsson, I.~Goodfellow, and K.~Talwar,
  ``Semi-supervised knowledge transfer for deep learning from private training
  data,'' in \emph{ICLR}, 2017.

\bibitem{blum1998combining}
A.~Blum and T.~Mitchell, ``Combining labeled and unlabeled data with
  co-training,'' in \emph{COLT}, 1998.

\bibitem{nigam2000analyzing}
K.~Nigam and R.~Ghani, ``Analyzing the effectiveness and applicability of
  co-training,'' in \emph{CIKM}, 2000.

\bibitem{libbrecht2015machine}
M.~W. Libbrecht and W.~S. Noble, ``Machine learning applications in genetics
  and genomics,'' \emph{Nature Reviews Genetics}, 2015.

\bibitem{berthelot2019mixmatch}
D.~Berthelot, N.~Carlini, I.~Goodfellow, N.~Papernot, A.~Oliver, and C.~Raffel,
  ``Mixmatch: A holistic approach to semi-supervised learning,'' in
  \emph{NeurIPS}, 2019.

\bibitem{berthelot2019remixmatch}
D.~Berthelot, N.~Carlini, E.~D. Cubuk, A.~Kurakin, K.~Sohn, H.~Zhang, and
  C.~Raffel, ``Remixmatch: Semi-supervised learning with distribution alignment
  and augmentation anchoring,'' in \emph{ICLR}, 2020.

\bibitem{Jang_2020_CVPR}
Y.~K. Jang and N.~I. Cho, ``Generalized product quantization network for
  semi-supervised image retrieval,'' in \emph{CVPR}, 2020.

\bibitem{Gao_2019_ICCV}
J.~Gao, J.~Wang, S.~Dai, L.-J. Li, and R.~Nevatia, ``Note-rcnn: Noise tolerant
  ensemble rcnn for semi-supervised object detection,'' in \emph{ICCV}, 2019.

\bibitem{tang2021humble}
Y.~Tang, W.~Chen, Y.~Luo, and Y.~Zhang, ``Humble teachers teach better students
  for semi-supervised object detection,'' in \emph{CVPR}, 2021.

\bibitem{Kalluri_2019_ICCV}
T.~Kalluri, G.~Varma, M.~Chandraker, and C.~Jawahar, ``Universal
  semi-supervised semantic segmentation,'' in \emph{ICCV}, 2019.

\bibitem{Ouali_2020_CVPR}
Y.~Ouali, C.~Hudelot, and M.~Tami, ``Semi-supervised semantic segmentation with
  cross-consistency training,'' in \emph{CVPR}, 2020.

\bibitem{Ibrahim_2020_CVPR}
M.~S. Ibrahim, A.~Vahdat, M.~Ranjbar, and W.~G. Macready, ``Semi-supervised
  semantic image segmentation with self-correcting networks,'' in \emph{CVPR},
  2020.

\bibitem{Chen_2019_ICCV}
Y.~Chen, Z.~Tu, L.~Ge, D.~Zhang, R.~Chen, and J.~Yuan, ``So-handnet:
  Self-organizing network for 3d hand pose estimation with semi-supervised
  learning,'' in \emph{ICCV}, 2019.

\bibitem{radosavovic2018data}
I.~Radosavovic, P.~Doll{\'a}r, R.~Girshick, G.~Gkioxari, and K.~He, ``Data
  distillation: Towards omni-supervised learning,'' in \emph{CVPR}, 2018.

\bibitem{Mitra_2020_CVPR}
R.~Mitra, N.~B. Gundavarapu, A.~Sharma, and A.~Jain, ``Multiview-consistent
  semi-supervised learning for 3d human pose estimation,'' in \emph{CVPR},
  2020.

\bibitem{laine2016temporal}
S.~Laine and T.~Aila, ``Temporal ensembling for semi-supervised learning,'' in
  \emph{ICLR}, 2017.

\bibitem{sohn2020fixmatch}
K.~Sohn, D.~Berthelot, C.-L. Li, Z.~Zhang, N.~Carlini, E.~D. Cubuk, A.~Kurakin,
  H.~Zhang, and C.~Raffel, ``Fixmatch: Simplifying semi-supervised learning
  with consistency and confidence,'' in \emph{NeurIPS}, 2020.

\bibitem{tarvainen2017mean}
A.~Tarvainen and H.~Valpola, ``Mean teachers are better role models:
  Weight-averaged consistency targets improve semi-supervised deep learning
  results,'' in \emph{NeurIPS}, 2017.

\bibitem{oliver2018realistic}
A.~Oliver, A.~Odena, C.~A. Raffel, E.~D. Cubuk, and I.~Goodfellow, ``Realistic
  evaluation of deep semi-supervised learning algorithms,'' in \emph{NeurIPS},
  2018.

\bibitem{athiwaratkun2018there}
B.~Athiwaratkun, M.~Finzi, P.~Izmailov, and A.~G. Wilson, ``There are many
  consistent explanations of unlabeled data: Why you should average,'' in
  \emph{ICLR}, 2019.

\bibitem{zhou2004learning}
D.~Zhou, O.~Bousquet, T.~N. Lal, J.~Weston, and B.~Sch{\"o}lkopf, ``Learning
  with local and global consistency,'' in \emph{NeurIPS}, 2004.

\bibitem{chapelle2002cluster}
O.~Chapelle, J.~Weston, and B.~Sch{\"o}lkopf, ``Cluster kernels for
  semi-supervised learning,'' in \emph{NeurIPS}, 2002.

\bibitem{weston2012deep}
J.~Weston, F.~Ratle, H.~Mobahi, and R.~Collobert, ``Deep learning via
  semi-supervised embedding,'' in \emph{ICML}, 2008.

\bibitem{chapelle2005semi}
O.~Chapelle, A.~Zien, C.~Z. Ghahramani \emph{et~al.}, ``Semi-supervised
  classification by low density separation,'' in \emph{AISTATSW}, 2005.

\bibitem{sajjadi2016regularization}
M.~Sajjadi, M.~Javanmardi, and T.~Tasdizen, ``Regularization with stochastic
  transformations and perturbations for deep semi-supervised learning,'' in
  \emph{NeurIPS}, 2016.

\bibitem{wang2020enaet}
X.~Wang, D.~Kihara, J.~Luo, and G.-J. Qi, ``Enaet: A self-trained framework for
  semi-supervised and supervised learning with ensemble transformations,''
  \emph{IEEE TIP}, 2020.

\bibitem{miyato2015distributional}
T.~Miyato, S.-i. Maeda, M.~Koyama, K.~Nakae, and S.~Ishii, ``Distributional
  smoothing with virtual adversarial training,'' in \emph{ICLR}, 2016.

\bibitem{miyato2018virtual}
T.~Miyato, S.-i. Maeda, M.~Koyama, and S.~Ishii, ``Virtual adversarial
  training: a regularization method for supervised and semi-supervised
  learning,'' \emph{IEEE TPAMI}, 2018.

\bibitem{verma2019interpolation}
V.~Verma, A.~Lamb, J.~Kannala, Y.~Bengio, and D.~Lopez-Paz, ``Interpolation
  consistency training for semi-supervised learning,'' in \emph{IJCAI}, 2019.

\bibitem{xie2020unsupervised}
Q.~Xie, Z.~Dai, E.~Hovy, T.~Luong, and Q.~Le, ``Unsupervised data augmentation
  for consistency training,'' in \emph{NeurIPS}, 2020.

\bibitem{bachman2014learning}
P.~Bachman, O.~Alsharif, and D.~Precup, ``Learning with pseudo-ensembles,'' in
  \emph{NeurIPS}, 2014.

\bibitem{rasmus2015semi}
A.~Rasmus, M.~Berglund, M.~Honkala, H.~Valpola, and T.~Raiko, ``Semi-supervised
  learning with ladder networks,'' in \emph{NeurIPS}, 2015.

\bibitem{park2017adversarial}
S.~Park, J.-K. Park, S.-J. Shin, and I.-C. Moon, ``Adversarial dropout for
  supervised and semi-supervised learning,'' in \emph{AAAI}, 2018.

\bibitem{zhang2020wcp}
L.~Zhang and G.-J. Qi, ``Wcp: Worst-case perturbations for semi-supervised deep
  learning,'' in \emph{CVPR}, 2020.

\bibitem{lee2013pseudo}
D.-H. Lee, ``Pseudo-label: The simple and efficient semi-supervised learning
  method for deep neural networks,'' in \emph{ICMLW}, 2013.

\bibitem{chen2018semi}
Y.~Chen, X.~Zhu, and S.~Gong, ``Semi-supervised deep learning with memory,'' in
  \emph{ECCV}, 2018.

\bibitem{qiao2018deep}
S.~Qiao, W.~Shen, Z.~Zhang, B.~Wang, and A.~Yuille, ``Deep co-training for
  semi-supervised image recognition,'' in \emph{ECCV}, 2018.

\bibitem{dong2018tri}
W.~Dong-DongChen and Z.-H. WeiGao, ``Tri-net for semi-supervised deep
  learning,'' in \emph{IJCAI}, 2018.

\bibitem{xie2020self}
Q.~Xie, M.-T. Luong, E.~Hovy, and Q.~V. Le, ``Self-training with noisy student
  improves imagenet classification,'' in \emph{CVPR}, 2020.

\bibitem{luo2018smooth}
Y.~Luo, J.~Zhu, M.~Li, Y.~Ren, and B.~Zhang, ``Smooth neighbors on teacher
  graphs for semi-supervised learning,'' in \emph{CVPR}, 2018.

\bibitem{kipf2016semi}
T.~N. Kipf and M.~Welling, ``Semi-supervised classification with graph
  convolutional networks,'' in \emph{ICLR}, 2017.

\bibitem{iscen2019label}
A.~Iscen, G.~Tolias, Y.~Avrithis, and O.~Chum, ``Label propagation for deep
  semi-supervised learning,'' in \emph{CVPR}, 2019.

\bibitem{kingma2014semi}
D.~P. Kingma, S.~Mohamed, D.~J. Rezende, and M.~Welling, ``Semi-supervised
  learning with deep generative models,'' in \emph{NeurIPS}, 2014.

\bibitem{maaloe2016auxiliary}
L.~Maal{\o}e, C.~K. S{\o}nderby, S.~K. S{\o}nderby, and O.~Winther, ``Auxiliary
  deep generative models,'' in \emph{ICML}, 2016.

\bibitem{springenberg2015unsupervised}
J.~T. Springenberg, ``Unsupervised and semi-supervised learning with
  categorical generative adversarial networks,'' in \emph{ICLR}, 2016.

\bibitem{salimans2016improved}
T.~Salimans, I.~Goodfellow, W.~Zaremba, V.~Cheung, A.~Radford, and X.~Chen,
  ``Improved techniques for training gans,'' in \emph{NeurIPS}, 2016.

\bibitem{dumoulin2016adversarially}
V.~Dumoulin, I.~Belghazi, B.~Poole, A.~Lamb, M.~Arjovsky, O.~Mastropietro, and
  A.~Courville, ``Adversarially learned inference,'' in \emph{ICLR}, 2017.

\bibitem{dai2017good}
Z.~Dai, Z.~Yang, F.~Yang, W.~W. Cohen, and R.~R. Salakhutdinov, ``Good
  semi-supervised learning that requires a bad gan,'' in \emph{NeurIPS}, 2017.

\bibitem{qi2018global}
G.-J. Qi, L.~Zhang, H.~Hu, M.~Edraki, J.~Wang, and X.-S. Hua, ``Global versus
  localized generative adversarial nets,'' in \emph{CVPR}, 2018.

\bibitem{zhai2019s4l}
X.~Zhai, A.~Oliver, A.~Kolesnikov, and L.~Beyer, ``S4l: Self-supervised
  semi-supervised learning,'' in \emph{ICCV}, 2019.

\bibitem{chen2020big}
T.~Chen, S.~Kornblith, K.~Swersky, M.~Norouzi, and G.~E. Hinton, ``Big
  self-supervised models are strong semi-supervised learners,'' in
  \emph{NeurIPS}, 2020.

\bibitem{zhu2002learning}
X.~Zhu and Z.~Ghahramani, ``Learning from labeled and unlabeled data with label
  propagation,'' \emph{Technical Report, Carnegie Mellon University}, 2002.

\bibitem{suzuki2020adversarial}
T.~Suzuki and I.~Sato, ``Adversarial transformations for semi-supervised
  learning.'' in \emph{AAAI}, 2020.

\bibitem{zhang2017mixup}
H.~Zhang, M.~Cisse, Y.~N. Dauphin, and D.~Lopez-Paz, ``mixup: Beyond empirical
  risk minimization,'' in \emph{ICLR}, 2018.

\bibitem{cubuk2018autoaugment}
E.~D. Cubuk, B.~Zoph, D.~Mane, V.~Vasudevan, and Q.~V. Le, ``Autoaugment:
  Learning augmentation policies from data,'' \emph{arXiv:1805.09501}, 2018.

\bibitem{cubuk2020randaugment}
E.~D. Cubuk, B.~Zoph, J.~Shlens, and Q.~V. Le, ``Randaugment: Practical
  automated data augmentation with a reduced search space,'' in \emph{CVPRW},
  2020.

\bibitem{devries2017improved}
T.~DeVries and G.~W. Taylor, ``Improved regularization of convolutional neural
  networks with cutout,'' \emph{arXiv:1708.04552}, 2017.

\bibitem{szegedy2013intriguing}
C.~Szegedy, W.~Zaremba, I.~Sutskever, J.~Bruna, D.~Erhan, I.~Goodfellow, and
  R.~Fergus, ``Intriguing properties of neural networks,'' in \emph{ICLR},
  2014.

\bibitem{kurakin2016adversarial}
A.~Kurakin, I.~Goodfellow, and S.~Bengio, ``Adversarial examples in the
  physical world,'' in \emph{ICLR}, 2017.

\bibitem{najafi2019robustness}
A.~Najafi, S.-i. Maeda, M.~Koyama, and T.~Miyato, ``Robustness to adversarial
  perturbations in learning from incomplete data,'' in \emph{NeurIPS}, 2019.

\bibitem{carmon2019unlabeled}
Y.~Carmon, A.~Raghunathan, L.~Schmidt, J.~C. Duchi, and P.~S. Liang,
  ``Unlabeled data improves adversarial robustness,'' in \emph{NeurIPS}, 2019.

\bibitem{lim2019fast}
S.~Lim, I.~Kim, T.~Kim, C.~Kim, and S.~Kim, ``Fast autoaugment,'' in
  \emph{NeurIPS}, 2019.

\bibitem{ho2019population}
D.~Ho, E.~Liang, X.~Chen, I.~Stoica, and P.~Abbeel, ``Population based
  augmentation: Efficient learning of augmentation policy schedules,'' in
  \emph{ICML}, 2019.

\bibitem{zhang2019adversarial}
X.~Zhang, Q.~Wang, J.~Zhang, and Z.~Zhong, ``Adversarial autoaugment,'' in
  \emph{ICLR}, 2019.

\bibitem{izmailov2018averaging}
P.~Izmailov, D.~Podoprikhin, T.~Garipov, D.~Vetrov, and A.~G. Wilson,
  ``Averaging weights leads to wider optima and better generalization,'' in
  \emph{UAI}, 2018.

\bibitem{mitchell1982generalization}
T.~M. Mitchell, ``Generalization as search,'' \emph{AI}, 1982.

\bibitem{schapire1990strength}
R.~E. Schapire, ``The strength of weak learnability,'' \emph{ML}, 1990.

\bibitem{breiman2001random}
L.~Breiman, ``Random forests,'' \emph{ML}, 2001.

\bibitem{freund1999short}
Y.~Freund, R.~Schapire, and N.~Abe, ``A short introduction to boosting,''
  \emph{Journal-Japanese Society For Artificial Intelligence}, 1999.

\bibitem{lakshminarayanan2017simple}
B.~Lakshminarayanan, A.~Pritzel, and C.~Blundell, ``Simple and scalable
  predictive uncertainty estimation using deep ensembles,'' in \emph{NeurIPS},
  2017.

\bibitem{grandvalet2005semi}
Y.~Grandvalet and Y.~Bengio, ``Semi-supervised learning by entropy
  minimization,'' in \emph{NeurIPS}, 2005.

\bibitem{sajjadi2016mutual}
M.~Sajjadi, M.~Javanmardi, and T.~Tasdizen, ``Mutual exclusivity loss for
  semi-supervised deep learning,'' in \emph{ICIP}, 2016.

\bibitem{snell2017prototypical}
J.~Snell, K.~Swersky, and R.~Zemel, ``Prototypical networks for few-shot
  learning,'' in \emph{NeurIPS}, 2017.

\bibitem{ke2019dual}
Z.~Ke, D.~Wang, Q.~Yan, J.~Ren, and R.~W. Lau, ``Dual student: Breaking the
  limits of the teacher in semi-supervised learning,'' in \emph{CVPR}, 2019.

\bibitem{goodfellow2014explaining}
I.~J. Goodfellow, J.~Shlens, and C.~Szegedy, ``Explaining and harnessing
  adversarial examples,'' in \emph{ICLR}, 2015.

\bibitem{hinton2015distilling}
G.~Hinton, O.~Vinyals, and J.~Dean, ``Distilling the knowledge in a neural
  network,'' \emph{arXiv:1503.02531}, 2015.

\bibitem{bucilua2006model}
C.~Buciluǎ, R.~Caruana, and A.~Niculescu-Mizil, ``Model compression,'' in
  \emph{ACM SIGKDD}, 2006.

\bibitem{ba2014deep}
J.~Ba and R.~Caruana, ``Do deep nets really need to be deep?'' \emph{NeurIPS},
  2014.

\bibitem{yalniz2019billion}
I.~Z. Yalniz, H.~J{\'e}gou, K.~Chen, M.~Paluri, and D.~Mahajan, ``Billion-scale
  semi-supervised learning for image classification,'' \emph{arXiv:1905.00546},
  2019.

\bibitem{zhu2003semi}
X.~Zhu, Z.~Ghahramani, and J.~D. Lafferty, ``Semi-supervised learning using
  gaussian fields and harmonic functions,'' in \emph{ICML}, 2003.

\bibitem{belkin06a}
M.~Belkin, P.~Niyogi, and V.~Sindhwani, ``Manifold regularization: A geometric
  framework for learning from labeled and unlabeled examples,'' \emph{JMLR},
  2006.

\bibitem{wang2013dynamic}
B.~Wang, Z.~Tu, and J.~K. Tsotsos, ``Dynamic label propagation for
  semi-supervised multi-class multi-label classification,'' in \emph{CVPR},
  2013.

\bibitem{jiang2019semi}
B.~Jiang, Z.~Zhang, D.~Lin, J.~Tang, and B.~Luo, ``Semi-supervised learning
  with graph learning-convolutional networks,'' in \emph{CVPR}, 2019.

\bibitem{bromley1993signature}
J.~Bromley, J.~W. Bentz, L.~Bottou, I.~Guyon, Y.~LeCun, C.~Moore,
  E.~S{\"a}ckinger, and R.~Shah, ``Signature verification using a “siamese”
  time delay neural network,'' \emph{IJPRAI}, 1993.

\bibitem{hadsell2006dimensionality}
R.~Hadsell, S.~Chopra, and Y.~LeCun, ``Dimensionality reduction by learning an
  invariant mapping,'' in \emph{CVPR}, 2006.

\bibitem{lin2020shoestring}
W.~Lin, Z.~Gao, and B.~Li, ``Shoestring: Graph-based semi-supervised
  classification with severely limited labeled data,'' in \emph{CVPR}, 2020.

\bibitem{li2020density}
S.~Li, B.~Liu, D.~Chen, Q.~Chu, L.~Yuan, and N.~Yu, ``Density-aware graph for
  deep semi-supervised visual recognition,'' in \emph{CVPR}, 2020.

\bibitem{kingma2013auto}
D.~P. Kingma and M.~Welling, ``Auto-encoding variational bayes,''
  \emph{arXiv:1312.6114}, 2013.

\bibitem{goodfellow2014generative}
I.~Goodfellow, J.~Pouget-Abadie, M.~Mirza, B.~Xu, D.~Warde-Farley, S.~Ozair,
  A.~Courville, and Y.~Bengio, ``Generative adversarial nets,'' in
  \emph{NeurIPS}, 2014.

\bibitem{doersch2016tutorial}
C.~Doersch, ``Tutorial on variational autoencoders,'' \emph{arXiv:1606.05908},
  2016.

\bibitem{ehsan2017infinite}
M.~Ehsan~Abbasnejad, A.~Dick, and A.~van~den Hengel, ``Infinite variational
  autoencoder for semi-supervised learning,'' in \emph{CVPR}, 2017.

\bibitem{kumar2017semi}
A.~Kumar, P.~Sattigeri, and T.~Fletcher, ``Semi-supervised learning with gans:
  Manifold invariance with improved inference,'' in \emph{NeurIPS}, 2017.

\bibitem{li2017triple}
C.~Li, T.~Xu, J.~Zhu, and B.~Zhang, ``Triple generative adversarial nets,'' in
  \emph{NeurIPS}, 2017.

\bibitem{RAYLLS16}
S.~Reed, Z.~Akata, X.~Yan, L.~Logeswaran, B.~Schiele, and H.~Lee, ``Generative
  adversarial text to image synthesis.'' in \emph{ICML}, 2016.

\bibitem{RAMTLS16}
S.~Reed, Z.~Akata, S.~Mohan, S.~Tenka, B.~Schiele, and H.~Lee, ``Learning what
  and where to draw.'' in \emph{NIPS}, 2016.

\bibitem{XLSA18}
Y.~Xian, T.~Lorenz, B.~Schiele, and Z.~Akata, ``Feature generating networks for
  zero-shot learning,'' in \emph{CVPR}, 2018.

\bibitem{XSSA19}
Y.~Xian, S.~Sharma, B.~Schiele, and Z.~Akata, ``F-vaegan-d2: A feature
  generating framework for any-shot learning,'' in \emph{CVPR}, 2019.

\bibitem{dosovitskiy2014discriminative}
A.~Dosovitskiy, J.~T. Springenberg, M.~Riedmiller, and T.~Brox,
  ``Discriminative unsupervised feature learning with convolutional neural
  networks,'' in \emph{NeurIPS}, 2014.

\bibitem{dosovitskiy2015discriminative}
A.~Dosovitskiy, P.~Fischer, J.~T. Springenberg, M.~Riedmiller, and T.~Brox,
  ``Discriminative unsupervised feature learning with exemplar convolutional
  neural networks,'' \emph{IEEE TPAMI}, 2015.

\bibitem{noroozi2016unsupervised}
M.~Noroozi and P.~Favaro, ``Unsupervised learning of visual representations by
  solving jigsaw puzzles,'' in \emph{ECCV}, 2016.

\bibitem{noroozi2017representation}
M.~Noroozi, H.~Pirsiavash, and P.~Favaro, ``Representation learning by learning
  to count,'' in \emph{ICCV}, 2017.

\bibitem{gidaris2018unsupervised}
S.~Gidaris, P.~Singh, and N.~Komodakis, ``Unsupervised representation learning
  by predicting image rotations,'' in \emph{ICLR}, 2018.

\bibitem{hinton2006reducing}
G.~E. Hinton and R.~R. Salakhutdinov, ``Reducing the dimensionality of data
  with neural networks,'' \emph{Science}, 2006.

\bibitem{masci2011stacked}
J.~Masci, U.~Meier, D.~Cire{\c{s}}an, and J.~Schmidhuber, ``Stacked
  convolutional auto-encoders for hierarchical feature extraction,'' in
  \emph{ICANN}, 2011.

\bibitem{pathak2016context}
D.~Pathak, P.~Krahenbuhl, J.~Donahue, T.~Darrell, and A.~A. Efros, ``Context
  encoders: Feature learning by inpainting,'' in \emph{CVPR}, 2016.

\bibitem{he2022masked}
K.~He, X.~Chen, S.~Xie, Y.~Li, P.~Doll{\'a}r, and R.~Girshick, ``Masked
  autoencoders are scalable vision learners,'' in \emph{CVPR}, 2022.

\bibitem{vincent2008extracting}
P.~Vincent, H.~Larochelle, Y.~Bengio, and P.-A. Manzagol, ``Extracting and
  composing robust features with denoising autoencoders,'' in \emph{ICML},
  2008.

\bibitem{zhang2016colorful}
R.~Zhang, P.~Isola, and A.~A. Efros, ``Colorful image colorization,'' in
  \emph{ECCV}, 2016.

\bibitem{zhang2017split}
------, ``Split-brain autoencoders: Unsupervised learning by cross-channel
  prediction,'' in \emph{CVPR}, 2017.

\bibitem{larsson2017colorization}
G.~Larsson, M.~Maire, and G.~Shakhnarovich, ``Colorization as a proxy task for
  visual understanding,'' in \emph{CVPR}, 2017.

\bibitem{wu2018unsupervised}
Z.~Wu, Y.~Xiong, S.~X. Yu, and D.~Lin, ``Unsupervised feature learning via
  non-parametric instance discrimination,'' in \emph{CVPR}, 2018.

\bibitem{misra2020self}
I.~Misra and L.~v.~d. Maaten, ``Self-supervised learning of pretext-invariant
  representations,'' in \emph{CVPR}, 2020.

\bibitem{tian2019contrastive}
Y.~Tian, D.~Krishnan, and P.~Isola, ``Contrastive multiview coding,'' in
  \emph{ECCV}, 2020.

\bibitem{chen2020exploring}
X.~Chen and K.~He, ``Exploring simple siamese representation learning,'' in
  \emph{CVPR}, 2021.

\bibitem{grill2020bootstrap}
J.-B. Grill, F.~Strub, F.~Altch{\'e}, C.~Tallec, P.~H. Richemond,
  E.~Buchatskaya, C.~Doersch, B.~A. Pires, Z.~D. Guo, M.~G. Azar \emph{et~al.},
  ``Bootstrap your own latent: A new approach to self-supervised learning,'' in
  \emph{NeurIPS}, 2020.

\bibitem{tian2021understanding}
Y.~Tian, X.~Chen, and S.~Ganguli, ``Understanding self-supervised learning
  dynamics without contrastive pairs,'' in \emph{ICML}, 2021.

\bibitem{caron2018deep}
M.~Caron, P.~Bojanowski, A.~Joulin, and M.~Douze, ``Deep clustering for
  unsupervised learning of visual features,'' in \emph{ECCV}, 2018.

\bibitem{yang2016joint}
J.~Yang, D.~Parikh, and D.~Batra, ``Joint unsupervised learning of deep
  representations and image clusters,'' in \emph{CVPR}, 2016.

\bibitem{asano2019self}
Y.~Asano, C.~Rupprecht, and A.~Vedaldi, ``Self-labelling via simultaneous
  clustering and representation learning,'' in \emph{ICLR}, 2020.

\bibitem{ji2019invariant}
X.~Ji, J.~F. Henriques, and A.~Vedaldi, ``Invariant information clustering for
  unsupervised image classification and segmentation,'' in \emph{CVPR}, 2019.

\bibitem{huang2020deep}
J.~Huang, S.~Gong, and X.~Zhu, ``Deep semantic clustering by partition
  confidence maximisation,'' in \emph{CVPR}, 2020.

\bibitem{haeusser2018associative}
P.~Haeusser, J.~Plapp, V.~Golkov, E.~Aljalbout, and D.~Cremers, ``Associative
  deep clustering: Training a classification network with no labels,'' in
  \emph{GCPR}, 2018.

\bibitem{caron2020unsupervised}
M.~Caron, I.~Misra, J.~Mairal, P.~Goyal, P.~Bojanowski, and A.~Joulin,
  ``Unsupervised learning of visual features by contrasting cluster
  assignments,'' in \emph{NeurIPS}, 2020.

\bibitem{radford2015unsupervised}
A.~Radford, L.~Metz, and S.~Chintala, ``Unsupervised representation learning
  with deep convolutional generative adversarial networks,''
  \emph{arXiv:1511.06434}, 2015.

\bibitem{chen2019self}
T.~Chen, X.~Zhai, M.~Ritter, M.~Lucic, and N.~Houlsby, ``Self-supervised gans
  via auxiliary rotation loss,'' in \emph{CVPR}, 2019.

\bibitem{wang2020transformation}
J.~Wang, W.~Zhou, G.-J. Qi, Z.~Fu, Q.~Tian, and H.~Li, ``Transformation gan for
  unsupervised image synthesis and representation learning,'' in \emph{CVPR},
  2020.

\bibitem{brock2018large}
A.~Brock, J.~Donahue, and K.~Simonyan, ``Large scale gan training for high
  fidelity natural image synthesis,'' in \emph{ICLR}, 2019.

\bibitem{donahue2019large}
J.~Donahue and K.~Simonyan, ``Large scale adversarial representation
  learning,'' in \emph{NeurIPS}, 2019.

\bibitem{wang2015unsupervised}
X.~Wang and A.~Gupta, ``Unsupervised learning of visual representations using
  videos,'' in \emph{ICCV}, 2015.

\bibitem{vondrick2018tracking}
C.~Vondrick, A.~Shrivastava, A.~Fathi, S.~Guadarrama, and K.~Murphy, ``Tracking
  emerges by colorizing videos,'' in \emph{ECCV}, 2018.

\bibitem{aganj2018unsupervised}
I.~Aganj, M.~G. Harisinghani, R.~Weissleder, and B.~Fischl, ``Unsupervised
  medical image segmentation based on the local center of mass,''
  \emph{Nature}, 2018.

\bibitem{he2019rethinking}
K.~He, R.~Girshick, and P.~Doll{\'a}r, ``Rethinking imagenet pre-training,'' in
  \emph{CVPR}, 2019.

\bibitem{feichtenhofer2021large}
C.~Feichtenhofer, H.~Fan, B.~Xiong, R.~Girshick, and K.~He, ``A large-scale
  study on unsupervised spatiotemporal representation learning,'' in
  \emph{CVPR}, 2021.

\bibitem{ren2016faster}
S.~Ren, K.~He, R.~Girshick, and J.~Sun, ``Faster r-cnn: Towards real-time
  object detection with region proposal networks,'' \emph{IEEE TPAMI}, 2016.

\bibitem{everingham2010pascal}
M.~Everingham, L.~Van~Gool, C.~K. Williams, J.~Winn, and A.~Zisserman, ``The
  pascal visual object classes (voc) challenge,'' \emph{IJCV}, 2010.

\bibitem{he2017mask}
K.~He, G.~Gkioxari, P.~Doll{\'a}r, and R.~Girshick, ``Mask r-cnn,'' in
  \emph{CVPR}, 2017.

\bibitem{lin2014microsoft}
T.-Y. Lin, M.~Maire, S.~Belongie, J.~Hays, P.~Perona, D.~Ramanan,
  P.~Doll{\'a}r, and C.~L. Zitnick, ``Microsoft coco: Common objects in
  context,'' in \emph{ECCV}, 2014.

\bibitem{santa2018visual}
R.~Santa~Cruz, B.~Fernando, A.~Cherian, and S.~Gould, ``Visual permutation
  learning,'' \emph{IEEE TPAMI}, 2018.

\bibitem{nathan2018improvements}
T.~Nathan~Mundhenk, D.~Ho, and B.~Y. Chen, ``Improvements to context based
  self-supervised learning,'' in \emph{CVPR}, 2018.

\bibitem{wei2019iterative}
C.~Wei, L.~Xie, X.~Ren, Y.~Xia, C.~Su, J.~Liu, Q.~Tian, and A.~L. Yuille,
  ``Iterative reorganization with weak spatial constraints: Solving arbitrary
  jigsaw puzzles for unsupervised representation learning,'' in \emph{CVPR},
  2019.

\bibitem{goyal2019scaling}
P.~Goyal, D.~Mahajan, A.~Gupta, and I.~Misra, ``Scaling and benchmarking
  self-supervised visual representation learning,'' in \emph{ICCV}, 2019.

\bibitem{goh2021multimodal}
G.~Goh, N.~Cammarata, C.~Voss, S.~Carter, M.~Petrov, L.~Schubert, A.~Radford,
  and C.~Olah, ``Multimodal neurons in artificial neural networks,''
  \emph{Distill}, 2021.

\bibitem{carlucci2019domain}
F.~M. Carlucci, A.~D'Innocente, S.~Bucci, B.~Caputo, and T.~Tommasi, ``Domain
  generalization by solving jigsaw puzzles,'' in \emph{CVPR}, 2019.

\bibitem{hjelm2018learning}
R.~D. Hjelm, A.~Fedorov, S.~Lavoie-Marchildon, K.~Grewal, P.~Bachman,
  A.~Trischler, and Y.~Bengio, ``Learning deep representations by mutual
  information estimation and maximization,'' in \emph{ICLR}, 2019.

\bibitem{bachman2019learning}
P.~Bachman, R.~D. Hjelm, and W.~Buchwalter, ``Learning representations by
  maximizing mutual information across views,'' in \emph{NeurIPS}, 2019.

\bibitem{tschannen2019mutual}
M.~Tschannen, J.~Djolonga, P.~K. Rubenstein, S.~Gelly, and M.~Lucic, ``On
  mutual information maximization for representation learning,'' in
  \emph{ICLR}, 2019.

\bibitem{tian2020makes}
Y.~Tian, C.~Sun, B.~Poole, D.~Krishnan, C.~Schmid, and P.~Isola, ``What makes
  for good views for contrastive learning,'' in \emph{NeurIPS}, 2020.

\bibitem{huang2019unsupervised}
J.~Huang, Q.~Dong, S.~Gong, and X.~Zhu, ``Unsupervised deep learning by
  neighbourhood discovery,'' in \emph{ICML}, 2019.

\bibitem{van2020learning}
W.~Van~Gansbeke, S.~Vandenhende, S.~Georgoulis, M.~Proesmans, and L.~Van~Gool,
  ``Learning to classify images without labels,'' in \emph{ECCV}, 2020.

\bibitem{caron2019unsupervised}
M.~Caron, P.~Bojanowski, J.~Mairal, and A.~Joulin, ``Unsupervised pre-training
  of image features on non-curated data,'' in \emph{ICCV}, 2019.

\bibitem{novotny2018self}
D.~Novotny, S.~Albanie, D.~Larlus, and A.~Vedaldi, ``Self-supervised learning
  of geometrically stable features through probabilistic introspection,'' in
  \emph{CVPR}, 2018.

\bibitem{ye2019unsupervised}
M.~Ye, X.~Zhang, P.~C. Yuen, and S.-F. Chang, ``Unsupervised embedding learning
  via invariant and spreading instance feature,'' in \emph{CVPR}, 2019.

\bibitem{chen2020improved}
X.~Chen, H.~Fan, R.~Girshick, and K.~He, ``Improved baselines with momentum
  contrastive learning,'' \emph{arXiv:2003.04297}, 2020.

\bibitem{gutmann2010noise}
M.~Gutmann and A.~Hyv{\"a}rinen, ``Noise-contrastive estimation: A new
  estimation principle for unnormalized statistical models,'' in
  \emph{AISTATS}, 2010.

\bibitem{FDFKA20}
M.~Federici, A.~Dutta, P.~Forré, N.~Kushman, and Z.~Akata, ``Learning robust
  representations via multi-view information bottleneck,'' in \emph{ICLR},
  2020.

\bibitem{oord2018representation}
A.~v.~d. Oord, Y.~Li, and O.~Vinyals, ``Representation learning with
  contrastive predictive coding,'' \emph{arXiv:1807.03748}, 2018.

\bibitem{robinson2020contrastive}
J.~Robinson, C.-Y. Chuang, S.~Sra, and S.~Jegelka, ``Contrastive learning with
  hard negative samples,'' in \emph{ICLR}, 2021.

\bibitem{hu2021adco}
Q.~Hu, X.~Wang, W.~Hu, and G.-J. Qi, ``Adco: Adversarial contrast for efficient
  learning of unsupervised representations from self-trained negative
  adversaries,'' in \emph{CVPR}, 2021.

\bibitem{wang2022caco}
X.~Wang, Y.~Huang, D.~Zeng, and G.-J. Qi, ``Caco: Both positive and negative
  samples are directly learnable via cooperative-adversarial contrastive
  learning,'' \emph{arXiv:2203.14370}, 2022.

\bibitem{zbontar2021barlow}
J.~Zbontar, L.~Jing, I.~Misra, Y.~LeCun, and S.~Deny, ``Barlow twins:
  Self-supervised learning via redundancy reduction,'' in \emph{ICML}, 2021.

\bibitem{xie2016unsupervised}
J.~Xie, R.~Girshick, and A.~Farhadi, ``Unsupervised deep embedding for
  clustering analysis,'' in \emph{ICML}, 2016.

\bibitem{hu2017learning}
W.~Hu, T.~Miyato, S.~Tokui, E.~Matsumoto, and M.~Sugiyama, ``Learning discrete
  representations via information maximizing self-augmented training,'' in
  \emph{ICML}, 2017.

\bibitem{zhuang2019local}
C.~Zhuang, A.~L. Zhai, and D.~Yamins, ``Local aggregation for unsupervised
  learning of visual embeddings,'' in \emph{CVPR}, 2019.

\bibitem{yan2020clusterfit}
X.~Yan, I.~Misra, A.~Gupta, D.~Ghadiyaram, and D.~Mahajan, ``Clusterfit:
  Improving generalization of visual representations,'' in \emph{CVPR}, 2020.

\bibitem{wang2020unsupervised}
X.~Wang, Z.~Liu, and S.~X. Yu, ``Unsupervised feature learning by cross-level
  discrimination between instances and groups,'' in \emph{CVPR}, 2021.

\bibitem{gidaris2020learning}
S.~Gidaris, A.~Bursuc, N.~Komodakis, P.~P{\'e}rez, and M.~Cord, ``Learning
  representations by predicting bags of visual words,'' in \emph{CVPR}, 2020.

\bibitem{jain1999data}
A.~K. Jain, M.~N. Murty, and P.~J. Flynn, ``Data clustering: a review,''
  \emph{CSUR}, 1999.

\bibitem{coates2012learning}
A.~Coates and A.~Y. Ng, ``Learning feature representations with k-means,'' in
  \emph{Neural networks: Tricks of the trade}, 2012.

\bibitem{von2007tutorial}
U.~Von~Luxburg, ``A tutorial on spectral clustering,'' \emph{Statistics and
  computing}, 2007.

\bibitem{chang2017deep}
J.~Chang, L.~Wang, G.~Meng, S.~Xiang, and C.~Pan, ``Deep adaptive image
  clustering,'' in \emph{CVPR}, 2017.

\bibitem{guo2017improved}
X.~Guo, L.~Gao, X.~Liu, and J.~Yin, ``Improved deep embedded clustering with
  local structure preservation,'' in \emph{IJCAI}, 2017.

\bibitem{yang2017towards}
B.~Yang, X.~Fu, N.~D. Sidiropoulos, and M.~Hong, ``Towards k-means-friendly
  spaces: Simultaneous deep learning and clustering,'' in \emph{ICML}, 2017.

\bibitem{gowda1978agglomerative}
K.~C. Gowda and G.~Krishna, ``Agglomerative clustering using the concept of
  mutual nearest neighbourhood,'' \emph{PR}, 1978.

\bibitem{cuturi2013sinkhorn}
M.~Cuturi, ``Sinkhorn distances: Lightspeed computation of optimal transport,''
  in \emph{NeurIPS}, 2013.

\bibitem{li2020contrastive}
Y.~Li, P.~Hu, Z.~Liu, D.~Peng, J.~T. Zhou, and X.~Peng, ``Contrastive
  clustering,'' in \emph{AAAI}, 2020.

\bibitem{hinton2006fast}
G.~E. Hinton, S.~Osindero, and Y.-W. Teh, ``A fast learning algorithm for deep
  belief nets,'' \emph{Neural computation}, 2006.

\bibitem{gabbay2019demystifying}
A.~Gabbay and Y.~Hoshen, ``Demystifying inter-class disentanglement,'' in
  \emph{ICLR}, 2020.

\bibitem{donahue2016adversarial}
J.~Donahue, P.~Kr{\"a}henb{\"u}hl, and T.~Darrell, ``Adversarial feature
  learning,'' in \emph{ICLR}, 2017.

\bibitem{qi2019avt}
G.-J. Qi, L.~Zhang, C.~W. Chen, and Q.~Tian, ``Avt: Unsupervised learning of
  transformation equivariant representations by autoencoding variational
  transformations,'' in \emph{ICCV}, 2019.

\bibitem{zhang2019aet}
L.~Zhang, G.-J. Qi, L.~Wang, and J.~Luo, ``Aet vs. aed: Unsupervised
  representation learning by auto-encoding transformations rather than data,''
  in \emph{CVPR}, 2019.

\bibitem{chen2021semi}
X.~Chen, Y.~Yuan, G.~Zeng, and J.~Wang, ``Semi-supervised semantic segmentation
  with cross pseudo supervision,'' in \emph{CVPR}, 2021.

\bibitem{jeong2019consistency}
J.~Jeong, S.~Lee, J.~Kim, and N.~Kwak, ``Consistency-based semi-supervised
  learning for object detection,'' in \emph{NeurIPS}, 2019.

\bibitem{sun2019unsupervised-self}
Y.~Sun, E.~Tzeng, T.~Darrell, and A.~A. Efros, ``Unsupervised domain adaptation
  through self-supervision,'' \emph{arXiv:1909.11825}, 2019.

\bibitem{saito2020universal}
K.~Saito, D.~Kim, S.~Sclaroff, and K.~Saenko, ``Universal domain adaptation
  through self supervision,'' in \emph{NeurIPS}, 2020.

\bibitem{yang2021semihand}
L.~Yang, S.~Chen, and A.~Yao, ``Semihand: Semi-supervised hand pose estimation
  with consistency,'' in \emph{ICCV}, 2021.

\bibitem{kim2021just}
T.~Kim, J.~Choi, S.~Choi, D.~Jung, and C.~Kim, ``Just a few points are all you
  need for multi-view stereo: A novel semi-supervised learning method for
  multi-view stereo,'' in \emph{ICCV}, 2021.

\bibitem{alwassel2019self}
H.~Alwassel, D.~Mahajan, B.~Korbar, L.~Torresani, B.~Ghanem, and D.~Tran,
  ``Self-supervised learning by cross-modal audio-video clustering,'' in
  \emph{NeurIPS}, 2020.

\bibitem{ronneberger2015u}
O.~Ronneberger, P.~Fischer, and T.~Brox, ``U-net: Convolutional networks for
  biomedical image segmentation,'' in \emph{MICCAI}, 2015.

\bibitem{huo2021atso}
X.~Huo, L.~Xie, J.~He, Z.~Yang, W.~Zhou, H.~Li, and Q.~Tian, ``Atso:
  Asynchronous teacher-student optimization for semi-supervised image
  segmentation,'' in \emph{CVPR}, 2021.

\bibitem{wu2021collaborative}
H.~Wu, G.~Chen, Z.~Wen, and J.~Qin, ``Collaborative and adversarial learning of
  focused and dispersive representations for semi-supervised polyp
  segmentation,'' in \emph{ICCV}, 2021.

\bibitem{huang2021graph}
H.~Huang, L.~Lin, Y.~Zhang, Y.~Xu, J.~Zheng, X.~Mao, X.~Qian, Z.~Peng, J.~Zhou,
  Y.-W. Chen \emph{et~al.}, ``Graph-bas3net: Boundary-aware semi-supervised
  segmentation network with bilateral graph convolution,'' in \emph{ICCV},
  2021.

\bibitem{cordts2016cityscapes}
M.~Cordts, M.~Omran, S.~Ramos, T.~Rehfeld, M.~Enzweiler, R.~Benenson,
  U.~Franke, S.~Roth, and B.~Schiele, ``The cityscapes dataset for semantic
  urban scene understanding,'' in \emph{CVPR}, 2016.

\bibitem{yang2018denseaspp}
M.~Yang, K.~Yu, C.~Zhang, Z.~Li, and K.~Yang, ``Denseaspp for semantic
  segmentation in street scenes,'' in \emph{CVPR}, 2018.

\bibitem{behley2019semantickitti}
J.~Behley, M.~Garbade, A.~Milioto, J.~Quenzel, S.~Behnke, C.~Stachniss, and
  J.~Gall, ``Semantickitti: A dataset for semantic scene understanding of lidar
  sequences,'' in \emph{ICCV}, 2019.

\bibitem{french2019semi}
G.~French, T.~Aila, S.~Laine, M.~Mackiewicz, and G.~Finlayson,
  ``Semi-supervised semantic segmentation needs strong, high-dimensional
  perturbations,'' in \emph{BMVC}, 2020.

\bibitem{ouali2020semi}
Y.~Ouali, C.~Hudelot, and M.~Tami, ``Semi-supervised semantic segmentation with
  cross-consistency training,'' in \emph{CVPR}, 2020.

\bibitem{ke2020guided}
Z.~Ke, D.~Qiu, K.~Li, Q.~Yan, and R.~W. Lau, ``Guided collaborative training
  for pixel-wise semi-supervised learning,'' in \emph{ECCV}, 2020.

\bibitem{hu2021semi}
H.~Hu, F.~Wei, H.~Hu, Q.~Ye, J.~Cui, and L.~Wang, ``Semi-supervised semantic
  segmentation via adaptive equalization learning,'' in \emph{NeurIPS}, 2021.

\bibitem{mendel2020semi}
R.~Mendel, L.~A. De~Souza, D.~Rauber, J.~P. Papa, and C.~Palm,
  ``Semi-supervised segmentation based on error-correcting supervision,'' in
  \emph{ECCV}, 2020.

\bibitem{zou2020pseudoseg}
Y.~Zou, Z.~Zhang, H.~Zhang, C.-L. Li, X.~Bian, J.-B. Huang, and T.~Pfister,
  ``Pseudoseg: Designing pseudo labels for semantic segmentation,'' in
  \emph{ICLR}, 2021.

\bibitem{ibrahim2020semi}
M.~S. Ibrahim, A.~Vahdat, M.~Ranjbar, and W.~G. Macready, ``Semi-supervised
  semantic image segmentation with self-correcting networks,'' in \emph{CVPR},
  2020.

\bibitem{he2021re}
R.~He, J.~Yang, and X.~Qi, ``Re-distributing biased pseudo labels for
  semi-supervised semantic segmentation: A baseline investigation,'' in
  \emph{ICCV}, 2021.

\bibitem{yuan2021simple}
J.~Yuan, Y.~Liu, C.~Shen, Z.~Wang, and H.~Li, ``A simple baseline for
  semi-supervised semantic segmentation with strong data augmentation,'' in
  \emph{ICCV}, 2021.

\bibitem{souly2017semi}
N.~Souly, C.~Spampinato, and M.~Shah, ``Semi supervised semantic segmentation
  using generative adversarial network,'' in \emph{ICCV}, 2017.

\bibitem{hung2018adversarial}
W.-C. Hung, Y.-H. Tsai, Y.-T. Liou, Y.-Y. Lin, and M.-H. Yang, ``Adversarial
  learning for semi-supervised semantic segmentation,'' in \emph{BMVC}, 2018.

\bibitem{mittal2019semi}
S.~Mittal, M.~Tatarchenko, and T.~Brox, ``Semi-supervised semantic segmentation
  with high-and low-level consistency,'' \emph{IEEE TPAMI}, 2021.

\bibitem{lai2021semi}
X.~Lai, Z.~Tian, L.~Jiang, S.~Liu, H.~Zhao, L.~Wang, and J.~Jia,
  ``Semi-supervised semantic segmentation with directional context-aware
  consistency,'' in \emph{CVPR}, 2021.

\bibitem{alonso2021semi}
I.~Alonso, A.~Sabater, D.~Ferstl, L.~Montesano, and A.~C. Murillo,
  ``Semi-supervised semantic segmentation with pixel-level contrastive learning
  from a class-wise memory bank,'' in \emph{ICCV}, 2021.

\bibitem{zhong2021pixel}
Y.~Zhong, B.~Yuan, H.~Wu, Z.~Yuan, J.~Peng, and Y.-X. Wang, ``Pixel
  contrastive-consistent semi-supervised semantic segmentation,'' in
  \emph{ICCV}, 2021.

\bibitem{zhou2021c3}
Y.~Zhou, H.~Xu, W.~Zhang, B.~Gao, and P.-A. Heng, ``C3-semiseg: Contrastive
  semi-supervised segmentation via cross-set learning and dynamic
  class-balancing,'' in \emph{ICCV}, 2021.

\bibitem{yun2019cutmix}
S.~Yun, D.~Han, S.~J. Oh, S.~Chun, J.~Choe, and Y.~Yoo, ``Cutmix:
  Regularization strategy to train strong classifiers with localizable
  features,'' in \emph{ICCV}, 2019.

\bibitem{carion2020end}
N.~Carion, F.~Massa, G.~Synnaeve, N.~Usunier, A.~Kirillov, and S.~Zagoruyko,
  ``End-to-end object detection with transformers,'' in \emph{ECCV}, 2020.

\bibitem{xiao2017joint}
T.~Xiao, S.~Li, B.~Wang, L.~Lin, and X.~Wang, ``Joint detection and
  identification feature learning for person search,'' in \emph{CVPR}, 2017.

\bibitem{qian2021robust}
K.~Qian, S.~Zhu, X.~Zhang, and L.~E. Li, ``Robust multimodal vehicle detection
  in foggy weather using complementary lidar and radar signals,'' in
  \emph{CVPR}, 2021.

\bibitem{su2021multi}
H.~Su, S.~Gong, and X.~Zhu, ``Multi-perspective cross-class domain adaptation
  for open logo detection,'' \emph{CVIU}, 2021.

\bibitem{feng2021semantic}
W.~Feng, F.~Yin, X.-Y. Zhang, and C.-L. Liu, ``Semantic-aware video text
  detection,'' in \emph{CVPR}, 2021.

\bibitem{russakovsky2015best}
O.~Russakovsky, L.-J. Li, and L.~Fei-Fei, ``Best of both worlds: human-machine
  collaboration for object annotation,'' in \emph{CVPR}, 2015.

\bibitem{jeong2021interpolation}
J.~Jeong, V.~Verma, M.~Hyun, J.~Kannala, and N.~Kwak, ``Interpolation-based
  semi-supervised learning for object detection,'' in \emph{CVPR}, 2021.

\bibitem{liu2021unbiased}
Y.-C. Liu, C.-Y. Ma, Z.~He, C.-W. Kuo, K.~Chen, P.~Zhang, B.~Wu, Z.~Kira, and
  P.~Vajda, ``Unbiased teacher for semi-supervised object detection,'' in
  \emph{ICLR}, 2021.

\bibitem{zhou2021instant}
Q.~Zhou, C.~Yu, Z.~Wang, Q.~Qian, and H.~Li, ``Instant-teaching: An end-to-end
  semi-supervised object detection framework,'' in \emph{CVPR}, 2021.

\bibitem{xu2021end}
M.~Xu, Z.~Zhang, H.~Hu, J.~Wang, L.~Wang, F.~Wei, X.~Bai, and Z.~Liu,
  ``End-to-end semi-supervised object detection with soft teacher,'' in
  \emph{ICCV}, 2021.

\bibitem{sohn2020simple}
K.~Sohn, Z.~Zhang, C.-L. Li, H.~Zhang, C.-Y. Lee, and T.~Pfister, ``A simple
  semi-supervised learning framework for object detection,''
  \emph{arXiv:2005.04757}, 2020.

\bibitem{yang2021interactive}
Q.~Yang, X.~Wei, B.~Wang, X.-S. Hua, and L.~Zhang, ``Interactive self-training
  with mean teachers for semi-supervised object detection,'' in \emph{CVPR},
  2021.

\bibitem{wang2021data}
Z.~Wang, Y.~Li, Y.~Guo, L.~Fang, and S.~Wang, ``Data-uncertainty guided
  multi-phase learning for semi-supervised object detection,'' in \emph{CVPR},
  2021.

\bibitem{wang2021combating}
Z.~Wang, Y.~Li, Y.~Guo, and S.~Wang, ``Combating noise: Semi-supervised
  learning by region uncertainty quantification,'' in \emph{NeurIPS}, 2021.

\bibitem{saenko2010adapting}
K.~Saenko, B.~Kulis, M.~Fritz, and T.~Darrell, ``Adapting visual category
  models to new domains,'' in \emph{ECCV}, 2010.

\bibitem{wu2020dualmixup}
Y.~Wu, D.~Inkpen, and A.~El-Roby, ``Dual mixup regularized learning for
  adversarial domain adaptation,'' in \emph{ECCV}, 2020.

\bibitem{french2018self}
G.~French, M.~Mackiewicz, and M.~Fisher, ``Self-ensembling for visual domain
  adaptation,'' in \emph{ICLR}, 2018.

\bibitem{deng2019cluster}
Z.~Deng, Y.~Luo, and J.~Zhu, ``Cluster alignment with a teacher for
  unsupervised domain adaptation,'' in \emph{ICCV}, 2019.

\bibitem{carlucci2017autodial}
F.~M. Carlucci, L.~Porzi, B.~Caputo, E.~Ricci, and S.~R. Bulo, ``Autodial:
  Automatic domain alignment layers,'' in \emph{ICCV}, 2017.

\bibitem{morerio2018minimal}
P.~Morerio, J.~Cavazza, and V.~Murino, ``Minimal-entropy correlation alignment
  for unsupervised deep domain adaptation,'' in \emph{ICLR}, 2018.

\bibitem{hu2020synchronization}
L.~Hu, M.~Kan, S.~Shan, and X.~Chen, ``Unsupervised domain adaptation with
  hierarchical gradient synchronization,'' in \emph{CVPR}, 2020.

\bibitem{saito2017asymmetric}
K.~Saito, Y.~Ushiku, and T.~Harada, ``Asymmetric tri-training for unsupervised
  domain adaptation,'' in \emph{ICML}, 2017.

\bibitem{bousmalis2017unsupervised}
K.~Bousmalis, N.~Silberman, D.~Dohan, D.~Erhan, and D.~Krishnan, ``Unsupervised
  pixel-level domain adaptation with generative adversarial networks,'' in
  \emph{CVPR}, 2017.

\bibitem{sankaranarayanan2018learning}
S.~Sankaranarayanan, Y.~Balaji, A.~Jain, S.~Nam~Lim, and R.~Chellappa,
  ``Learning from synthetic data: Addressing domain shift for semantic
  segmentation,'' in \emph{CVPR}, 2018.

\bibitem{hoffman18cycada}
J.~Hoffman, E.~Tzeng, T.~Park, J.-Y. Zhu, P.~Isola, K.~Saenko, A.~Efros, and
  T.~Darrell, ``{C}y{CADA}: Cycle-consistent adversarial domain adaptation,''
  in \emph{ICML}, 2018.

\bibitem{russo2018sbadagan}
P.~Russo, F.~M. Carlucci, T.~Tommasi, and B.~Caputo, ``From source to target
  and back: symmetric bi-directional adaptive gan,'' in \emph{CVPR}, 2018.

\bibitem{murez2018image}
Z.~Murez, S.~Kolouri, D.~Kriegman, R.~Ramamoorthi, and K.~Kim, ``Image to image
  translation for domain adaptation,'' in \emph{CVPR}, 2018.

\bibitem{chen2019crdoco}
Y.-C. Chen, Y.-Y. Lin, M.-H. Yang, and J.-B. Huang, ``Crdoco: Pixel-level
  domain transfer with cross-domain consistency,'' in \emph{CVPR}, 2019.

\bibitem{yoo2016pixel}
D.~Yoo, N.~Kim, S.~Park, A.~S. Paek, and I.~S. Kweon, ``Pixel-level domain
  transfer,'' in \emph{ECCV}, 2016.

\bibitem{shrivastava2017learning}
A.~Shrivastava, T.~Pfister, O.~Tuzel, J.~Susskind, W.~Wang, and R.~Webb,
  ``Learning from simulated and unsupervised images through adversarial
  training.'' in \emph{CVPR}, 2017.

\bibitem{xu2019self-da}
J.~Xu, L.~Xiao, and A.~M. L{\'o}pez, ``Self-supervised domain adaptation for
  computer vision tasks,'' \emph{IEEE Access}, 2019.

\bibitem{bucci2020self}
S.~Bucci, A.~D'Innocente, Y.~Liao, F.~M. Carlucci, B.~Caputo, and T.~Tommasi,
  ``Self-supervised learning across domains,'' \emph{IEEE TPAMI}, 2020.

\bibitem{su2020gradient}
P.~Su, S.~Tang, P.~Gao, D.~Qiu, N.~Zhao, and X.~Wang, ``Gradient regularized
  contrastive learning for continual domain adaptation,'' in \emph{AAAI}, 2021.

\bibitem{wang2022cross}
R.~Wang, Z.~Wu, Z.~Weng, J.~Chen, G.-J. Qi, and Y.-G. Jiang, ``Cross-domain
  contrastive learning for unsupervised domain adaptation,'' \emph{IEEE
  Transactions on Multimedia}, 2022.

\bibitem{yu2020multi}
Q.~Yu, D.~Ikami, G.~Irie, and K.~Aizawa, ``Multi-task curriculum framework for
  open-set semi-supervised learning,'' in \emph{ECCV}, 2020.

\bibitem{augustin2020out}
M.~Augustin and M.~Hein, ``Out-distribution aware self-training in an open
  world setting,'' \emph{arXiv:2012.12372}, 2020.

\bibitem{huang2021trash}
J.~Huang, C.~Fang, W.~Chen, Z.~Chai, X.~Wei, P.~Wei, L.~Lin, and G.~Li, ``Trash
  to treasure: Harvesting ood data with cross-modal matching for open-set
  semi-supervised learning,'' in \emph{ICCV}, 2021.

\bibitem{saito2021openmatch}
K.~Saito, D.~Kim, and K.~Saenko, ``Openmatch: Open-set consistency
  regularization for semi-supervised learning with outliers,'' in
  \emph{NeurIPS}, 2021.

\bibitem{huang2021universal}
Z.~Huang, C.~Xue, B.~Han, J.~Yang, and C.~Gong, ``Universal semi-supervised
  learning,'' in \emph{Thirty-Fifth Conference on Neural Information Processing
  Systems}, 2021.

\bibitem{cao2021open}
K.~Cao, M.~Brbic, and J.~Leskovec, ``Open-world semi-supervised learning,'' in
  \emph{ICLR}, 2022.

\bibitem{hendrycks2016baseline}
D.~Hendrycks and K.~Gimpel, ``A baseline for detecting misclassified and
  out-of-distribution examples in neural networks,'' in \emph{ICLR}, 2017.

\bibitem{zhong2021openmix}
Z.~Zhong, L.~Zhu, Z.~Luo, S.~Li, Y.~Yang, and N.~Sebe, ``Openmix: Reviving
  known knowledge for discovering novel visual categories in an open world,''
  in \emph{CVPR}, 2021.

\bibitem{hendrycks2018deep}
D.~Hendrycks, M.~Mazeika, and T.~Dietterich, ``Deep anomaly detection with
  outlier exposure,'' in \emph{ICLR}, 2019.

\bibitem{lee2018simple}
K.~Lee, K.~Lee, H.~Lee, and J.~Shin, ``A simple unified framework for detecting
  out-of-distribution samples and adversarial attacks,'' in \emph{NeurIPS},
  2018.

\bibitem{hein2019relu}
M.~Hein, M.~Andriushchenko, and J.~Bitterwolf, ``Why relu networks yield
  high-confidence predictions far away from the training data and how to
  mitigate the problem,'' in \emph{CVPR}, 2019.

\bibitem{kim2020distribution}
J.~Kim, Y.~Hur, S.~Park, E.~Yang, S.~J. Hwang, and J.~Shin, ``Distribution
  aligning refinery of pseudo-label for imbalanced semi-supervised learning,''
  in \emph{NeurIPS}, 2020.

\bibitem{lee2021abc}
H.~Lee, S.~Shin, and H.~Kim, ``Abc: Auxiliary balanced classifier for
  class-imbalanced semi-supervised learning,'' in \emph{NeurIPS}, 2021.

\bibitem{han2019learning}
K.~Han, A.~Vedaldi, and A.~Zisserman, ``Learning to discover novel visual
  categories via deep transfer clustering,'' in \emph{ICCV}, 2019.

\bibitem{han2020automatically}
K.~Han, S.-A. Rebuffi, S.~Ehrhardt, A.~Vedaldi, and A.~Zisserman,
  ``Automatically discovering and learning new visual categories with ranking
  statistics,'' in \emph{ICLR}, 2020.

\bibitem{lee2019overcoming}
K.~Lee, K.~Lee, J.~Shin, and H.~Lee, ``Overcoming catastrophic forgetting with
  unlabeled data in the wild,'' in \emph{ICCV}, 2019.

\bibitem{rao2019continual}
D.~Rao, F.~Visin, A.~A. Rusu, R.~Pascanu, Y.~W. Teh, and R.~Hadsell,
  ``Continual unsupervised representation learning,'' in \emph{NeurIPS}, 2019.

\bibitem{delange2021continual}
M.~Delange, R.~Aljundi, M.~Masana, S.~Parisot, X.~Jia, A.~Leonardis,
  G.~Slabaugh, and T.~Tuytelaars, ``A continual learning survey: Defying
  forgetting in classification tasks,'' \emph{IEEE TPAMI}, 2021.

\bibitem{mccloskey1989catastrophic}
M.~McCloskey and N.~J. Cohen, ``Catastrophic interference in connectionist
  networks: The sequential learning problem,'' in \emph{Psychology of learning
  and motivation}, 1989.

\bibitem{rebuffi2017icarl}
S.-A. Rebuffi, A.~Kolesnikov, G.~Sperl, and C.~H. Lampert, ``icarl: Incremental
  classifier and representation learning,'' in \emph{CVPR}, 2017.

\bibitem{kirkpatrick2017overcoming}
J.~Kirkpatrick, R.~Pascanu, N.~Rabinowitz, J.~Veness, G.~Desjardins, A.~A.
  Rusu, K.~Milan, J.~Quan, T.~Ramalho, A.~Grabska-Barwinska \emph{et~al.},
  ``Overcoming catastrophic forgetting in neural networks,'' \emph{PNAS}, 2017.

\bibitem{chaudhry2018riemannian}
A.~Chaudhry, P.~K. Dokania, T.~Ajanthan, and P.~H. Torr, ``Riemannian walk for
  incremental learning: Understanding forgetting and intransigence,'' in
  \emph{ECCV}, 2018.

\bibitem{aljundi2018memory}
R.~Aljundi, F.~Babiloni, M.~Elhoseiny, M.~Rohrbach, and T.~Tuytelaars, ``Memory
  aware synapses: Learning what (not) to forget,'' in \emph{ECCV}, 2018.

\bibitem{zhang2020class}
J.~Zhang, J.~Zhang, S.~Ghosh, D.~Li, S.~Tasci, L.~Heck, H.~Zhang, and C.-C.~J.
  Kuo, ``Class-incremental learning via deep model consolidation,'' in
  \emph{WACV}, 2020.

\bibitem{li2019incremental}
Y.~Li, Y.~Wang, Q.~Liu, C.~Bi, X.~Jiang, and S.~Sun, ``Incremental
  semi-supervised learning on streaming data,'' \emph{PR}, 2019.

\bibitem{smith2019unsupervised}
J.~Smith, S.~Baer, Z.~Kira, and C.~Dovrolis, ``Unsupervised continual learning
  and self-taught associative memory hierarchies,'' in \emph{ICLRW}, 2019.

\bibitem{sun2020test}
Y.~Sun, X.~Wang, Z.~Liu, J.~Miller, A.~Efros, and M.~Hardt, ``Test-time
  training with self-supervision for generalization under distribution
  shifts,'' in \emph{ICML}, 2020.

\bibitem{varsavsky2020test}
T.~Varsavsky, M.~Orbes-Arteaga, C.~H. Sudre, M.~S. Graham, P.~Nachev, and M.~J.
  Cardoso, ``Test-time unsupervised domain adaptation,'' in \emph{MICCAI},
  2020.

\bibitem{wangtent}
D.~Wang, E.~Shelhamer, S.~Liu, B.~Olshausen, and T.~Darrell, ``Tent: Fully
  test-time adaptation by entropy minimiaztion,'' in \emph{ICLR}, 2021.

\bibitem{hoffman2014continuous}
J.~Hoffman, T.~Darrell, and K.~Saenko, ``Continuous manifold based adaptation
  for evolving visual domains,'' in \emph{CVPR}, 2014.

\bibitem{bobu2018adapting}
A.~Bobu, E.~Tzeng, J.~Hoffman, and T.~Darrell, ``Adapting to continuously
  shifting domains,'' in \emph{ICLRW}, 2018.

\bibitem{arazo2020pseudo}
E.~Arazo, D.~Ortego, P.~Albert, N.~E. O’Connor, and K.~McGuinness,
  ``Pseudo-labeling and confirmation bias in deep semi-supervised learning,''
  in \emph{IJCNN}, 2020.

\bibitem{lu2019vilbert}
J.~Lu, D.~Batra, D.~Parikh, and S.~Lee, ``Vilbert: Pretraining task-agnostic
  visiolinguistic representations for vision-and-language tasks,'' in
  \emph{NeurIPS}, 2019.

\bibitem{tan2019lxmert}
H.~Tan and M.~Bansal, ``Lxmert: Learning cross-modality encoder representations
  from transformers,'' in \emph{ACL}, 2019.

\bibitem{su2019vl}
W.~Su, X.~Zhu, Y.~Cao, B.~Li, L.~Lu, F.~Wei, and J.~Dai, ``Vl-bert:
  Pre-training of generic visual-linguistic representations,'' in \emph{ICLR},
  2019.

\bibitem{chen2020uniter}
Y.-C. Chen, L.~Li, L.~Yu, A.~El~Kholy, F.~Ahmed, Z.~Gan, Y.~Cheng, and J.~Liu,
  ``Uniter: Universal image-text representation learning,'' in \emph{ECCV},
  2020.

\bibitem{li2020unicoder}
G.~Li, N.~Duan, Y.~Fang, M.~Gong, D.~Jiang, and M.~Zhou, ``Unicoder-vl: A
  universal encoder for vision and language by cross-modal pre-training.'' in
  \emph{AAAI}, 2020.

\bibitem{vaswani2017attention}
A.~Vaswani, N.~Shazeer, N.~Parmar, J.~Uszkoreit, L.~Jones, A.~N. Gomez,
  {\L}.~Kaiser, and I.~Polosukhin, ``Attention is all you need,'' in
  \emph{NeurIPS}, 2017.

\bibitem{devlin2018bert}
J.~Devlin, M.-W. Chang, K.~Lee, and K.~Toutanova, ``Bert: Pre-training of deep
  bidirectional transformers for language understanding,'' in \emph{ACL}, 2019.

\bibitem{sun2019videobert}
C.~Sun, A.~Myers, C.~Vondrick, K.~Murphy, and C.~Schmid, ``Videobert: A joint
  model for video and language representation learning,'' in \emph{ICCV}, 2019.

\bibitem{stroud2020learning}
J.~C. Stroud, D.~A. Ross, C.~Sun, J.~Deng, R.~Sukthankar, and C.~Schmid,
  ``Learning video representations from textual web supervision,''
  \emph{arXiv:2007.14937}, 2020.

\bibitem{alayrac2020self}
J.-B. Alayrac, A.~Recasens, R.~Schneider, R.~Arandjelovi{\'c}, J.~Ramapuram,
  J.~De~Fauw, L.~Smaira, S.~Dieleman, and A.~Zisserman, ``Self-supervised
  multimodal versatile networks,'' in \emph{NeurIPS}, 2020.

\bibitem{miech2020end}
A.~Miech, J.-B. Alayrac, L.~Smaira, I.~Laptev, J.~Sivic, and A.~Zisserman,
  ``End-to-end learning of visual representations from uncurated instructional
  videos,'' in \emph{CVPR}, 2020.

\bibitem{rouditchenko2020avlnet}
A.~Rouditchenko, A.~Boggust, D.~Harwath, D.~Joshi, S.~Thomas, K.~Audhkhasi,
  R.~Feris, B.~Kingsbury, M.~Picheny, A.~Torralba \emph{et~al.}, ``Avlnet:
  Learning audio-visual language representations from instructional videos,''
  \emph{arXiv:2006.09199}, 2020.

\bibitem{miech2019howto100m}
A.~Miech, D.~Zhukov, J.-B. Alayrac, M.~Tapaswi, I.~Laptev, and J.~Sivic,
  ``Howto100m: Learning a text-video embedding by watching hundred million
  narrated video clips,'' in \emph{ICCV}, 2019.

\bibitem{radford2021learning}
A.~Radford, J.~W. Kim, C.~Hallacy, A.~Ramesh, G.~Goh, S.~Agarwal, G.~Sastry,
  A.~Askell, P.~Mishkin, J.~Clark \emph{et~al.}, ``Learning transferable visual
  models from natural language supervision,'' in \emph{ICML}, 2021.

\bibitem{jia2021scaling}
C.~Jia, Y.~Yang, Y.~Xia, Y.-T. Chen, Z.~Parekh, H.~Pham, Q.~V. Le, Y.~Sung,
  Z.~Li, and T.~Duerig, ``Scaling up visual and vision-language representation
  learning with noisy text supervision,'' in \emph{ICML}, 2021.

\bibitem{arandjelovic2017look}
R.~Arandjelovic and A.~Zisserman, ``Look, listen and learn,'' in \emph{CVPR},
  2017.

\bibitem{owens2018audio}
A.~Owens and A.~A. Efros, ``Audio-visual scene analysis with self-supervised
  multisensory features,'' in \emph{ECCV}, 2018.

\bibitem{korbar2018cooperative}
B.~Korbar, D.~Tran, and L.~Torresani, ``Cooperative learning of audio and video
  models from self-supervised synchronization,'' in \emph{NeurIPS}, 2018.

\bibitem{morgado2020audio}
P.~Morgado, N.~Vasconcelos, and I.~Misra, ``Audio-visual instance
  discrimination with cross-modal agreement,'' in \emph{CVPR}, 2021.

\bibitem{patrick2020multi}
M.~Patrick, Y.~M. Asano, R.~Fong, J.~F. Henriques, G.~Zweig, and A.~Vedaldi,
  ``Multi-modal self-supervision from generalized data transformations,''
  \emph{arXiv:2003.04298}, 2020.

\bibitem{owens2016ambient}
A.~Owens, J.~Wu, J.~H. McDermott, W.~T. Freeman, and A.~Torralba, ``Ambient
  sound provides supervision for visual learning,'' in \emph{ECCV}, 2016.

\bibitem{hu2020semi}
P.~Hu, H.~Zhu, X.~Peng, and J.~Lin, ``Semi-supervised multi-modal learning with
  balanced spectral decomposition,'' in \emph{AAAI}, 2020.

\bibitem{li2020hda-semantic}
S.~Li, B.~Xie, J.~Wu, Y.~Zhao, C.~H. Liu, and Z.~Ding, ``Simultaneous semantic
  alignment network for heterogeneous domain adaptation,'' in \emph{ACM MM},
  2020.

\bibitem{chen2021distilling}
Y.~Chen, Y.~Xian, A.~S. Koepke, Y.~Shan, and Z.~Akata, ``Distilling
  audio-visual knowledge by compositional contrastive learning,'' in
  \emph{CVPR}, 2021.

\end{thebibliography}

\end{document}